%% file: main.tex
\documentclass[10pt,twocolumn,letterpaper]{article}

\usepackage[pagenumbers]{cvpr}

\input{preamble}
\input{macros}
\definecolor{cvprblue}{rgb}{0.21,0.49,0.74}
\usepackage[pagebackref,breaklinks,colorlinks,allcolors=cvprblue]{hyperref}

\usepackage[most]{tcolorbox}
\definecolor{sbred}{HTML}{94070A}
\definecolor{cardbg}{HTML}{F3E4E4}
\definecolor{cardfg}{HTML}{1C2B33}
\newcommand{\scenebound}{\texorpdfstring{\textcolor{sbred}{SceneBound}}{SceneBound}}
\newcommand{\displayface}{\usefont{OT1}{Inter-TLF}{b}{n}}
\newtcolorbox{titlecard}{enhanced, frame hidden, colback=cardbg, arc=10pt, boxsep=0pt, width=\textwidth,
  left=0.55cm, right=0.55cm, top=0.4cm, bottom=0.4cm, grow to left by=0.55cm, grow to right by=0.55cm}
\newcommand{\afflfont}{\fontsize{10pt}{12pt}\selectfont}
\newcommand{\names}[1]{{\displayface\fontsize{9.5pt}{13pt}\selectfont #1}}
\newcommand{\cardtitle}{{\displayface\fontsize{14.1pt}{16.7pt}\selectfont \scenebound: How Much Should a Streaming\\
  3D Reconstruction Model Remember?\par}}
\newcommand{\emblem}[1]{\raisebox{-0.38\height}[0.62\height][5pt]{\includegraphics[height=1.9\baselineskip]{#1}}}
\makeatletter
\def\@maketitle{%
  \begin{titlecard}
    \color{cardfg}\centering
    \cardtitle
    \vskip 0.12in
    {\fontsize{10.8pt}{13pt}\selectfont\begin{tabular}[t]{c}\@author\end{tabular}\par}
    \vskip 0.16in
    \begin{minipage}{\dimexpr\linewidth-0.5cm\relax}
      \fontsize{11pt}{13.6pt}\selectfont\setlength{\parindent}{0pt}%
      \def\abstract{\ignorespaces}\def\endabstract{\par}%
      \input{sec/0_abstract}    \end{minipage}
    \vskip 0.10in
    \begin{minipage}{\dimexpr\linewidth-0.5cm\relax}
      \fontsize{10.5pt}{16pt}\selectfont
      \hfill\emblem{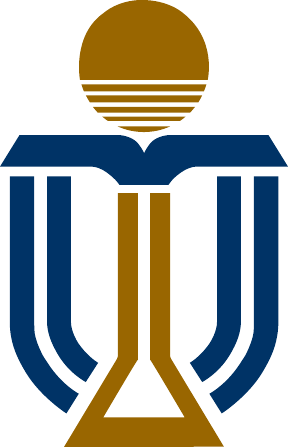}\hspace{0.32cm}\emblem{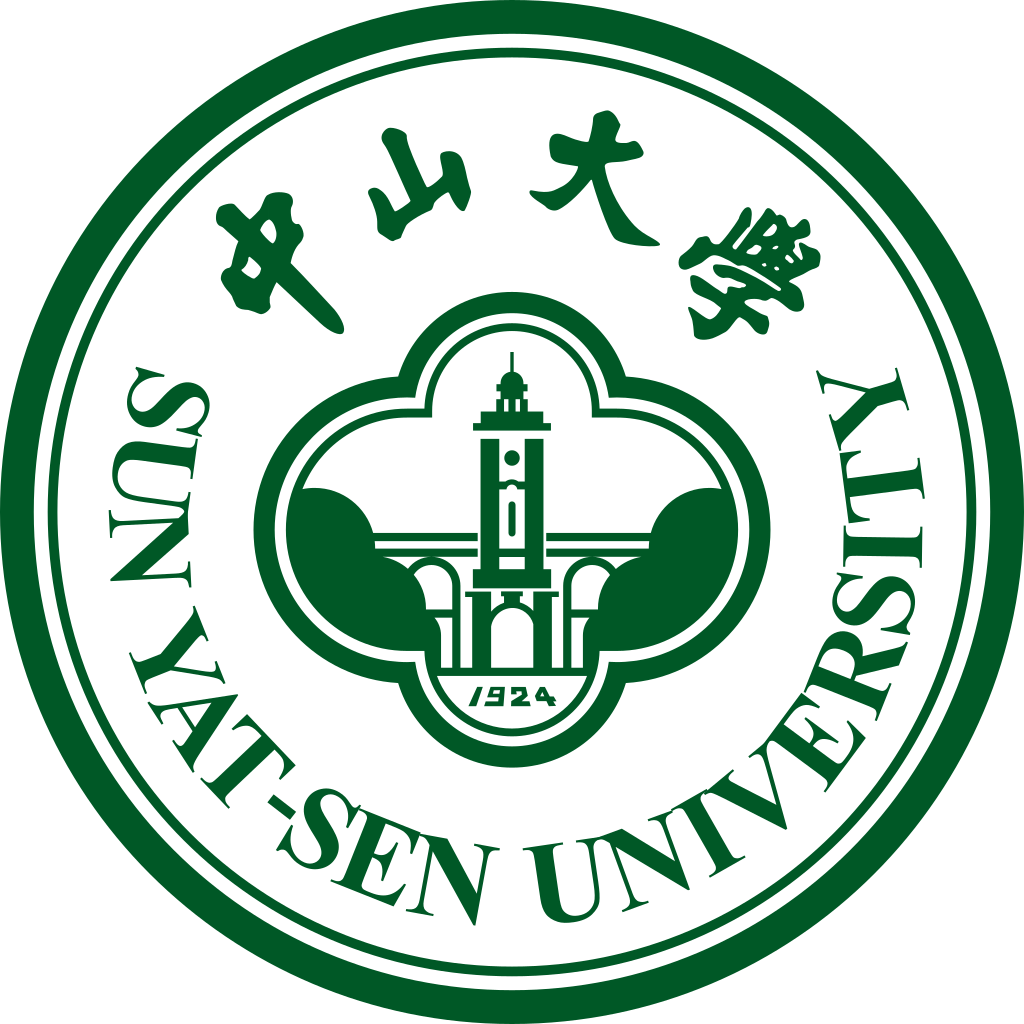}\hspace{0.32cm}\emblem{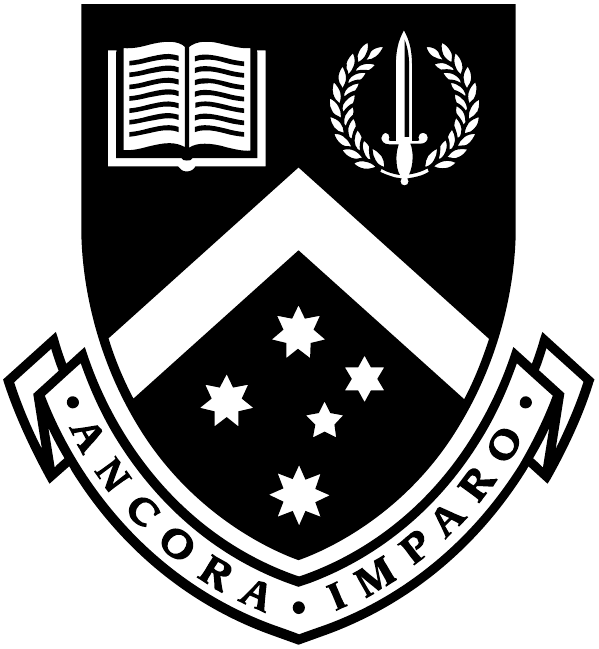}\hspace{0.32cm}%
        \emblem{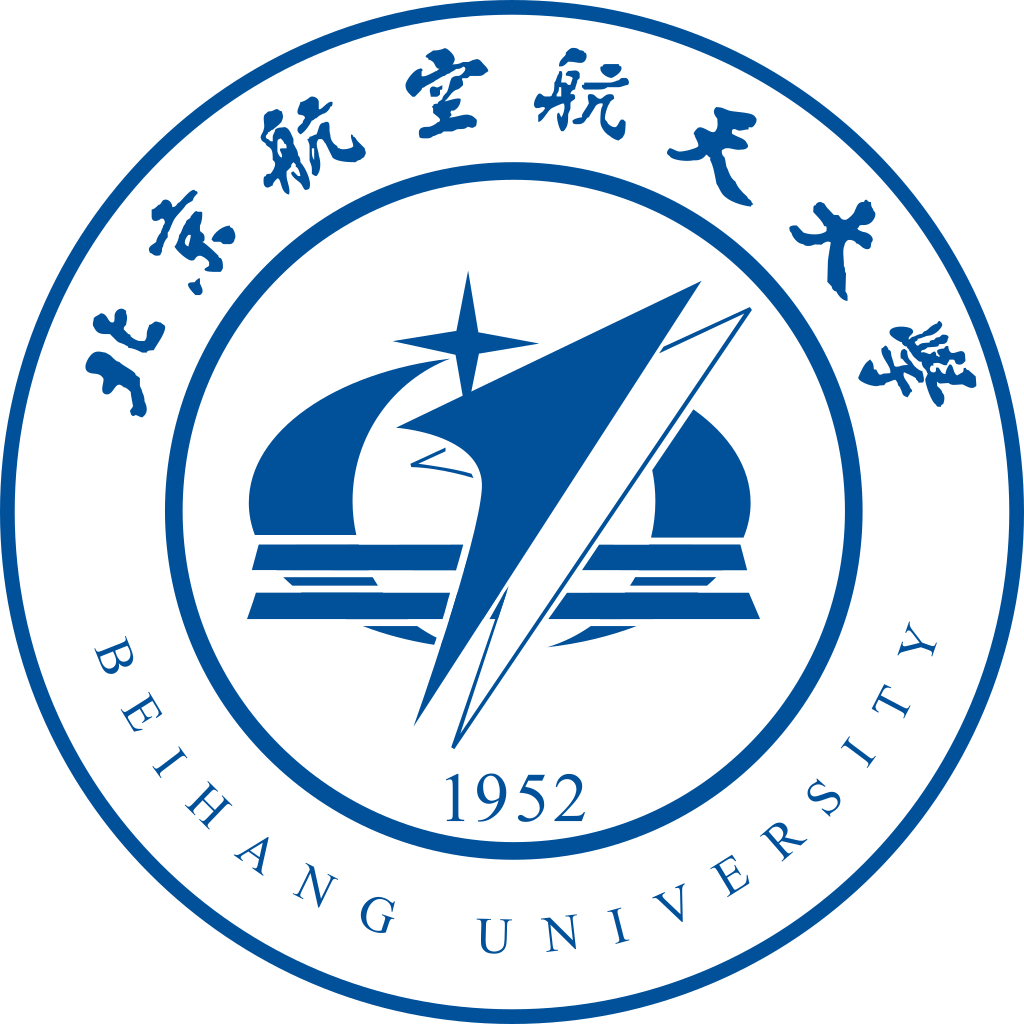}\hspace{0.32cm}\emblem{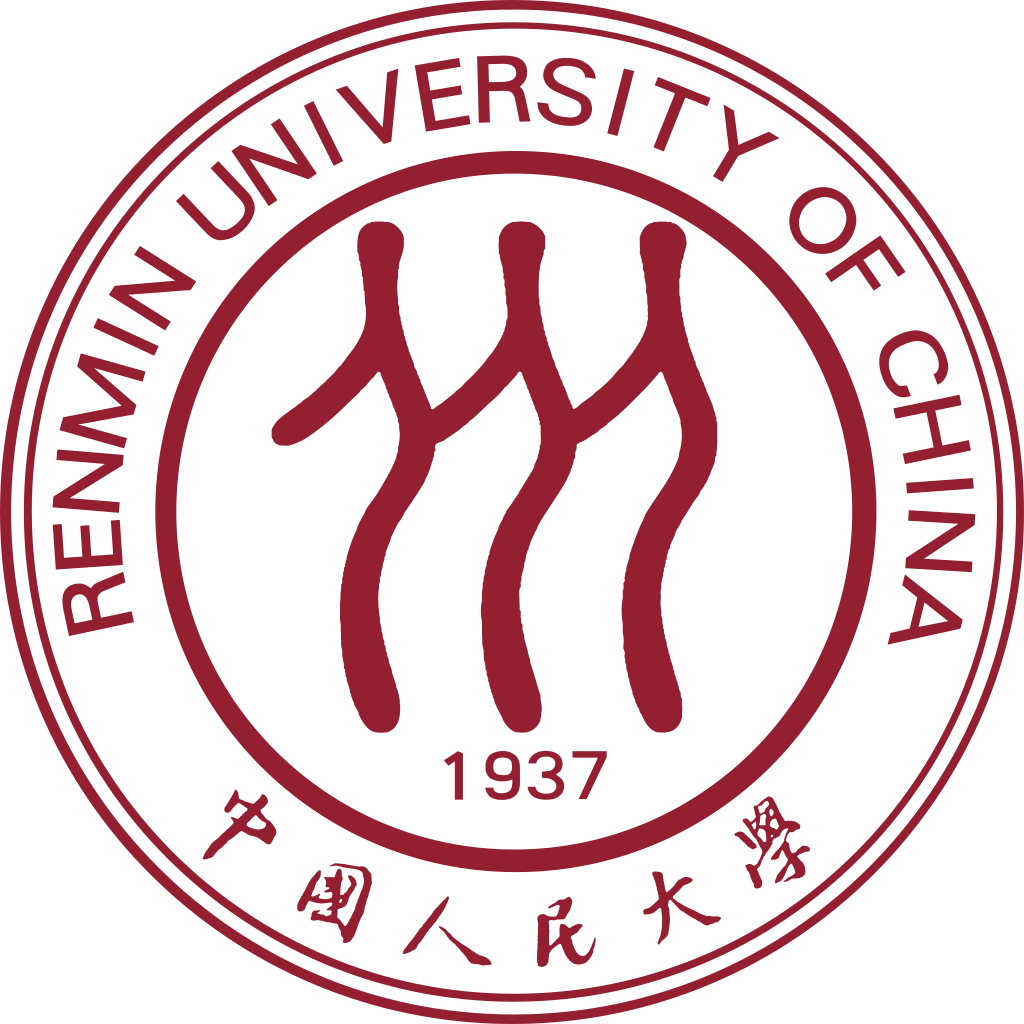}
    \end{minipage}
    \vskip 0.02in
  \end{titlecard}%
}
\def\maketitlesupplementary{%
  \newpage
  \twocolumn[{%
    \begin{titlecard}
      \color{cardfg}\centering
      {\afflfont Supplementary Material\par}
      \vskip 0.06in
      \cardtitle
    \end{titlecard}
    \vskip 12pt}]%
}
\makeatother

\def\confName{CVPR}
\def\confYear{2027}

\title{SceneBound:\\How Much Should a Streaming 3D Reconstruction Model Remember?}
\author{\names{Ziming Wang\textsuperscript{1}, Chuyue Gong\textsuperscript{1}, Chutian Wang\textsuperscript{2}, Jingmin Zhu\textsuperscript{3},}\\
  \names{Hongliang Cao\textsuperscript{4}, Yuheng He\textsuperscript{2}, Xiuyu Li\textsuperscript{5}}\\
  \\
  \afflfont \textsuperscript{1}The Hong Kong University of Science and Technology, \textsuperscript{2}Sun Yat-sen University,\\
  \afflfont \textsuperscript{3}Monash University, \textsuperscript{4}Beihang University, \textsuperscript{5}Renmin University of China}
\hypersetup{pdftitle={SceneBound: How Much Should a Streaming 3D Reconstruction Model Remember?},
  pdfauthor={Ziming Wang, Chuyue Gong, Chutian Wang, Jingmin Zhu, Hongliang Cao, Yuheng He, Xiuyu Li}}

\begin{document}
\maketitle
\input{fig/teaser}
\input{sec/1_intro}
\input{sec/2_related}
\input{sec/3_method}
\input{sec/4_experiments}
\input{sec/5_conclusion}
\input{sec/6_llm}

{
    \small
    \bibliographystyle{ieeenat_fullname}
    \bibliography{references}
}

\maketitlesupplementary
\appendix
\crefalias{section}{appendix}
\counterwithin{proposition}{section}
\renewcommand{\eqlabel}[1]{\tag{\ref*{#1}}}
\input{supp/A_proofs}
\input{supp/B_baselines}
\input{supp/C_method}
\input{supp/D_results}
\input{supp/E_limitations}

\end{document}

%% file: preamble.tex
\usepackage{cuted} 

\usepackage{currfile} 

\usepackage{caption} 



%% file: sec/0_abstract.tex
\begin{abstract}
Streaming 3D reconstruction models keep past video frames in a recurrent state, fast weights from test-time training, or a key-value (KV) cache.
Each written frame shifts the coordinate frame of later cameras or occupies GPU memory.
Training-free rules decide what to write from a signal of the current frame.
How much they write accounts for their mean accuracy: spending the update budget of a gate at uniformly spaced frames, with no signal, matches nine published gates, and the budget bounds the drift of the state wherever the writes fall.
We propose \scenebound{}, a training-free rule that writes a frame only when its predicted camera pose lies at least a fixed radius from every stored keyframe.
Its keyframe count stays within a constant factor of the covering number of the visited poses, the fewest views that keep a keyframe near every camera, so the count grows with the places visited and stays flat when the camera returns.
In large scenes, \scenebound{} closes a submap once decoding its first frame again reveals drift.
\scenebound{} uses one radius across four released models and three memory types.
In the recurrent state of CUT3R and the KV cache of StreamVGGT, it achieves the lowest mean trajectory error among the published training-free rules we evaluate, in one pass and over nine passes of a scene, where its keyframes grow by 3\% to 6\%.
\end{abstract}

%% file: fig/teaser.tex
\begin{strip}
\centering
\includegraphics[width=\linewidth]{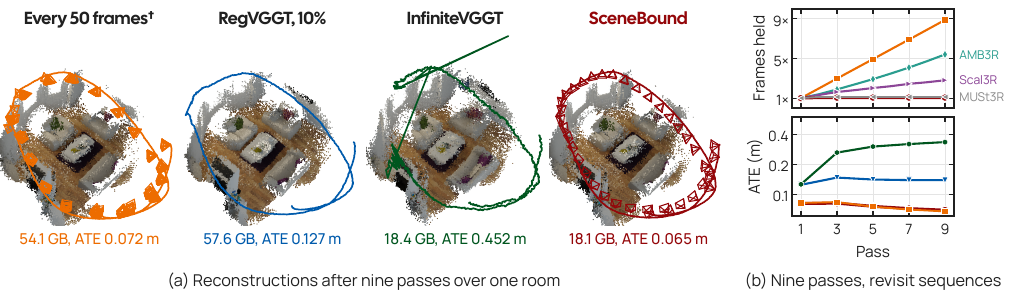}
\captionsetup{hypcap=false}
\captionof{figure}{Over nine passes in the KV cache of StreamVGGT, \scenebound{} holds its memory near the first-pass size at the accuracy of a fixed interval of 50 frames\textsuperscript{\dag}, whose memory grows with every pass; \dag{} marks a constructed baseline. (a) One room with the retained cameras. (b) Sums over the revisit sequences each rule completed.}
\captionsetup{hypcap=true}
\label{fig:\currfilebase}%
\end{strip}

%% file: sec/1_intro.tex
\section{Introduction}
\label{sec:intro}

A home robot vacuuming an apartment or augmented reality glasses worn around an office record video for hours, spending most of that time in the same few rooms.
To act in that space, the device must recover 3D scene structure and track the camera.
Feed-forward networks map a set of images to camera poses, the position and orientation of each view~\cite{wang2024dust3r,leroy2024grounding,wang2025vggt,yang2025fast3r}, and a 3D point per pixel in one pass~\cite{wang2026pi3,keetha2026mapanything,lin2026depth}.
In a video stream, frames arrive in sequence and the full video cannot fit in memory; \emph{streaming 3D reconstruction} predicts the pose and 3D points of each frame online from the frames seen so far~\cite{wang2025continuous,zhuo2026streaming,wang2025reconstruction,chen2025long3r}.

To relate the current view to earlier views, a streaming model equips a feed-forward network with an online memory of past frames~\cite{wu2025point3r,liu2025slam3r,cabon2025must3r,li2025wint3r}.
These are the same memories that language models use to carry context.
CUT3R~\cite{wang2025continuous} compresses past observations into a fixed-size \emph{recurrent state} of latent tokens updated at each step, following recurrent networks and state-space models~\cite{hochreiter1997long,gu2024mamba}.
ZipMap~\cite{jin2026zipmap} stores past frames in \emph{fast weights}, layer parameters updated by gradient descent at inference through \emph{test-time training}~\cite{sun2025learning,behrouz2025titans,liu2026mem3r}.
StreamVGGT~\cite{zhuo2026streaming} and STream3R~\cite{lan2026stream3r} append the attention features of each frame to a growing \emph{key-value (KV) cache}, the memory that transformers use to attend to past tokens~\cite{vaswani2017attention,zhang2023h2o,xiao2024efficient}.

Writing a frame into any of these memories carries a cost.
Each update to a recurrent state or fast weights shifts the coordinate frame in which later cameras and points are expressed, so the trajectory \emph{drifts} from the true path.
In a KV cache, each stored frame occupies GPU memory, and the cache of StreamVGGT fills 80\,GB within 10 seconds of video.
Both costs grow with the number of frames written, and the rooms stay the same size.

Training-free rules manage the memory of a frozen network at test time by reading a signal from the current frame to decide how much to write.
In recurrent states, \emph{update gates} scale each update by a \emph{gain} computed from attention weights or state uncertainty~\cite{chen2026ttt3r,zheng2026ttsa3r,dong2026memix,jin2026filt3r}, reliability and information content~\cite{li2026recal3r,ren2026rethinking,baek2026info3r,li2026bat3r}, or motion and geometric cues~\cite{deng2026ssr,wang2026raymap3r,shen2025mut3r,anonymous2026ddd3r}.
In KV caches, compression rules keep the highest-scoring tokens of each frame~\cite{mao2026regvggt,mahdi2025evict3r,su2026xstreamvggt,liu2026streamcachevggt}, keep a fixed number of tokens or frames~\cite{yuan2026infinitevggt,lu2026ovggt,xu2026framevggt,chen2026ghost}, or index kept frames by place~\cite{wang2026stac,zou2026attention,yao2026blazevggt}.
These rules share a premise: a sharper signal of the current frame makes a better memory.
Like a visitor sketching the living room on a fresh page every time they pass the sofa, a camera returning to a known room produces the same signals, so these memories write the room again and grow with the video, or, at fixed capacity, forget the earlier visit (\cref{fig:teaser}).

To test this premise, we remove the signal and keep the amount written, following budget-matched baselines in network pruning and language-model caches~\cite{liu2019rethinking,liu2022unreasonable,wang2026random}.
Our \emph{budget-matched baseline} spends the \emph{update budget} of a gate, the sum of its gains on a video, at uniformly spaced frames with no image signal.
On our long-sequence benchmark, this blind schedule matches nine published gates with a geometric-mean error ratio of 0.99, so their signals add no mean accuracy beyond their budget.
We prove that when the expected coordinate shift per write keeps a direction, the drift of every rule stays within a constant factor of its budget wherever writes fall, and measured drift grows almost in proportion to the budget.
In the fast weights of ZipMap, a sparse schedule cuts the error of writing every frame by nearly an order of magnitude.
The right question for a memory rule is therefore how much to write.
A fixed interval answers it by counting frames, rewriting a revisited room on every visit and leaving views uncovered when the camera moves fast.

We propose \scenebound{}, a training-free rule that lets the scene set the amount, following the \emph{keyframe} principle of visual simultaneous localisation and mapping (SLAM)~\cite{davison2007monoslam,klein2007parallel,forster2014svo,cadena2016past}.
Visual SLAM stores a view as a keyframe when the camera moves far enough from stored ones and culls redundant keyframes so the map grows with the scene~\cite{murartal2015orbslam,engel2018direct,johannsson2013temporally,campos2021orbslam3}.
\scenebound{} writes a frame into the memory of a frozen network as a keyframe when its predicted pose lies at least a fixed radius from every stored keyframe, and decodes closer frames from the unchanged memory.
Because each frame is compared with all stored keyframes, the keyframes lie near the best placement for their count, which stays within a constant factor of the \emph{covering number} of the visited poses, the fewest views keeping a stored view near every camera~\cite{gonzalez1985clustering,dumitrescu2020online}; the count thus grows with the places visited and stops growing on revisits, regardless of video length.
The poses come from the model's own pose head, so the test needs no extra network pass and applies to any memory that can keep or drop a frame.
In large scenes, where one state drifts as the visited region grows, \scenebound{} splits memory into \emph{submaps}, as SLAM does~\cite{bosse2004simultaneous,elvira2019orbslamatlas,maggio2025vggtslam,deng2026vggtlong}, and closes the active submap once its first frame, decoded again by the frozen model, moves beyond a threshold, a check needing no ground truth.

With one radius across CUT3R, ZipMap, StreamVGGT and STream3R, \scenebound{} achieves the lowest mean absolute trajectory error (ATE) among the published training-free rules we evaluate in the recurrent state of CUT3R and the KV cache of StreamVGGT, in one pass and over nine passes of a scene.
Over those repeated passes, its keyframes grow by 3\% to 6\%, published gates multiply their initial budget, and fixed-capacity caches multiply their error.

%% file: sec/2_related.tex
\section{Related work}
\label{sec:related}

\begin{figure*}[t]
\centering
\includegraphics[width=\linewidth]{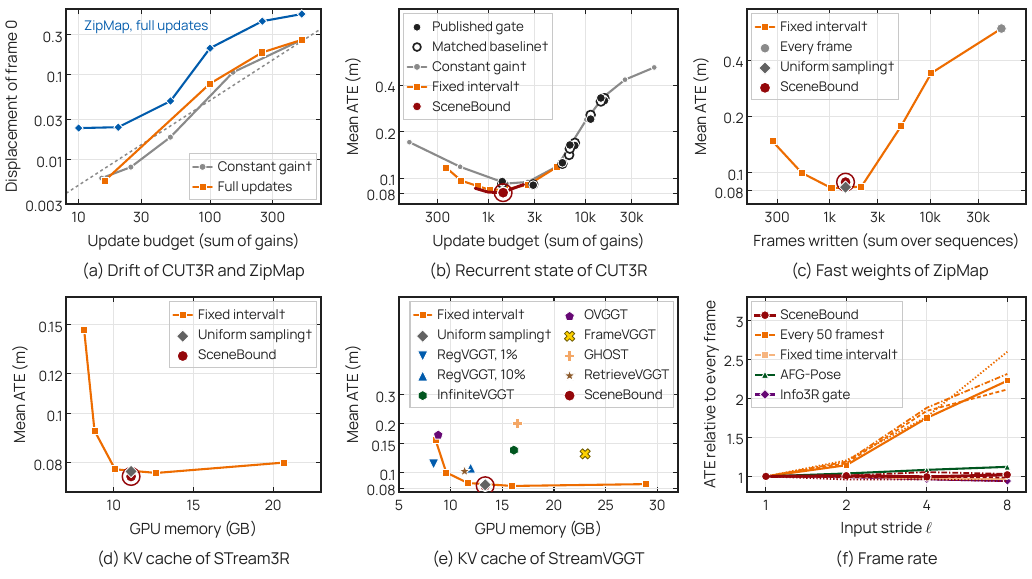}
\caption{The amount written orders the drift and the error in four released models. (a) After 500 frames, geometric mean over the benchmark datasets; the dashed line is proportional to the budget. (b--e) Benchmark means; lines in (b) join each gate to its budget-matched baseline. (f) CUT3R solid, StreamVGGT dashed, STream3R dash-dotted, ZipMap dotted; \dag{} marks constructed baselines.}
\label{fig:budget}
\end{figure*}

\paragraph{Streaming 3D reconstruction.}
Streaming models extend multi-view feed-forward networks~\cite{leroy2024grounding,yang2025fast3r,keetha2026mapanything} with online memories of spatial tokens, keyframes or KV caches~\cite{wu2025point3r,liu2025slam3r,li2025wint3r,lan2026stream3r}.
Models trained for long videos learn what to keep through persistent tokens, hybrid memories or fast weights~\cite{li2026s2tok,cheng2026longstream,zhang2026loger,liu2026mem3r}, and recent ones read their memory through test-time training, geometric context or long-horizon attention~\cite{jin2026zipmap,chen2026lingbotmap,cheng2026horizonstream,han2026revisiting}.
\scenebound{} decides, without training, what enters the memory of a released model.

\paragraph{Training-free memory management.}
Training-free rules filter the updates of a recurrent state~\cite{deng2026ssr,wang2026raymap3r,shen2025mut3r}, evict or merge cache tokens~\cite{anonymous2026diastream,jiang2026motion}, or index memory by place in a state~\cite{zhang2026slot3r} or a cache~\cite{wang2026stac,chen2026ghost,zou2026attention,yao2026blazevggt}.
Adaptive gains have been reported to settle near constant values~\cite{ren2026rethinking,anonymous2026ddd3r}, and gate papers compare with constant gains or fixed intervals at budgets set by hand~\cite{jin2026filt3r,dong2026memix}.
We give a signal-free schedule the budget that each gate logs on each video and let the visited places set the budget.

\paragraph{Keyframes and submaps.}
Classical SLAM bounds its map by keyframe insertion and culling~\cite{forster2014svo,engel2018direct,campos2021orbslam3} and by submaps in large scenes~\cite{bosse2004simultaneous,elvira2019orbslamatlas}, and learned systems optimise keyframes or chunks in a back end that closes loops~\cite{teed2021droidslam,murai2025mast3rslam,maggio2025vggtslam,deng2026vggtlong} across local and global maps~\cite{wang2026amb3rslam,li2026closer}.
Feed-forward and tracking methods select keyframes by pose distance or point overlap against recent or stored views~\cite{wang2026amb3r,lin2026scal3r,taher2026kvtracker,cabon2025must3r}.
\scenebound{} applies the test against all stored keyframes to the memory of a frozen network and bounds it by the covering number of the predicted poses.

%% file: sec/3_method.tex
\section{Method}
\label{sec:method}

\subsection{Preliminaries}
\label{sec:setting}

A frozen streaming model reads frames $x_0,x_1,\dots,x_N$ in order and predicts the camera pose and 3D points of each frame in the coordinate frame of $x_0$.
In CUT3R~\cite{wang2025continuous}, the memory is a state $s_t$ of $n$ tokens initialised from $x_0$, beside a fixed-size pose memory updated at every frame.
At frame $t$, the decoder reads $s_{t-1}$ and $x_t$ to predict the pose, the points and a candidate state $c_t$.
An update rule gives state token $k$ a gain $b_{t,k}\in[0,1]$ scaling the increment $\xi_t$ towards $c_t$,
\begin{equation}
  s_t=s_{t-1}+b_t\odot\xi_t,\qquad \xi_t=c_t-s_{t-1}.
  \label{eq:update}
\end{equation}
Unit gain across the state is a \emph{full update}, applied at every frame in CUT3R; zero gain \emph{skips} the update.
An update gate computes gains from a frame signal, and a \emph{schedule}, such as a constant gain or fixed interval, fixes them beforehand.
With mean frame gain $\beta_t$, a rule has \emph{update budget} $W=\sum_t\beta_t$ and sum of squared gains $Q=\sum_t\frac1n\sum_k b_{t,k}^2$; under full updates, both equal the count of updated frames.
The distance $d$ between two predicted poses is their relative rotation angle in radians plus their camera-centre distance divided by the median predicted depth of $x_0$, so one radius $M$ serves both a desk and a street.

\subsection{The budget determines the drift}
\label{sec:baselines}

To see what a gate signal adds, we first ask how much drift the update budget $W$ causes by itself.
The state increments $\xi_t$ are nearly uncorrelated, so the change that adds up lies in the coordinate frame that $s_t$ encodes, and we measure it through a read-out $\phi$, the pose of $x_0$ decoded again from $s_t$ with no write.
With $B$ the largest increment norm and $L$ the Lipschitz constant of $\phi$, a write of gain $\beta_t$ moves $s_t$ by at most $B\beta_t$ and $\phi$ by at most $LB\beta_t$.
When each write pushes $\phi$ on average along a common direction, the coherent parts add in proportion to $W$, and the unpredictable parts add like a random walk.

\begin{theorem}[Budget bound]
\label{thm:budget}
\input{thm/budget}
\end{theorem}

The budget thus confines expected drift to a factor of $LB/\kappa$, wherever writes fall and whatever signal sets the gains, and a larger budget raises drift without limit.
Comparing a constant gain $W/N$ ($Q=W^2/N$) with full updates ($Q=W$) at equal budget separates the two terms: on 7-Scenes and Neural RGB-D, full updates at our keyframes have 22 times the $Q$ of a constant gain and after 500 frames shift the pose of $x_0$ by 0.92 and 1.11 times as much.
Across five constant gains and three full-update rules, this shift grows with exponent 1.2 in $W$, against 0.5 for increments without a common direction, and stays within a factor of 2.4 per unit budget (\cref{fig:budget}a), as in the fast weights of ZipMap.

Within this band, a signal can lower drift at a fixed budget by writing where the expected change is small or giving small gains where the random part is large.
To test whether published signals do this, our \emph{budget-matched baseline} takes the sums $(W,Q)$ of a gate on a video of $N$ gated frames and applies one gain $g=W/J$ at $J$ uniformly spaced frames, with $J$ the integer in $[\lceil W\rceil,N]$ closest to $W^2/Q$.
Its budget is $W$, and its sum of squared gains $W^2/J$ lies within 0.3\% of $Q$ across all 549 gate-sequence pairs.
Reading no frame, its gap to the gate measures what the signal adds at the gate's chosen budget; we mark constructed baselines with~\dag.

On the 61 benchmark sequences, the baseline matches nine published gates from eight papers (\cref{fig:budget}b): the ratio of its mean ATE to the gate's ranges from 0.92 for AFG-Img to 1.06 for TTSA3R, with a geometric mean of 0.99, and lies within resampling noise for seven of the nine.
This ratio is 1.04 in the second released checkpoint at 224 pixels, 0.98 on held-out Replica and 0.96 under the periodic reset of TTT3R.
Under that reset on full-length KITTI drives~\cite{geiger2012ready} and TUM RGB-D sequences, the signals add accuracy, and the baselines reach a geometric mean of 1.12 and 1.13 times the gate error.

The budget thus sets mean accuracy on the benchmark, and the increments set the form of update: at equal budget, full updates every 10 to 150 frames reach 1\% to 18\% lower ATE than a constant gain (\cref{fig:budget}b), whose increments 8 to 16 frames apart repeat one push at a gain of 0.05, with a mean inner product of 0.46 times their mean squared norm.

\begin{figure*}[t]
\centering
\includegraphics[width=\linewidth]{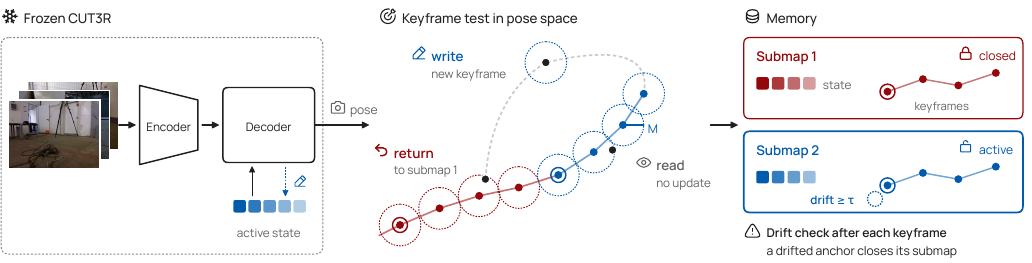}
\caption{\scenebound{} writes a frame at least $M$ from every keyframe, returns to a closed submap within $M$ of its keyframes, reads every other frame with no update, and closes a submap once its anchor drifts by at least a threshold $\tau$.}
\label{fig:method}
\end{figure*}

\subsection{Spatial keyframes}
\label{sec:keyframes}

The decoder localises a frame against stored views of the same place, so a state needs one view near every camera, and by the budget bound further writes add drift.
\scenebound{} therefore lets the visited poses set the budget (\cref{fig:method}).
When the predicted pose of frame $t$ lies at least $M$ from every stored keyframe, we apply a full update, as CUT3R is trained with, and store the pose as a keyframe.
Closer frames skip the update and are decoded from the unchanged state; as for the gates, the pose memory receives every frame.
Because each keyframe lies at least $M$ from earlier ones, no open ball of radius $M/2$ holds two keyframes.

\begin{theorem}[Keyframe count]
\label{thm:keyframes}
\input{thm/keyframes}
\end{theorem}

The covering number $N_d(X,r)$ measures the visited region at resolution $r$ and stays unchanged on revisits up to pose error.
\scenebound{} thus stays online within the factor $\lambda$ of the least budget $N_d(X,M)$ at any length and places keyframes near the best for their count: no choice of one frame fewer covers $X$ at half the radius, the guarantee of greedy $k$-centre clustering~\cite{gonzalez1985clustering}.
Uniform sampling in time has no such guarantee, since a lingering camera draws most frames in one place.
With predicted poses within $\varepsilon<M/2$ of the truth, the bounds hold for the true poses at radii $M+2\varepsilon$ and $M/2-\varepsilon$, and a rigid break in the predicted poses adds at most the covering number of the region seen after it.
Since $d$ is a metric for every fixed depth unit, a first frame $\rho$ times closer refines the counting radius by at most $\rho$.
Since a full update moves $\phi$ by at most $LB$, one state moves $\phi$ by at most $LB\,N_d(X,M/2)$ at any length, and updating every $P$ frames moves it by at least $\kappa\lfloor N/P\rfloor$ in expectation.

This bound requires comparing each frame with all stored keyframes, as in keyframe-based SLAM~\cite{klein2007parallel}; testing the latest keyframe or a recent window writes a revisited place again once its keyframes leave that set.
Predicted poses can index the memory because revisits receive the same pose: on the revisit sequences of \cref{sec:setup} in the KV cache, 99.5\% of frames in later passes lie within $M$ of their first-pass pose under \scenebound{}, against 66.8\% under InfiniteVGGT.
On the benchmark in the KV cache, \scenebound{} stores 1,433 keyframes, with $N_d(X,M)$ between 571 and 965.

\subsection{Submaps from measured drift}
\label{sec:submaps}

\begin{figure*}[t]
\centering
\includegraphics[width=\linewidth]{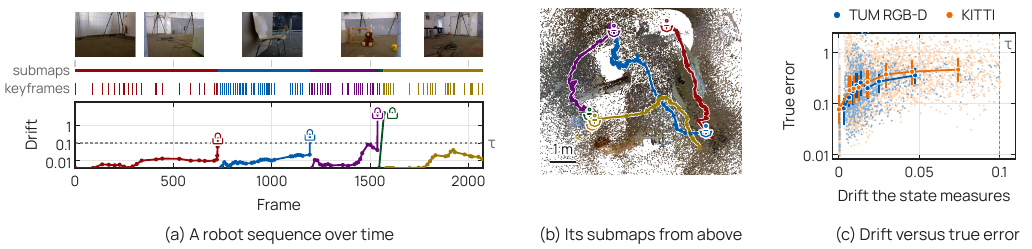}
\caption{The anchor drift that the state measures tracks the true error and closes a submap at $\tau$. (a, b) A robot sequence of TUM RGB-D. (c) TUM RGB-D and KITTI, with medians and quartiles per bin.}
\label{fig:submaps}
\end{figure*}

On a long trajectory, the covering number grows with the visited region, and with it the drift bound of one state.
Periodic resets and the chunks of VGGT-Long~\cite{deng2026vggtlong} close a state at fixed frame counts; \scenebound{} closes it when it measures drift.
A \emph{submap} is a state with its keyframes, and its \emph{anchor} is the frame that opened it; \emph{\scenebound{}, one state} is the single-submap variant.
After each keyframe write, the \emph{drift check} decodes the anchor again with the current state and closes the submap once this pose moves by at least $\tau$ from its initial pose.
The next frame opens a new submap with a fresh state and pose memory, placed in the closed submap's coordinates by decoding the frame once with the closed state, and every submap keeps the depth unit of $x_0$.
Closed submaps stay in memory beside the active one, and a frame at distance below $M$ from a closed submap's keyframe \emph{returns} to it, which resumes as it was left.

For the counts below, consecutive predicted poses lie less than $\delta<M$ apart, and the decoded pose of a fixed frame moves by at most $L$ times a change of the state, so a full update moves the anchor by at most $LB$.

\begin{theorem}[Submap counts]
\label{thm:submaps}
\input{thm/submaps}
\end{theorem}

The covering number thus bounds the keyframes and states in memory, a revisit adds neither, and a room that one state holds without drift never splits.

Because the anchor image is fixed, its pose shift measures state drift without ground truth (\cref{fig:submaps}a, b).
Inside a submap this shift rises with the true pose error, with median rank correlations of 0.74 on TUM RGB-D and 0.89 on KITTI (\cref{fig:submaps}c), and partial rank correlations of 0.27 and 0.02 given the frame count since the anchor.

\subsection{Fast weights and the KV cache}
\label{sec:cache}

ZipMap~\cite{jin2026zipmap} replaces global attention by test-time training: each frame updates the fast weights of every global block by a normalised gradient step and is decoded with them.
After decoding a non-keyframe, we restore the fast weights to their values before it, so each frame applies a full update or a skip, as the budget bound treats.
In the KV caches of StreamVGGT~\cite{zhuo2026streaming} and STream3R~\cite{lan2026stream3r}, every camera is expressed in the coordinate frame of $x_0$, whose keys and values never change.
A written frame leaves the coordinate frame in place and costs memory, and the number of retained frames sets the error (\cref{fig:budget}d, e).
After decoding a non-keyframe, we remove its keys and values from every aggregator layer and the camera head, so the keyframe count bounds the cache by the covering number.

%% file: thm/budget.tex
Let $\phi$ be a read-out of the state, such as the pose decoded for a fixed frame with no write, whose change $\Delta_t=\phi(s_t)-\phi(s_{t-1})$ has norm at most $LB\beta_t$, as it does for one gain per frame.
If, given the earlier frames and the gains of frame $t$, $\Delta_t$ has an expected component of at least $\kappa\beta_t$ along a fixed unit vector $u$, then every update rule, gates included, has
\begin{equation}
  \kappa\,\E[W]\;\le\;\E\|\phi(s_N)-\phi(s_0)\|\;\le\;LB\,\E[W],
  \eqlabel{eq:budget-band}
\end{equation}
and the parts of $\Delta_t$ that the earlier frames and the gains do not predict sum to a vector of mean squared norm at most $(LB)^2\,\E[Q]$.

%% file: thm/keyframes.tex
Let $X$ be the set of predicted poses of the frames of a video, $K$ the number of keyframes that \scenebound{} stores at radius $M$, frame 0 included, and $N_d(X,r)$ the least number of open balls of radius $r$ in the distance $d$, centred anywhere in pose space, that cover $X$.
Every pose of $X$ lies at distance below $M$ from a keyframe stored no later than it, and
\begin{equation}
  N_d(X,M)\le K\le N_d(X,M/2)\le\lambda N_d(X,M),
  \eqlabel{eq:keyframe-count}
\end{equation}
where $\lambda$ balls of radius $M/2$ cover a ball of radius $M$.
Every rule that keeps a stored pose at distance below $M$ from every pose of $X$ stores at least $N_d(X,M)$ frames, and every rule that stores $K-1$ frames leaves a pose of $X$ at distance $M/2$ or more from all of them.
If the frames split into $H$ sets, in each of which one rigid motion carries the true poses $Y_h$ of the set to within $\varepsilon<M/2$ of the predicted ones, then $K\le\sum_{h=1}^{H}N_d(Y_h,M/2-\varepsilon)$.
Fed every $\ell$-th frame, with every pose of $X$ within $\delta_\ell$ of the last pose fed, it stores between $N_d(X,M+\delta_\ell)$ and $N_d(X,M/2)$ keyframes.

%% file: thm/submaps.tex
Let $U$ be the number of frames, other than those that open a submap, whose pose lies at distance $M$ or more from every keyframe when the frame arrives, $C$ the number of submaps and $K$ the number of keyframes.
Every frame is decoded with a submap that holds a keyframe at distance below $M+\delta$ from its pose, and
\begin{equation}
  U\le N_d(X,M/2),\qquad C\le 1+U,\qquad K\le 1+2U.
  \eqlabel{eq:submap-counts}
\end{equation}
A return to a kept submap adds no submap, no keyframe and no placement.
Under the drift check at threshold $\tau$, every submap that the check closes holds more than $\tau/(LB)$ keyframes, and a video on which one state stores at most $\tau/(LB)$ keyframes, in particular every video with $N_d(X,M/2)\le\tau/(LB)$, keeps one state.

%% file: sec/4_experiments.tex
\section{Experiments}
\label{sec:experiments}

\subsection{Experimental setup}
\label{sec:setup}

\input{tab/main}

\paragraph{Implementation.}
Methods in the recurrent state share one evaluator and the released 512-pixel CUT3R checkpoint.
Cache methods and ZipMap run released checkpoints with official streaming settings in the RegVGGT~\cite{mao2026regvggt} evaluator, ZipMap with a one-frame online window.
AFG-Img and AFG-Pose are the image and pose gates of AFG, Info3R in full adds its resets to its gate, TTT3R with reset restarts every 100 frames, and cache memory counts the GPU memory beyond the stored input frames.
Where papers report TUM-dynamics at 1,000 frames, our runs lie within 13\% of their values, OVGGT within 22\%, our implementations of AFG-Pose and Info3R included.
Constructed baselines (\dag) replace the keyframe test of \scenebound{}: a \emph{fixed interval} writes every 50 frames, the best of seven intervals from 10 to 150 frames in the state on the test sequences, \emph{uniform sampling} spreads the count of \scenebound{} evenly over each video, a count known only after the run, and the \emph{latest keyframe} tests the radius against the most recent keyframe alone.
We chose the keyframe radius $M=0.3$ from 0.15, 0.2 and 0.3 on 15 TUM sequences including the eight from TUM-dynamics; a later benchmark sweep from 0.2 to 0.5 is lowest at 0.3 and within 1\% up to 0.4.
We set the depth unit of $d$ on the benchmark and $\tau=0.1$ on seven TUM RGB-D sequences, fixing all three across models and datasets.
Inference is deterministic, so each configuration runs once.

\paragraph{Datasets and metrics.}
Our long-sequence benchmark applies the TTT3R~\cite{chen2026ttt3r} pose protocol (stride 1, up to 1,000 frames) to 61 sequences (51,256 frames) from TUM-dynamics~\cite{sturm2012benchmark}, Bonn~\cite{palazzolo2019refusion}, 7-Scenes~\cite{shotton2013scene} and Neural RGB-D~\cite{azinovic2022neural}.
Revisit sequences play 24 sequences forward and backward nine times, 17 from TUM-dynamics and Neural RGB-D in every memory and seven from 7-Scenes in the state.
Concatenated rooms stream all sequences of six 7-Scenes rooms in turn, returning to each along a new path.
At full length, we evaluate on 19 TUM RGB-D and 11 KITTI odometry sequences~\cite{geiger2012ready}, and once with fixed constants on four held-out sets: eight Replica scenes~\cite{straub2019replica}, 11 EuRoC~\cite{burri2016euroc}, 61 ETH3D SLAM training~\cite{schops2019bad} and 19 further TUM RGB-D sequences.
We report ATE in metres after similarity alignment~\cite{umeyama1991least}, fused-point accuracy, memory and frame rate.

\subsection{Main results}
\label{sec:benchmark}

We ask whether the count set by the visited poses is as accurate in one pass as the best tuned amount, and whether memory and accuracy hold on revisits (\cref{tab:main}).
In one pass, \scenebound{} has the lowest mean ATE in both memories, 9\% below Info3R in full and 18\% below RetrieveVGGT, and ranks first or second in each of the eight dataset columns.
On TUM-dynamics, where we chose $M$, both its values lie below every value reported at 1,000 frames, including that of the trained LoGeR~\cite{zhang2026loger}.
Every 50 frames reaches 0.084 in the state, and the best interval in the cache, every 25 frames, matches \scenebound{} with 2.7\,GB more memory (\cref{fig:budget}b, e); outside TUM-dynamics, \scenebound{} lies within 2\% of every 50 frames in the state.
\scenebound{} thus finds online the amount that a sweep of intervals tunes on the test sequences.
Placement also matters: at the same count, uniform sampling and random frames raise the state's error by 12\% and 13\%.

Over nine passes, the keyframes of \scenebound{} grow by 3\% in the state and 6\% in the cache, and every other rule pays in update budget, memory or accuracy.
In the state, published rules end at 1.5 to 9.7 times its error, and the closest, the Info3R gate, spends 11 times its budget.
In the cache, the fixed-capacity InfiniteVGGT and GHOST forget the room, their error more than doubling after the first pass; FrameVGGT takes over twice the memory of \scenebound{}, and OVGGT holds less at 2.4 times its error.

On the concatenated rooms, where new paths bring new poses, \scenebound{} reaches 0.139 against 0.142 for AFG-Pose and every 50 frames.
On the Bonn sequences where a person moves boxes, it reaches 0.085 against 0.094 for the Info3R gate, since each frame's points come from its own image.
The count set by the visited poses thus matches the best tuned amount in one pass, and \scenebound{} alone holds both its cost and its accuracy over revisits.

\subsection{Four released models and four frame rates}
\label{sec:models}

We test whether one radius transfers across memory types and camera speeds.
In ZipMap, \scenebound{} writes 24 frames per sequence and cuts the error from 0.590 to 0.089 (\cref{fig:budget}c); every 50 frames and uniform sampling at its count reach 0.083 and 0.084, so the count brings the gain.
In STream3R, \scenebound{} reaches 0.075, ahead of every fixed interval and of uniform sampling at its count (\cref{fig:budget}d), and in both models its writes grow by at most 1\% over nine passes.

To vary camera speed along the same paths, we feed every second, fourth and eighth frame of the benchmark, scaling the motion per frame by $\ell$ (\cref{fig:budget}f).
At stride 8, \scenebound{} writes about 15\% fewer frames in every model and changes its error by at most 6\%; every 50 input frames more than doubles its error, and a fixed time interval stays within 5\% of \scenebound{} and writes again on revisits.
A first frame facing a far scene raises the error by 3\%, and one facing a near wall, at half or a quarter of the depth unit, writes more frames and raises the error by 11\% and 20\%, as the budget bound predicts at larger budgets.
A unit that rises with the median depth of later keyframes, never falling, keeps the count bound and brings these to 2\% and 3\%.
One radius thus serves three memory types and an eightfold range of camera speeds.

\subsection{Long sequences and held-out data}
\label{sec:heldout}

\begin{figure*}[t]
\centering
\includegraphics[width=\linewidth]{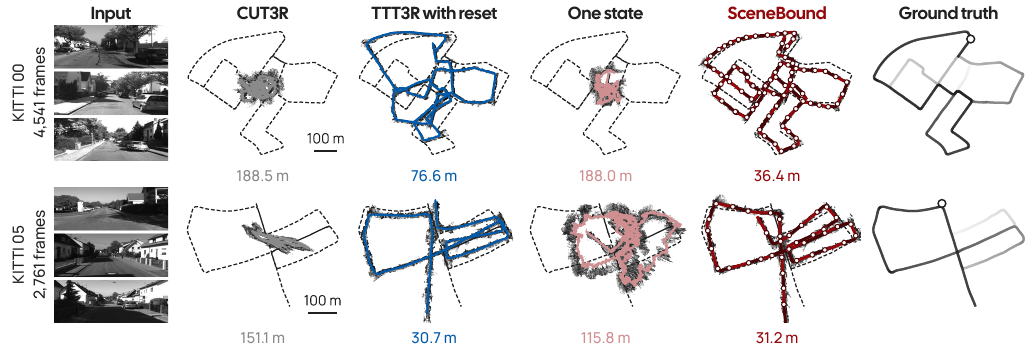}
\caption{Two KITTI drives in the recurrent state, from above, with the true path dashed and the ATE under each map. Rings mark the anchors of \scenebound{}; its submaps alternate in shade.}
\label{fig:drives}
\end{figure*}

We test whether the drift check bounds the drift of one state on long trajectories and held-out data.
Submaps cut the error of one state by 29\% on the TUM RGB-D sequences not used for $\tau$ and by 71\% on the KITTI drives (\cref{fig:drives}), where \scenebound{} reaches 50.7\,m against 61.7\,m for TTT3R with reset.
On these drives, closing the state every 30 frames is lower, and Info3R with constants set for driving, AFG-Pose and the trained LoGeR report 30.5, 44.0 and 25.4\,m; on TUM, Info3R in full and TTT3R with reset stay about 11\% lower.
Aligned to the truth one by one, the KITTI submaps have an error of 4.2\,m, so most of the error lies in placing the submaps, the task of a back end with its own place recognition and geometric verification, as in VGGT-Long~\cite{deng2026vggtlong}.
\scenebound{} runs at 22 frames per second, and its keyframe test against 10,000 stored keyframes takes 0.2\,ms on one CPU core.

On held-out ETH3D and TUM, \scenebound{} has the lowest error, 6\% and 30\% below the next published rules.
On EuRoC, the drift check splits fast drone flights into up to 52 submaps and raises the error of one state from 1.78 to 2.14, behind AFG-Pose at 1.25 and both forms of Info3R; aligned one by one, the submaps are more accurate than one state over the same frames, so placing them causes the excess.
On Replica, where the error of gates grows with their budget, \scenebound{} reaches 0.181, behind both forms of Info3R, the gate at 0.152, as the covering number at $M$ exceeds the best budget.
The drift check thus cuts the drift of one state on long trajectories and held-out cameras.

\subsection{Point clouds}
\label{sec:points}

To measure surface accuracy, we fuse the points of the first 1,000 frames of the 7-Scenes and Neural RGB-D benchmark sequences with the CUT3R evaluator.
In the state, \scenebound{} reaches accuracy errors of 0.026 and 0.086\,m, TTT3R with reset has 3.3 and 1.6 times these, and every 50 frames is lowest, about 10\% below them.
Fed every eighth frame, \scenebound{} is lowest on both, and every 50 frames has 4.2 and 1.2 times its error.
In the cache, the rules that keep whole frames lead on 7-Scenes, and on Neural RGB-D RetrieveVGGT and OVGGT, which read tokens of every earlier frame, have 21\% to 25\% less error than \scenebound{}.

\subsection{Ablation study}
\label{sec:ablations}

\input{tab/ablation}

\Cref{tab:ablation} tests whether a keyframe rule should measure poses or points and check all stored keyframes or recent ones.
Beside the latest keyframe, we port three tests from their official code at the count of \scenebound{} without their trained components: AMB3R~\cite{wang2026amb3r} on the rotation and centre distance to its 10 active keyframes, and the point tests of Scal3R~\cite{lin2026scal3r} on the first and last four writes and MUSt3R~\cite{cabon2025must3r} on all writes.

What the test measures matters on moving content.
In the state, the ported tests lie 7\% to 10\% above \scenebound{}, losing mostly on TUM-dynamics: the point tests write where people walk, 22 frames of a static camera against one for poses, and 12 to 14 of one half-sphere sweep against 26, with four times the error.
In the cache, all rules at this count lie within 2\%, and MUSt3R runs at 4.8 frames per second against 6.3 for \scenebound{}.

The test scope decides the writes on revisits.
On the benchmark, the latest keyframe writes 14\% more frames on returns within a pass, within 2\% in error.
On the revisit sequences, it has 2.5 times the error of \scenebound{}; only MUSt3R and \scenebound{} check all stored views and keep their growth small.
In the cache, MUSt3R ends at 0.062 against 0.069 through one video where predicted poses jump up to nine radii between frames; holding writes to steps below $M$, as the submap counts assume, closes this gap.
Checking all stored views thus keeps the writes flat on revisits, and measuring poses writes less on moving content.

%% file: tab/main.tex
\begin{table*}[t]
\caption{Published training-free rules and \scenebound{} in one pass of the long-sequence benchmark and over nine passes of the revisit sequences, with the cost after them: the update budget in the state and the GPU memory in the cache.}
\label{tab:main}
\centering
\footnotesize
\begin{fulltabular}{@{}lccccccr@{\hspace{3.5pt}}}
\toprule
 & \multicolumn{5}{c}{One pass of the benchmark, ATE (m) $\downarrow$} & \multicolumn{2}{c}{Revisits, nine passes} \\
\cmidrule(lr){2-6}\cmidrule(l){7-8}
Method & TUM-dynamics & Bonn & 7-Scenes & Neural RGB-D & Mean & ATE (m) $\downarrow$ & Cost $\downarrow$ \\
\midrule
\multicolumn{7}{@{}l}{\emph{Recurrent state of CUT3R}} & \emph{budget} \\
CUT3R~\cite{wang2025continuous} & 0.161 & 0.322 & 0.655 & 1.156 & 0.522 & 0.815 & 196,518 \\
MeMix~\cite{dong2026memix} & 0.102 & 0.264 & 0.393 & 0.604 & 0.331 & 0.673 & 54,398 \\
TTT3R~\cite{chen2026ttt3r} & 0.115 & 0.269 & 0.410 & 0.457 & 0.318 & 0.672 & 59,122 \\
TTSA3R~\cite{zheng2026ttsa3r} & 0.093 & 0.220 & 0.315 & 0.289 & 0.241 & 0.653 & 42,908 \\
RayMap3R~\cite{wang2026raymap3r} & 0.131 & 0.127 & 0.206 & 0.257 & 0.170 & 0.700 & 33,413 \\
BAT3R~\cite{li2026bat3r} & 0.069 & 0.155 & 0.170 & 0.262 & 0.164 & 0.608 & 22,241 \\
FILT3R~\cite{jin2026filt3r} & 0.074 & 0.143 & 0.190 & 0.245 & 0.163 & 0.601 & 29,939 \\
AFG-Img\textsuperscript{\ddag}~\cite{ren2026rethinking} & 0.083 & 0.165 & 0.163 & 0.162 & 0.153 & 0.516 & 23,389 \\
ReCal3R~\cite{li2026recal3r} & 0.067 & 0.110 & 0.146 & 0.185 & 0.126 & 0.552 & 22,129 \\
AFG-Pose\textsuperscript{\ddag}~\cite{ren2026rethinking} & \underline{0.055} & 0.107 & \textbf{0.094} & 0.098 & 0.095 & 0.134 & \underline{5,304} \\
Info3R\textsuperscript{\ddag}~\cite{baek2026info3r}, gate & 0.060 & 0.094 & 0.101 & \textbf{0.085} & 0.090 & \underline{0.124} & 7,642 \\
Info3R\textsuperscript{\ddag}, in full & 0.056 & \underline{0.091} & \underline{0.095} & 0.099 & \underline{0.089} & 0.252 & 7,618 \\
TTT3R with reset & 0.096 & 0.115 & 0.163 & 0.167 & 0.134 & 0.573 & 68,399 \\
\oursrow \scenebound{} (ours) & \textbf{0.048} & \textbf{0.078} & \underline{0.095} & \underline{0.088} & \textbf{0.080} & \textbf{0.084} & \oursend[3.5pt] \textbf{698} \\
\midrule
\multicolumn{7}{@{}l}{\emph{KV cache of StreamVGGT}} & \emph{GB} \\
RegVGGT~\cite{mao2026regvggt}, 1\% of keys & 0.058 & 0.099 & 0.116 & 0.195 & 0.113 & 0.129 & 15.2 \\
RegVGGT, 10\% of keys & 0.051 & \underline{0.091} & \underline{0.113} & 0.188 & 0.107 & 0.138 & 52.6 \\
InfiniteVGGT~\cite{yuan2026infinitevggt} & 0.067 & 0.129 & 0.161 & 0.178 & 0.137 & 0.335 & 18.1 \\
OVGGT~\cite{lu2026ovggt} & 0.071 & 0.192 & 0.167 & 0.203 & 0.170 & 0.166 & \textbf{8.9} \\
GHOST~\cite{chen2026ghost} & 0.072 & 0.173 & 0.248 & 0.302 & 0.201 & 0.768 & 18.6 \\
FrameVGGT~\cite{xu2026framevggt} & \underline{0.047} & 0.139 & 0.153 & 0.133 & 0.130 & \underline{0.077} & 33.1 \\
XStreamVGGT~\cite{su2026xstreamvggt} & 0.065 & 0.224 & 0.287 & 0.316 & 0.235 & 0.197 & 27.2 \\
RetrieveVGGT~\cite{zou2026attention} & 0.050 & 0.101 & 0.126 & \textbf{0.102} & \underline{0.102} & 0.080\rlap{$^{\S}$} & 15.6\rlap{$^{\S}$} \\
\oursrow \scenebound{} (ours) & \textbf{0.033} & \textbf{0.071} & \textbf{0.104} & \underline{0.122} & \textbf{0.083} & \textbf{0.069} & \oursend[3.5pt] \underline{14.8} \\
\bottomrule
\end{fulltabular}

\smallskip
\raggedright\scriptsize
$^{\ddag}$Our implementation of a method without released code. $^{\S}$With its pose-aware memory on, which holds 211\,GB of host memory.
\end{table*}

%% file: tab/ablation.tex
\begin{table}[t]
\caption{Write rules on the 61 benchmark sequences in one state and in the KV cache, and the growth of their writes over nine passes of the revisit sequences.}
\label{tab:ablation}
\centering
\footnotesize
\begin{fulltabular}{@{}llccccc@{}}
\toprule
 & & & \multicolumn{2}{c}{ATE (m) $\downarrow$} & \multicolumn{2}{c}{Growth $\downarrow$} \\
\cmidrule(lr){4-5}\cmidrule(l){6-7}
Write rule & Test & Writes & State & Cache & State & Cache \\
\midrule
\multicolumn{7}{@{}l}{\emph{No test}} \\
Every 50 frames\textsuperscript{\dag} & none & 1,039 & 0.0839 & 0.0863 & 8.93 & 8.89 \\
Uniform sampling\textsuperscript{\dag} & none & 1,421 & 0.0904 & 0.0841 & 6.39 & 6.34 \\
\multicolumn{7}{@{}l}{\emph{Test against recent views}} \\
Latest keyframe\textsuperscript{\dag} & pose & 1,616 & \underline{0.0820} & 0.0846 & 8.44 & 7.95\rlap{$^{*}$} \\
AMB3R~\cite{wang2026amb3r} & pose & 1,422 & 0.0875 & \textbf{0.0825} & 6.54 & 5.41\rlap{$^{*}$} \\
Scal3R~\cite{lin2026scal3r} & points & 1,417 & 0.0864 & 0.0845 & 5.52 & 2.80\rlap{$^{*}$} \\
\multicolumn{7}{@{}l}{\emph{Test against all stored views}} \\
MUSt3R~\cite{cabon2025must3r} & points & 1,415 & 0.0887 & \underline{0.0831} & \underline{1.16} & \underline{1.12} \\
\oursrow \scenebound{} (ours) & pose & 1,421 & \textbf{0.0805} & \underline{0.0831} & \textbf{1.03} & \oursend \textbf{1.06} \\
\bottomrule
\end{fulltabular}

\smallskip
\raggedright\scriptsize
$^{\dag}$Constructed baseline. $^{*}$Five, nine and 15 of the 17 videos fit in GPU memory.
\end{table}

%% file: sec/5_conclusion.tex
\section{Conclusion}
\label{sec:conclusion}

The update budget of a training-free memory rule fixes its drift up to a constant factor wherever the writes fall and sets its mean trajectory error on the long-sequence benchmark, so a rule must first decide how much to write.
\scenebound{} lets the visited poses decide, placing keyframes near the best for their count, so the amount it writes follows the scene and stops growing on revisits, with one radius across recurrent states, fast weights and KV caches in four released models and an eightfold range of camera speeds.
Its drift check cuts the drift of one state on long trajectories, where the remaining error lies in placing submaps, a task for a back end with independent loop verification, and scenes that change call for a test that retires stored views.

%% file: sec/6_llm.tex
\section*{LLM Usage Statement}
\label{sec:llm}

We used Claude Code (Anthropic), Codex (OpenAI) and Gemini (Google) to assist our research and its execution: implementing and running the experiments and the figures, searching the literature, and preparing and editing the text.
The research questions, the theory, the design of the method and the division of the work are the authors' own.
The authors checked every result, proof and citation and take full responsibility for the content.

%% file: supp/A_proofs.tex
\section{Proofs}
\label{app:proofs}

\subsection{Proofs of the budget bound}
\label{app:proof-budget}

Let $\mathcal F_t$ hold the frames before $t$, frame 0 included, their states, increments and gains, and the gains $b_t$ of frame $t$, so that $\mathcal F_{t'}\subseteq\mathcal F_t$ for $t'<t$.
Write $\Delta_t=\phi(s_t)-\phi(s_{t-1})$ for the change of the read-out at frame $t$, $m_t=\E[\Delta_t\mid\mathcal F_t]$ for its prediction from the past and $\eta_t=\Delta_t-m_t$ for the rest, so that $\phi(s_N)-\phi(s_0)=\sum_tm_t+\sum_t\eta_t$.
The conditions of \cref{thm:budget} read $\|\Delta_t\|\le LB\beta_t$ and $\langle u,m_t\rangle\ge\kappa\beta_t$. With one gain per frame the first follows from $\|\xi_t\|\le B$ and a read-out that moves by at most $L$ times a change of the state, since then $\|s_t-s_{t-1}\|=\beta_t\|\xi_t\|\le B\beta_t$; with gains per token it is a condition, since $\|b_t\odot\xi_t\|$ is bounded only by $B\max_kb_{t,k}$.

\begin{restated}{thm:budget}{Budget bound}
\input{thm/budget}
\end{restated}

\begin{proof}
On every video, $\|\phi(s_N)-\phi(s_0)\|\le\sum_t\|\Delta_t\|\le LB\sum_t\beta_t=LB\,W$, which gives the upper bound.
For the lower bound, $\E\langle u,\Delta_t\rangle=\E\langle u,m_t\rangle\ge\kappa\,\E\beta_t$, and summing over the frames gives $\E\langle u,\phi(s_N)-\phi(s_0)\rangle\ge\kappa\,\E W$; since $u$ is a unit vector, $\langle u,v\rangle\le\|v\|$ for every $v$.
For $t'<t$ the rest $\eta_{t'}$ is fixed by $\mathcal F_t$, so $\E\langle\eta_{t'},\eta_t\rangle=\E\langle\eta_{t'},\E[\eta_t\mid\mathcal F_t]\rangle=0$, and $\E\|\eta_t\|^2\le\E\|\Delta_t\|^2\le(LB)^2\E\beta_t^2$.
Summing over pairs of frames gives $\E\|\sum_t\eta_t\|^2\le(LB)^2\E\sum_t\beta_t^2\le(LB)^2\E[Q]$, since $\beta_t^2\le\frac1n\sum_kb_{t,k}^2$ by Jensen's inequality, with equality for one gain per frame.
\end{proof}

The bound needs no model of the state, and its two constants can be measured: $\kappa$ is the least expected change of the read-out per unit gain along the direction of drift, and $LB$ the largest change.
The condition is stated given the gain of the frame, so it covers gates, whose gains depend on the frame they scale.
Read as a learning rate, as TTT3R reads its gate, $W$ and $Q$ are the two sums of the conditions of stochastic approximation~\cite{robbins1951stochastic}.

An identity for a constant push is the special case in which the read-out is the state itself.

\begin{proposition}[Constant push]
\label{prop:push}
Let the increments, of norm at most $B$, split as $\xi_t=\mu+\zeta_t$, with a push $\mu\ne0$ shared by all frames and rules and a random part $\zeta_t$ each of whose $n$ tokens has, given $\mathcal F_t$, zero mean and mean squared norm $\omega/n$.
Every update rule moves the state by $s_N-s_0=\sum_tb_t\odot\mu+R$ with $R=\sum_tb_t\odot\zeta_t$ and $\E\|R\|^2=\omega\,\E[Q]$.
A schedule with one gain per frame has $\E\|s_N-s_0\|^2=\|\mu\|^2W^2+\omega Q$ for every placement of its gains, and every rule with one gain per frame has $\E\|s_N-s_0\|^2\ge\|\mu\|^2(\E W)^2$.
\end{proposition}

\begin{proof}
The recursion gives $s_N-s_0=\sum_tb_t\odot\xi_t=\sum_tb_t\odot\mu+R$, and with one gain $\beta_t$ per frame $\sum_tb_t\odot\mu=W\mu$.
For $t'<t$ the gains $b_{t'}$, $b_t$ and the increment $\zeta_{t'}$ are fixed by $\mathcal F_t$, so
\begin{equation}
  \E\bigl[\langle b_{t'}\odot\zeta_{t'},\,b_t\odot\zeta_t\rangle\bigr]=\E\Bigl[\sum_k b_{t',k}b_{t,k}\bigl\langle\zeta_{t',k},\E[\zeta_{t,k}\mid\mathcal F_t]\bigr\rangle\Bigr]=0,
\end{equation}
and $\E\|b_t\odot\zeta_t\|^2=\E\sum_k b_{t,k}^2\,\E[\|\zeta_{t,k}\|^2\mid\mathcal F_t]=\omega\,\E\bigl[\tfrac1n\sum_k b_{t,k}^2\bigr]$.
Summing over pairs of frames gives $\E\|R\|^2=\omega\,\E[Q]$; every expectation is finite since $\|\zeta_t\|\le\|\xi_t\|+\|\mu\|\le2B$.
A schedule fixes its gains before the video, so $W$ and $Q$ are constants and $\E R=0$, which gives $\E\|W\mu+R\|^2=\|\mu\|^2W^2+\omega Q$.
For gains computed from the video, $b_t$ is fixed by $\mathcal F_t$, so $\E R=0$ and $\E[s_N-s_0]=\mu\,\E W$ for one gain per frame; the second moment of a vector is at least the squared norm of its mean.
\end{proof}

With the read-out $\phi(s)=s$ and one gain per frame, the conditions of \cref{thm:budget} hold with $u=\mu/\|\mu\|$, $\kappa=\|\mu\|$ and $L=1$, and for schedules the proposition sharpens the band to an identity.
A gate that sets small gains where $\|\zeta_t\|$ is large has $\E[b_{t,k}^2\|\zeta_{t,k}\|^2]<\tfrac{\omega}{n}\E[b_{t,k}^2]$, which is the way the identity leaves for a signal to lower the drift at fixed sums.
In the state of CUT3R this push is small and the read-out moves coherently (\cref{app:increments}), which is why \cref{thm:budget} bounds a read-out of the state and not the state itself.

\subsection{Proofs of the keyframe count}
\label{app:proof-keyframes}

Let $(\mathcal X,d)$ be a pseudometric space of views and $X=\{p_0,\dots,p_N\}$ the predicted poses of the frames, of which the rule accepts $p_0$.
$N_d(X,r)$ is the least number of open balls of radius $r$ with centres in $\mathcal X$ that cover $X$, and $P_d(X,r)$ is the largest size of a subset of $X$ with pairwise distances at least $r$.
The rule accepts $p_t$ when $d(p_t,q)\ge M$ for every accepted $q$.

\begin{restated}{thm:keyframes}{Keyframe count}
\input{thm/keyframes}
\end{restated}

\begin{proof}
The later of two accepted views passed the test against the earlier one, so accepted views are pairwise at distance at least $M$.
An accepted view is at distance zero from itself and a rejected one failed the test against an accepted view, which gives the guarantee.
The balls of radius $M$ around the accepted views therefore cover $X$, so $N_d(X,M)\le K$, and the accepted views form a subset with pairwise distances at least $M$, so $K\le P_d(X,M)$.
Two points of one ball of radius $M/2$ lie at distance below $M$, so such a subset has at most one point in each ball of a cover at radius $M/2$, and $P_d(X,M)\le N_d(X,M/2)$.
Every rule that keeps a stored pose at distance below $M$ from every pose of $X$ supplies a cover of $X$ by balls of radius $M$ around its stored poses, and so stores at least $N_d(X,M)$ frames.
If a rule stored $K-1$ frames within distance below $M/2$ of every keyframe, two keyframes would lie within $M/2$ of one stored frame and so at distance below $M$ from each other, which the test excludes; hence one keyframe, a pose of $X$, lies at distance $M/2$ or more from all $K-1$ stored frames.
For the split into $H$ sets, let $A_h$ be the rigid motion of set $h$; since $d$ is invariant under a common rigid motion of both poses, two keyframes $a$, $b$ of set $h$ have $d(y_a,y_b)=d(A_hy_a,A_hy_b)\ge d(p_a,p_b)-2\varepsilon\ge M-2\varepsilon$, so the packing argument at radius $M-2\varepsilon$ bounds the keyframes of set $h$ by $N_d(Y_h,M/2-\varepsilon)$, and summing over the sets gives the bound.
A similarity of scale $\sigma_h\ne1$ changes the centre term of $d$ by $\sigma_h$, which is a change of the depth unit, and by \cref{prop:unit} the term of set $h$ becomes $N_d(Y_h,(M/2-\varepsilon)/\hat\sigma_h)$ with $\hat\sigma_h=\max(\sigma_h,1/\sigma_h)$.
A cover of $X$ by $N_d(X,M)$ balls of radius $M$, each covered by $\lambda$ balls of radius $M/2$, covers $X$ at radius $M/2$, so $N_d(X,M/2)\le\lambda N_d(X,M)$.
Fed every $\ell$-th frame, the rule sees the poses $X_\ell\subseteq X$, so its count is at most $N_d(X_\ell,M/2)\le N_d(X,M/2)$, and its keyframes cover $X_\ell$ at radius $M$; every pose of $X$ lies within $\delta_\ell$ of a pose of $X_\ell$, so the same keyframes cover $X$ at radius $M+\delta_\ell$, and the count is at least $N_d(X,M+\delta_\ell)$.
\end{proof}

The statement carries over to the true poses $Y=\{y_0,\dots,y_N\}$ when the predicted poses are accurate to below half the radius.

\begin{proposition}[Keyframe count on true poses]
\label{prop:true-poses}
Let $Y=\{y_0,\dots,y_N\}$ be the true poses carried into the coordinates of the predictions by one similarity. If $d(p_t,y_t)\le\varepsilon<M/2$ for every frame, every true pose lies at distance below $M+2\varepsilon$ from the true pose of a keyframe stored no later than it, and $N_d(Y,M+2\varepsilon)\le K\le N_d(Y,M/2-\varepsilon)$.
\end{proposition}

\begin{proof}
The triangle inequality gives $d(y_t,y_q)<M+2\varepsilon$ for a frame and the accepted frame $q$ that covers it, so the balls of radius $M+2\varepsilon$ around the true poses of the keyframes cover $Y$, and $d(y_i,y_j)\ge M-2\varepsilon$ for two accepted frames, so the packing argument at radius $M-2\varepsilon$ gives $K\le N_d(Y,M/2-\varepsilon)$.
\end{proof}

The bounding quantities are functions of the set $X$, which is invariant to the order of the frames and to repeated visits, and a lower frame rate enters them through the single quantity $\delta_\ell$.
The second statement is the guarantee of greedy $k$-centre clustering~\cite{gonzalez1985clustering}, met online, since at its own count \scenebound{} covers $X$ at a radius within a factor 2 of the radius that any choice of one frame fewer achieves.
The split into sets bounds the cost of a break in the predicted poses, such as a jump of the camera head after which later poses are consistent among themselves and rigidly shifted from the earlier ones. Each break adds at most one covering number of the region seen after it, and a break that also rescales the later poses adds the factor of \cref{prop:unit}, never a cost that grows with the length of the video.
With $H=1$ the bound is that of \cref{prop:true-poses}.
For online covering by unit balls in a Euclidean space of dimension $D$, every deterministic rule has a competitive ratio of at least $D+1$ against the offline optimum~\cite{dumitrescu2020online}, so a constant factor is the best order an online rule achieves.
The distance of the method, $d(p_i,p_j)=\theta(R_i,R_j)+\|o_i-o_j\|/z_0$ with the rotation angle $\theta$, the camera centres $o_i$ and the fixed depth unit $z_0$, is a metric on poses for every fixed $z_0>0$, so all statements hold whatever depth the first frame sees.

\begin{proposition}[Depth unit]
\label{prop:unit}
Let \scenebound{} use the depth unit $\rho z_0$ for some $\rho>0$, and let $\hat\rho=\max(\rho,1/\rho)$.
Its keyframe count lies between $N_d(X,\hat\rho M)$ and $N_d(X,M/(2\hat\rho))$, with $d$ the distance at unit $z_0$, and \cref{thm:keyframes,thm:submaps} hold with the distance at unit $\rho z_0$.
\end{proposition}

\begin{proof}
Write $d'$ for the distance at unit $\rho z_0$.
For $\rho\ge1$, $d'\le d\le\rho\,d'$, and for $\rho<1$, $d\le d'\le d/\rho$; in both cases $d/\hat\rho\le d'\le\hat\rho\,d$.
A ball of radius $r/\hat\rho$ in $d$ lies inside the ball of radius $r$ in $d'$ with the same centre, so $N_{d'}(X,r)\le N_d(X,r/\hat\rho)$, and in the same way $N_{d'}(X,r)\ge N_d(X,\hat\rho r)$.
\Cref{thm:keyframes} at radius $M$ in $d'$ gives $N_{d'}(X,M)\le K\le N_{d'}(X,M/2)$, and the two inequalities give the stated range.
\end{proof}

A first frame that faces a wall four times closer than the rest of the scene thus counts the scene at a radius at most four times finer, and the count stays independent of the length of the video and of repeated visits.
A unit that changes along the video and never falls below $z_0$ keeps the upper bound $K\le N_d(X,M/2)$ at unit $z_0$, since a keyframe written at a unit $z\ge z_0$ lies at distance at least $M$ at unit $z$ from every earlier keyframe, and so at least $M$ at unit $z_0$.

\subsection{Drift at every length}
\label{app:proof-drift}

\Cref{thm:budget,thm:keyframes} together bound the displacement of one state at every length.

\begin{proposition}[Drift at every length]
\label{prop:drift}
\input{thm/drift}
\end{proposition}

\begin{proof}
With $\bar\beta_t$ the largest gain of frame $t$, $\|b_t\odot\xi_t\|^2=\sum_k b_{t,k}^2\|\xi_{t,k}\|^2\le\bar\beta_t^2\|\xi_t\|^2$, so $\|s_N-s_0\|\le\sum_t\bar\beta_t\|\xi_t\|\le B\sum_t\bar\beta_t$ for every choice of gains, gains chosen from the video included.
\scenebound{} in one state performs $K-1$ full updates after frame 0, so $\sum_t\bar\beta_t=K-1<N_d(X,M/2)$ by \cref{thm:keyframes}, which gives the first bound.
By the assumptions the pose that the decoder returns for a fixed frame moves by at most $L$ times the change of the state, so by at most $LB\,N_d(X,M/2)$.
A full update every $p$ frames has $W=\lfloor N/p\rfloor$ on every video, and \cref{thm:budget} gives $\E\|\phi(s_N)-\phi(s_0)\|\ge\kappa\lfloor N/p\rfloor$.
A rule of full updates that keeps frame 0 or a written frame at distance below $M$ from every pose of $X$ places the centres of open balls of radius $M$ that cover $X$ at frame 0 and at its writes, so $W\ge N_d(X,M)-1$ on every video, and \cref{thm:budget} gives $\E\|\phi(s_N)-\phi(s_0)\|\ge\kappa\,\E W\ge\kappa(\E N_d(X,M)-1)$.
For \scenebound{}, $\|\phi(s_N)-\phi(s_0)\|\le LB(K-1)\le LB\,N_d(X,M/2)\le\lambda LB\,N_d(X,M)$ by \cref{thm:keyframes}.
\end{proof}

The first bound uses the triangle inequality alone, so it holds for every choice of gains.

\subsection{Proofs of the submap counts}
\label{app:proof-submaps}

As in \cref{sec:submaps}, consecutive predicted poses lie less than $\delta<M$ apart, every increment has norm at most $B$, and the pose decoded for a fixed frame moves by at most $L$ times a change of the state, with $LB>0$ and $\tau>0$.
\scenebound{} with keyframe radius $M$ treats the view $p_t$ of frame $t$ by the first of three cases that applies.
\begin{enumerate}
  \item \emph{Read.} A view written in the current submap lies at distance below $M$ from $p_t$. The frame is read against the current submap.
  \item \emph{Return.} A view written in another submap lies at distance below $M$ from $p_t$. The frame is read against the current submap, and the submap that holds the nearest such view becomes current at frame $t+1$, with its state and its keyframes as they were when it was left.
  \item \emph{Write.} The view $p_t$ is written into the current submap. A new submap opens at frame $t+1$ when, after this write, the anchor read against the state lies at distance $\tau$ or more from its opening pose.
\end{enumerate}
A submap that opens at frame $t+1$ has the anchor $p_{t+1}$, written by a full update, and is placed by reading $p_{t+1}$ against the submap it replaces.
The anchor is exempt from the return test, and views of different submaps are compared in the frame of the first submap, into which each submap is carried by the composition of the placements that chain it to the first.
Frames of case~3 are the \emph{uncovered} frames, and $U$ is their number.
Under submaps, $X$ is the set of predicted poses carried into the frame of the first submap in this way, and the \emph{opening pose} of an anchor is the pose that its own submap gives it right after it is written.

\begin{restated}{thm:submaps}{Submap counts}
\input{thm/submaps}
\end{restated}

\begin{proof}
\emph{Reads.} A frame of case~1 has a keyframe of its submap below $M$.
A frame $t$ of case~2 or~3 is read against the current submap before anything is written, and for it consider frame $t-1$.
If frame $t-1$ was read by case~1, or written by case~3 or as an anchor in the submap that is current at frame $t$, a keyframe $q$ of that submap has $d(p_{t-1},q)<M$, so $d(p_t,q)<M+\delta$.
If frame $t-1$ was a return, the submap current at frame $t$ holds a view $q$ with $d(p_{t-1},q)<M$, and the same bound holds.
Frame $t-1$ is never a write that closed its submap, since frame $t$ would then be an anchor, and an anchor $p_{t+1}$ is placed by a read against the submap it replaces, which holds $p_t$, within $\delta$ of the anchor.

\emph{Counts.} An uncovered view failed cases~1 and~2, so it lies at distance at least $M$ from every view written before it in any submap, and it is written itself.
The uncovered views are therefore pairwise at distance at least $M$, and the packing argument of \cref{thm:keyframes} gives $U\le P_d(X,M)\le N_d(X,M/2)$.
Every submap after the first opens at the frame after an uncovered frame, and an uncovered frame opens at most one submap, so $C\le1+U$.
A keyframe is the view of frame 0, an uncovered view or an anchor, so $K\le1+U+(C-1)\le1+2U$.
These bounds are functions of the set $X$.

\emph{Drift check.} A return restores the state of a submap as it was left, so the state of a submap changes through its own writes alone.
Let $s'$ be the state of a submap after its anchor is written and $s$ its state after $j$ further full updates of case~3.
Each write changes the state by at most $B$, so $\|s-s'\|\le jB$, and the pose of the anchor decoded with $s$ lies within $LjB$ of its opening pose.
A submap that the check closes has this distance at least $\tau$, so $j\ge\tau/(LB)$ and the submap holds $1+j>\tau/(LB)$ keyframes.
If one state writes $K_1\le\tau/(LB)$ views on the video, \scenebound{} follows it write for write as long as its first submap is open, and after $j\le K_1-1$ writes the decoded anchor lies within $LB(K_1-1)<\tau$ of its opening pose, so the first submap never closes and the two rules coincide.
By \cref{thm:keyframes}, $K_1\le N_d(X,M/2)$.

\emph{Placements.} A submap is placed once, by one read against the submap it replaces, and cases~1 and~2 create no submap, so the submaps form a tree and a return adds no placement.
\end{proof}

%% file: thm/drift.tex
With increments of norm at most $B$ and a decoded pose that moves by at most $L$ times a change of the state, \scenebound{} in one state has $\|s_N-s_0\|\le B\,N_d(X,M/2)$ for every length $N$, and the pose it predicts for any fixed frame moves by at most $LB\,N_d(X,M/2)$.
Under the conditions of \cref{thm:budget}, a full update every $P$ frames has $\E\|\phi(s_N)-\phi(s_0)\|\ge\kappa\lfloor N/P\rfloor$, which grows with $N$, and every rule of full updates that keeps frame 0 or a written frame at distance below $M$ from every pose of $X$ has $\E\|\phi(s_N)-\phi(s_0)\|\ge\kappa(\E N_d(X,M)-1)$, and \scenebound{} has $\|\phi(s_N)-\phi(s_0)\|\le\lambda LB\,N_d(X,M)$.

%% file: supp/B_baselines.tex
\section{Budget-matched baselines}
\label{app:baselines}

\subsection{Construction}
\label{app:construction}

For the budget-matched baseline of \cref{sec:baselines}, we spend the logged budget of a gate on a uniformly spaced schedule.
\Cref{prop:schedule} states the schedule and the range of the two sums. By \cref{thm:budget}, $W$ sets the band of the drift and $Q$ bounds the part that the past does not predict, and under the constant push of \cref{prop:push} the expected squared drift depends on the gains through these two sums alone (\cref{prop:sums}), so matching them suffices.

\begin{proposition}[Uniformly spaced schedule]
\label{prop:schedule}
Let $N\ge1$ and $1\le J\le N$ be integers and $g\in(0,1]$.
The schedule with $b_t=g$ at the frames $t\in\{1,\dots,N\}$ with $\lceil tJ/N\rceil>\lceil(t-1)J/N\rceil$ and $b_t=0$ elsewhere writes at exactly $J$ frames, has $\sum_tb_t=gJ$ and $\sum_tb_t^2=g^2J$, and its prefix sums satisfy $0\le\sum_{u\le t}b_u-tgJ/N<g$ for every $t$.
Every gain sequence in $[0,1]^N$ with budget $W>0$ has $W^2/N\le Q\le W$, where the constant gain $W/N$ attains the first bound and gains in $\{0,1\}$ attain the second.
Hence $W\le W^2/Q\le N$, and for every integer $J$ with $W\le J\le N$ the gain $g=W/J$ lies in $(0,1]$ and gives the budget $W$ and the sum of squared gains $W^2/J$, which equals $Q$ when $J=W^2/Q$.
\end{proposition}

\begin{proof}
The number of writes up to frame $t$ is $\lceil tJ/N\rceil$ by telescoping, which equals $J$ at $t=N$, and $0\le\lceil x\rceil-x<1$ yields the bound on the prefix sums after multiplication by $g$.
By the Cauchy--Schwarz inequality, $W^2\le N\sum_t\beta_t^2$, with equality exactly for a constant $\beta$, and $\beta_t^2\le\beta_t$ on $[0,1]$, with equality exactly at the end points; hence $W^2/N\le Q\le W$ and $W\le W^2/Q\le N$.
Then $J\ge W$ implies $g=W/J\le1$, and the sums are $gJ=W$ and $g^2J=W^2/J$.
\end{proof}

\begin{proposition}[Two sums]
\label{prop:sums}
Under the constant push of \cref{prop:push} the increments of a schedule have the Gram matrix $G_{ij}=\E\langle\xi_i,\xi_j\rangle=\omega\,[i=j]+\|\mu\|^2$, so that $\E\|s_N-s_0\|^2=\beta^\top G\beta$ depends on the gains through $W$ and $Q$ alone.
If a symmetric matrix $G$ and a function $h$ satisfy $\beta^\top G\beta=h(W,Q)$ for every $\beta\in[0,1]^N$ with $N\ge2$, then $G=\omega I+\gamma\,\mathbf 1\mathbf 1^\top$ with $\gamma=G_{12}$ and $\omega=G_{11}-G_{12}$.
\end{proposition}

\begin{proof}
For $i<j$ the noise $\zeta_i$ is fixed by the information before frame $j$; thus $\E\langle\zeta_i,\zeta_j\rangle=0$, and $\E\langle\mu,\zeta_j\rangle=0$.
With $\E\|\zeta_j\|^2=\omega$, this yields the Gram matrix and $\beta^\top G\beta=\omega Q+\|\mu\|^2W^2$.
For the converse, let $\pi$ be a permutation matrix.
The cube $[0,1]^N$ is invariant under $\pi$, and $W$ and $Q$ take the same values at $\beta$ and at $\pi\beta$; therefore $\beta^\top(\pi^\top G\pi-G)\beta=0$ on the cube.
A quadratic form that vanishes on a set with nonempty interior has a zero matrix, and hence $G$ is invariant under every permutation of the indices.
A transposition yields $G_{ii}=G_{jj}$, and a permutation carries any ordered pair of distinct indices to any other; thus $G$ has one diagonal value and one off-diagonal value.
\end{proof}

In the experiments, we take $J$ as the integer in $[\lceil W\rceil,N]$ nearest $W^2/Q$, which makes the budget exact and places $W^2/J$ within a factor $1\pm1/J$ of $Q$.
Two schedules with equal sums move a state differently when the increments are correlated in time.
For scalar increments $(Z,Z,-Z)$ with a random sign $Z$, the schedules $(1,1,0)$ and $(1,0,1)$ have equal sums and displacements 2 and 0.
Since the baseline spreads its writes uniformly, the comparison of a gate with its baseline measures the placement of its gains and their feedback on later increments together.

\subsection{Gates versus their baselines}
\label{app:gates}

\begin{table*}[t]
\caption{Each gate against its budget-matched baseline and its matched constant on the 61 benchmark sequences, with a paired resampling interval, the ratio per dataset and the spread over sequences.}
\label{tab:gates-detail}
\centering
\footnotesize
\begin{fulltabular}{@{}lccccccccc@{}}
\toprule
 & \multicolumn{3}{c}{Mean ATE, baseline / gate} & \multicolumn{4}{c}{By dataset} & \multicolumn{2}{c}{Per sequence} \\
\cmidrule(lr){2-4}\cmidrule(lr){5-8}\cmidrule(lr){9-10}
Gate & Baseline\textsuperscript{\dag} & Interval & Constant\textsuperscript{\dag} & TUM-dyn & Bonn & 7-Scenes & NRGBD & Quartiles & Baseline lower \\
\midrule
TTT3R~\cite{chen2026ttt3r} & 1.04 & 0.99--1.09 & 1.07 & 0.97 & 1.05 & 1.02 & 1.05 & 0.91--1.12 & 26 of 61 \\
MeMix~\cite{dong2026memix} & 0.95 & 0.79--1.09 & 1.00 & 0.92 & 1.04 & 1.00 & 0.77 & 0.88--1.12 & 38 of 61 \\
TTSA3R~\cite{zheng2026ttsa3r} & 1.06 & 1.01--1.12 & 1.09 & 0.99 & 1.05 & 1.03 & 1.20 & 0.95--1.11 & 22 of 61 \\
FILT3R~\cite{jin2026filt3r} & 1.04 & 0.97--1.12 & 1.08 & 0.99 & 1.10 & 1.01 & 1.00 & 0.91--1.10 & 26 of 61 \\
ReCal3R~\cite{li2026recal3r} & 0.99 & 0.95--1.02 & 0.99 & 0.97 & 1.01 & 0.94 & 1.01 & 0.94--1.02 & 39 of 61 \\
AFG-Img\textsuperscript{\ddag}~\cite{ren2026rethinking} & 0.92 & 0.86--0.97 & 0.91 & 0.94 & 0.90 & 0.94 & 0.94 & 0.87--1.01 & 41 of 61 \\
AFG-Pose\textsuperscript{\ddag}~\cite{ren2026rethinking} & 0.97 & 0.91--1.06 & 0.97 & 1.51 & 0.89 & 0.98 & 0.93 & 0.84--1.04 & 44 of 61 \\
Info3R\textsuperscript{\ddag}~\cite{baek2026info3r}, gate & 1.01 & 0.98--1.05 & 1.02 & 1.07 & 1.00 & 0.98 & 1.08 & 0.97--1.03 & 29 of 61 \\
BAT3R~\cite{li2026bat3r} & 0.95 & 0.87--1.03 & 1.01 & 0.90 & 0.96 & 0.96 & 0.92 & 0.88--1.05 & 39 of 61 \\
\oursrow \scenebound{} (ours) & 1.12 & 1.02--1.29 & 1.21 & 1.69 & 1.16 & 1.00 & 1.03 & 0.97--1.09 & \oursend 28 of 61 \\
\bottomrule
\end{fulltabular}

\smallskip
\raggedright\scriptsize
\scenebound{} runs in one state. $^{\dag}$Constructed baseline. $^{\ddag}$Our implementation of a method without released code.
\end{table*}

\begin{figure}[t]
\centering
\includegraphics[width=\linewidth]{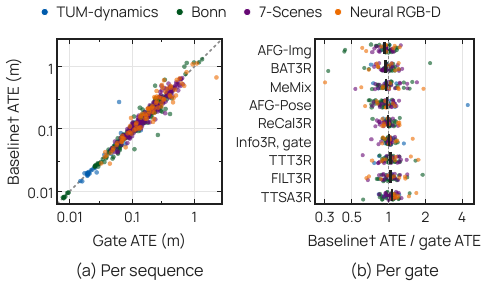}
\caption{Each published gate and its budget-matched baseline\textsuperscript{\dag} on the 61 benchmark sequences, coloured by dataset. (a) The two errors per sequence, with the line of equal error. (b) Their ratio per sequence; bars mark the ratio of the mean errors.}
\label{fig:gate-pairs}
\end{figure}

Per sequence, the gates and their baselines fall on both sides of equal error (\cref{fig:gate-pairs}).
We list each gate in \cref{tab:gates-detail} with a paired resampling interval of its ratio over the 61 sequences, taken as the central 95\% of 4,000 draws of sequences with replacement.
When we resample the 38 scenes, holding the sequences of one 7-Scenes room or one Bonn scene together, both ends of all intervals move by at most 0.01.
The gate of TTSA3R is ahead of its baseline at 1.06, with an interval from 1.01 to 1.12, and the baseline of AFG-Img is ahead of its gate at 0.92, with an interval from 0.86 to 0.97.
The largest ratio on one dataset, AFG-Pose on TUM-dynamics at 1.51 over eight sequences, comes from one sequence, sitting halfsphere, where the baseline and the matched constant reach 0.271 and 0.272 and the gate reaches 0.062.
Single sequences show differences between a gate and its baseline, with quartiles of the ratio between 0.84 and 1.12 across the gates.
The match therefore holds for the benchmark mean, and single scenes place a gate ahead of or behind its baseline.
The baselines of these soft gates write on 73\% to 99\% of the frames at mean gains of 0.04 to 0.32, close to a constant gain.

We run three further controls that remove what a baseline knows about its gate.
A constant gain on each sequence, the total budget of the gate divided by its frames, reaches 0.91 to 1.09 times the error of the gate, 1.01 in the geometric mean, and is lower on 24 to 44 of the 61 sequences.
When we move the writes of each baseline to frames drawn at random, the error is 1.00 to 1.06 times that of the uniformly spaced baseline.
The match thus holds for spaced, random and constant placements of the budget.
For \scenebound{} in one state, the three placements of its budget result in 0.0805 at new views, 0.0904 at uniform sampling and 0.0913 at random frames, and its lead over uniform sampling falls on the two datasets with moving people, with a ratio of 1.07 on the 53 sequences outside TUM-dynamics.

The published papers of these gates compare them with signal-free schedules at budgets chosen by hand.
AFG compares its gate with constants from 0.1 to 0.7 and with writes on alternate frames and on one frame in three, and in its own table a constant of 0.5 matches writes on alternate frames and a constant of 0.3 matches writes on one frame in three, as a budget would predict.
FILT3R reports a fixed gain of 0.05, which writes a third of what its gate writes, with a lower trajectory error and a higher rotation error than the gate.
MeMix reports that random tokens at an equal count reach the pose error of its rule.

\begin{table}[t]
\caption{Baseline\textsuperscript{\dag} over gate ATE inside the reset of the TTT3R release, in the second checkpoint at 224 pixels, on Replica, and inside the reset on the 11 KITTI drives and 12 TUM RGB-D sequences.}
\label{tab:settings}
\centering
\footnotesize
\begin{fulltabular}{@{}lccccc@{}}
\toprule
 & \multicolumn{3}{c}{Benchmark and Replica} & \multicolumn{2}{c}{Full length, reset} \\
\cmidrule(lr){2-4}\cmidrule(lr){5-6}
Gate & Reset & 224\,px & Replica & KITTI & TUM \\
\midrule
TTT3R~\cite{chen2026ttt3r} & 0.92 & 1.04 & 0.96 & 1.24 & 1.24 \\
MeMix~\cite{dong2026memix} & 0.88 & 1.04 & 0.97 & 1.15 & 1.17 \\
TTSA3R~\cite{zheng2026ttsa3r} & 1.01 & 1.06 & 1.16 & 1.15 & 1.21 \\
FILT3R~\cite{jin2026filt3r} & 1.01 & 1.06 & 1.09 & 1.06 & 0.99 \\
AFG-Img\textsuperscript{\ddag}~\cite{ren2026rethinking} & 0.92 & 0.99 & 0.90 & 1.25 & 1.35 \\
ReCal3R~\cite{li2026recal3r} & 1.00 & 1.01 & 0.95 & 1.07 & 1.02 \\
Info3R\textsuperscript{\ddag}~\cite{baek2026info3r}, gate & 0.99 & 1.13 & 1.03 & 1.05 & 0.97 \\
AFG-Pose\textsuperscript{\ddag}~\cite{ren2026rethinking} & 0.99 & 1.01 & 0.81 & 1.05 & 1.12 \\
\midrule
Geometric mean & 0.96 & 1.04 & 0.98 & 1.12 & 1.13 \\
\bottomrule
\end{fulltabular}

\smallskip
\raggedright\scriptsize
$^{\dag}$Constructed baseline. $^{\ddag}$Our implementation of a method without released code.
\end{table}

\begin{table*}[t]
\caption{Gram matrix of the state increments, the mean off-diagonal entry over the mean diagonal entry in each band of lags.}
\label{tab:gram}
\centering
\footnotesize
\begin{fulltabular}{@{}llcrrrrrrrr@{}}
\toprule
 & & & \multicolumn{8}{c}{Lag in frames} \\
\cmidrule(lr){4-11}
Schedule & Set & Sequences & 8--16 & 24--48 & 56--104 & 112--200 & 208--400 & 408--600 & 608--800 & 808--992 \\
\midrule
Constant gain 1 (CUT3R) & TUM-dyn & 8 & $-0.011$ & $-0.005$ & $-0.001$ & 0.000 & 0.000 & 0.000 & $-0.001$ & $-0.003$ \\
Constant gain 1 (CUT3R) & NRGBD & 9 & $-0.001$ & $-0.003$ & $-0.001$ & 0.000 & 0.000 & 0.000 & 0.000 & 0.000 \\
Constant gain 0.3\textsuperscript{\dag} & TUM-dyn & 8 & 0.018 & $-0.028$ & $-0.014$ & $-0.003$ & $-0.001$ & 0.000 & $-0.001$ & 0.000 \\
Constant gain 0.3\textsuperscript{\dag} & NRGBD & 9 & 0.035 & $-0.013$ & $-0.011$ & $-0.005$ & $-0.002$ & $-0.001$ & 0.000 & $-0.001$ \\
Constant gain 0.05\textsuperscript{\dag} & NRGBD & 9 & 0.464 & 0.162 & 0.003 & $-0.047$ & $-0.036$ & $-0.014$ & $-0.007$ & $-0.002$ \\
\oursrow \scenebound{} (ours) & NRGBD & 9 & 0.528 & 0.139 & 0.012 & $-0.025$ & $-0.024$ & $-0.013$ & $-0.006$ & \oursend 0.002 \\
\bottomrule
\end{fulltabular}

\smallskip
\raggedright\scriptsize
$^{\dag}$Constructed baseline.
\end{table*}

\subsection{Five further settings}
\label{app:settings}

We repeat the comparison in five settings (\cref{tab:settings}).
Inside the periodic reset of TTT3R, which starts a new state every 100 frames, the long-sequence benchmark shows the same picture as \cref{sec:baselines}, with a geometric mean of 0.96.
In the second released checkpoint, at 224 pixels, the baselines have 0.99 to 1.13 times the error of their gates, each rule has 1.4 to 2.7 times its error in the first checkpoint, and the eight gates and \scenebound{} rank identically by budget and by error.
On the held-out Replica, the ratio ranges from 0.81 to 1.16, with 0.98 in the geometric mean, and across the gates and CUT3R the rank correlation between budget and error is 1.00.

Inside short states on long sequences, the signals of several gates add accuracy beyond their budget.
Across the 11 KITTI drives, all gates are ahead of their baselines by 5\% to 25\%, and the rank correlation between budget and error falls to 0.45.
On the 12 TUM RGB-D sequences outside development, four gates are ahead by 17\% to 35\% and AFG-Pose by 12\%, and the baselines fall between 0.526 and 0.573 at budgets from 1,124 to 8,567.
We run two controls that preserve the gain of TTT3R frame by frame, one applying the mean gain of each frame to all tokens and one handing the gains of each frame to its tokens in random order.
On KITTI these controls reach 76.3 and 74.6\,m, compared with 61.7 for the gate and 76.3 for its baseline, and on TUM RGB-D they reach 0.531 and 0.551, compared with 0.449 and 0.557.
Inside the reset, the gate therefore gains by which tokens it writes.
In the cache of RegVGGT, three choices of keys that use no score are 4\% to 7\% below its scored selection at 10\% of the keys.
At 1\% of the keys, random keys are 6\% above it with more than twice its rotation RPE, and uniformly spaced keys reach 1.94 times its error.

\subsection{The increments}
\label{app:increments}

We decode frame 0 again with the current state and no write, which returns the pose that the state now assigns to its first camera, and the distance of this pose from the pose at the start measures the displacement of the coordinate frame.
Over five constant gains and three rules of full updates, the slope of the logarithm of this displacement versus the logarithm of the budget is 0.63, 1.00, 1.41 and 1.76 on TUM-dynamics, Bonn, 7-Scenes and Neural RGB-D after 500 frames, as in \cref{fig:budget}a, and 0.98, 0.98, 1.34 and 1.37 after 1,000 frames.
Over all four datasets the slope of the geometric mean is 1.2 after 500 frames, and the shift per unit budget stays within a factor of 2.4 across the eight rules, the band of \cref{thm:budget}; increments without a common direction would give a slope of 0.5.
For the ratio test of \cref{sec:baselines}, \scenebound{} writes 23.0 and 22.5 full updates per sequence over the first 500 frames of 7-Scenes and Neural RGB-D, so a constant gain of equal budget has $Q=W^2/N$, 22 times less, and a dominant uncorrelated random part would give $\sqrt{22}\approx4.7$ times its shift.
Interpolated in log budget between the constant gains of 0.03 and 0.05, the keyframes shift the first camera by 0.92 and 1.11 times as much after 500 frames and by 0.93 and 1.53 times after 1,000 frames.
The Gram matrix of the increments in \cref{tab:gram}, which we compute from one increment in eight along each sequence, is short-ranged and grows as the gain falls.
At gains of 1 and 0.3, the state increments are therefore nearly uncorrelated, with a common part near zero, and the change that grows with the budget is the coherent change of the read-out that \cref{thm:budget} bounds.
At small gains, neighbouring increments repeat one push, and the moments of the increments thus differ between schedules.

Two experiments test what the gains say about the tokens.
Permuting the gains of a gate over the state tokens of each frame preserves their distribution, their mean and their course over time, and results in 0.92 to 1.11 times the error of the gate for TTT3R, TTSA3R, FILT3R and AFG-Pose, with MeMix at 0.96 for random tokens of its count.
A keyframe that reaches part of the state, either with each token at the TTT3R gain of the frame or as a full update of the half of the tokens with the largest gain, is above the full update of \scenebound{} (0.0805) in all forms.
The half with the largest gain is also above a random half at both radii tried, with 0.230 versus 0.123 and 0.263 versus 0.116.

\subsection{Keyframe selection tests across thresholds}
\label{app:tests}

We port each test from the official code of its system without the trained parts, as a rule that writes a whole frame to the state of CUT3R or retains it in the cache of StreamVGGT\@.
AMB3R (commit \texttt{92c4081}) writes a frame when the rotation angle over $\pi$ plus the distance of its centre, in the depth unit of the first frame, exceeds the threshold for all active keyframes, and holds at most 10 active keyframes, resampled as released.
MUSt3R (commit \texttt{efc5b08}) writes a frame when the 85th percentile of the depth-normalised distances from its world points to the stored points seen in the same octant of view directions exceeds the threshold, with the confidence limit of its release.
Scal3R applies the same percentile to the points of the frame placed by the predicted camera, which stands in for its trained pose graph; in the state it follows commit \texttt{af4c8e8} and retains its first four frames and the last four written, and in the cache it follows the STream3R branch of its release (commit \texttt{87a7924}) and compares the frame with frame 0 and the last 12 frames kept.
We run each test at four thresholds, including the released one.

\begin{table}[t]
\caption{The ported keyframe selection tests at each threshold on the benchmark, with summed writes or kept frames and mean ATE (m), the cache with the camera head kept whole.}
\label{tab:tests-detail}
\centering
\footnotesize
\begin{fulltabular}{@{}lrrcrc@{}}
\toprule
 & & \multicolumn{2}{c}{State} & \multicolumn{2}{c}{Cache} \\
\cmidrule(lr){3-4}\cmidrule(lr){5-6}
Test & Threshold & Writes & ATE $\downarrow$ & Kept & ATE $\downarrow$ \\
\midrule
AMB3R & 0.125 & 2,069 & 0.0901 & 2,117 & 0.0888 \\
 & 0.15 & 1,687 & 0.0876 & 1,707 & 0.0873 \\
 & 0.175 & 1,422 & 0.0875 & 1,416 & 0.0879 \\
 & 0.2 & 1,219 & 0.0848 & 1,225 & 0.0908 \\
MUSt3R & 0.025 & 2,168 & 0.1113 & 2,798 & 0.0919 \\
 & 0.05 & 1,451 & 0.0885 & 1,650 & 0.0880 \\
 & 0.052 & 1,415 & 0.0887 & 1,604 & 0.0887 \\
 & 0.1 & 958 & 0.0867 & 995 & 0.0884 \\
 & 0.2 & 597 & 0.0855 & 566 & 0.0924 \\
Scal3R & 0.05 & 2,560 & 0.0896 & 1,981 & 0.0879 \\
 & 0.1 & 1,576 & \underline{0.0847} & 1,066 & 0.0873 \\
 & 0.115 & 1,417 & 0.0864 & 921 & 0.0881 \\
 & 0.2 & 960 & 0.0862 & 576 & 0.0940 \\
 & 0.4 & 572 & 0.0891 & 322 & 0.1022 \\
\midrule
Uniform sampling\textsuperscript{\dag} & none & 1,421 & 0.0904 & 1,426 & \textbf{0.0853} \\
\oursrow \scenebound{} (ours) & $M=0.3$ & 1,421 & \textbf{0.0805} & 1,426 & \oursend \underline{0.0870} \\
\bottomrule
\end{fulltabular}

\smallskip
\raggedright\scriptsize
Every rule runs in one state. $^{\dag}$Constructed baseline.
\end{table}

In the state, each test at each threshold is 5\% to 38\% above \scenebound{} (\cref{tab:tests-detail}).
At the thresholds whose total count matches that of \scenebound{}, AMB3R at 0.175, Scal3R at 0.115 and MUSt3R at 0.052 write 1,422, 1,417 and 1,415 frames and reach 0.0875, 0.0864 and 0.0887 against 0.0805, the rows of \cref{tab:ablation}.
Most of the gap lies on TUM-dynamics, where the two point tests reach 0.093 and 0.090 against 0.048: on the two sequences in which a static camera watches people walk past, they write 22 and 23 frames where the pose test writes one, and on the two half-sphere sweeps, which view one desk from new directions, they write 12 to 44 frames against 26 and 25 and reach 0.19 to 0.22 against 0.049 and 0.073.
On Bonn, which also holds moving people, they are 4\% and 5\% above \scenebound{}, on 7-Scenes within 2\%, and on Neural RGB-D Scal3R is 8\% below it.
When we interpolate in log count to the count of \scenebound{} on each sequence, AMB3R is 5.9\% above \scenebound{} on the 54 sequences that its thresholds bracket (paired interval 1.01 to 1.14), MUSt3R is 5.7\% above on 37 (0.97 to 1.17) and Scal3R is 8.7\% above on 39 (1.01 to 1.19).
With the camera head kept whole, the three tests in the cache are 1.6\%, 3.1\% and 1.7\% above \scenebound{} on the 55, 45 and 39 sequences that their thresholds bracket, with intervals from 0.95 to 1.08.
With the camera head pruned, as in \cref{tab:ablation}, AMB3R at 0.15 and 0.175, MUSt3R at 0.05 and 0.1 and Scal3R at 0.05 and 0.1 reach 0.0830, 0.0825, 0.0846, 0.0844, 0.0844 and 0.0834 with 1,703, 1,406, 1,662, 996, 1,990 and 1,066 frames, and at the 1,433 frames of \scenebound{}, interpolated in log count, 0.0826, 0.0845 and 0.0839 against 0.0831 for \scenebound{} and 0.0841 for uniform sampling; per sequence, the two thresholds of each test bracket 34, 21 and 24 sequences, on which the tests reach 1.00, 1.02 and 1.00 times \scenebound{}, with intervals from 0.91 to 1.09.
Run at the matched thresholds with the camera head pruned, Scal3R at 0.072 and MUSt3R at 0.061 keep 1,413 and 1,423 frames at 0.0845 and 0.0831, against 1,433 frames at 0.0831 for \scenebound{}, and process 5.7 and 4.8 frames per second against 6.3 for \scenebound{} and 6.4 for AMB3R.

Two tests against recent writes alone complete the comparison of history scopes.
A test against the latest keyframe, at radius $M$, writes 1,616 frames on the benchmark at 0.0820, against 1,421 at 0.0805, and on the 24 revisit sequences 788 frames in the first passes and 6,651 over all nine, at 0.206 against 0.084.
In the cache it keeps 1,638 frames on the benchmark at 0.0846, against 1,433 at 0.0831, and over nine passes it fills the GPU on 12 of the 17 revisit sequences, between frames 5,071 and 8,503, and keeps 7.95 times its first passes on the other five.
A test of the distance travelled since the last write, at the same radius, writes 2,751 frames on the benchmark at 0.086 and 11,788 over the nine passes at 0.354.
Over the nine passes in the state, AMB3R, Scal3R and MUSt3R write 4,577, 3,181 and 701 frames, 6.54, 5.52 and 1.16 times their first passes, at 0.170, 0.143 and 0.129, and uniform sampling, whose frames are fixed in advance, spreads the 722 writes of \scenebound{} evenly, 113 of them in the first passes, a growth of 6.39.

\begin{table}[t]
\caption{The ported keyframe selection tests on the 17 revisit sequences in the cache, beside \scenebound{} on the videos each test completed, every rule with the camera head pruned: GPU memory (GB) and ATE (m) after nine passes.}
\label{tab:tests-looped}
\centering
\footnotesize
\begin{fulltabular}{@{}lrcccc@{}}
\toprule
 & & \multicolumn{2}{c}{Frames kept} & \multicolumn{2}{c}{9 passes} \\
\cmidrule(lr){3-4}\cmidrule(lr){5-6}
Rule & Videos & 1 pass & 9 passes & GPU $\downarrow$ & ATE $\downarrow$ \\
\midrule
AMB3R, 0.175 & 9 of 17 & 107 & 579 & 24.4 & 0.033 \\
\oursrow \quad \scenebound{} (ours) & 9 & 131 & 132 & 11.6 & \oursend 0.032 \\
MUSt3R, 0.061 & 17 of 17 & 402 & 449 & 14.6 & 0.062 \\
\oursrow \quad \scenebound{} (ours) & 17 & 426 & 450 & 14.8 & \oursend 0.069 \\
Scal3R, 0.072 & 15 of 17 & 289 & 810 & 21.7 & 0.055 \\
\oursrow \quad \scenebound{} (ours) & 15 & 349 & 360 & 14.1 & \oursend 0.063 \\
\bottomrule
\end{fulltabular}
\end{table}

On the revisit sequences, we run each test in the cache at the threshold whose total count on the long-sequence benchmark matches that of \scenebound{} (\cref{tab:tests-looped}).
The test of MUSt3R compares a frame with the points of each stored frame and completes all 17 videos, with 402 frames kept in the first passes and 449 in all nine, 1.12 times, against 1.06 times for \scenebound{}, at 14.6 against 14.8\,GB.
AMB3R compares a frame with its 10 active keyframes and Scal3R with its first four and last four writes, so a place seen on an earlier pass is new to them once its keyframes have left that set.
AMB3R fills the 80\,GB GPU on the same eight of the 17 videos as with the camera head kept whole, where \scenebound{} retains 29 to 64 frames over all nine passes, and its kept frames grow 5.4 times on the other nine, against 1.01 times for \scenebound{}.
Scal3R fills the GPU on two videos and grows 2.8 times on the other 15, against 1.03 times for \scenebound{}.
The tests of MUSt3R and Scal3R end at 0.062 and 0.055 against 0.069 and 0.063 for \scenebound{}, and one video makes the difference, the breakfast room of Neural RGB-D.
On its second to fourth passes, six frames receive a predicted pose up to 2.6 from that of the frame before, beyond the radius $M$, \scenebound{} writes them as new views, and every later pass reads these misplaced views, so its error rises from 0.043 after one pass to 0.215 over nine; the test of MUSt3R finds the points of these frames stored and writes none.
Without this video, MUSt3R and Scal3R reach 0.063 and 0.055 against 0.060 and 0.052 for \scenebound{}.
\Cref{thm:submaps} assumes steps below $M$ between consecutive poses, and a write rule can hold its writes to this bound by measuring the step of each frame from the last frame that passed it; a frame beyond the bound is decoded and not written.
With this test, \scenebound{} keeps the six views out, writes three frames of later passes whose predicted pose lies beyond $M$ of every keyframe, and reaches 0.106 on the breakfast room with 46 frames.
Over the 17 videos its mean falls from 0.069 to 0.062, below the 0.067 of every 50 frames, with 447 frames, and on the long-sequence benchmark it moves from 0.0831 to 0.0832.

%% file: supp/C_method.tex
\section{\scenebound{} in detail}
\label{app:method}

\subsection{Constants and their selection}
\label{app:constants}

The rule has two constants.
We fixed the keyframe radius $M=0.3$ on 15 TUM sequences, seven static sequences outside the benchmark and the eight TUM-dynamics sequences of the benchmark, choosing from the values 0.15, 0.2 and 0.3.
The depth unit of the distance, the median depth of frame 0, replaced a running unit after a comparison on the 61 benchmark sequences (0.0805 versus 0.0831).
\Cref{tab:radius} reports the benchmark at six radii in one state, which we ran after these choices.
The error falls from 0.2 to 0.3 and rises again above 0.4.
A radius of 0.3 yields the lowest error of the six on the benchmark and on TUM-dynamics, and on the 53 sequences outside TUM-dynamics the response is flat within 1.5\% between 0.3 and 0.5.
On these 53 sequences, \scenebound{} reaches 0.0854 in the state and 0.0906 in the cache, 9\% and 17\% below the best published rules, Info3R in full (0.094) and RetrieveVGGT (0.110), and every 50 frames reaches 0.0839 and 0.0943.

\begin{table}[t]
\caption{Keyframe radius on the benchmark in one state.}
\label{tab:radius}
\centering
\footnotesize
\begin{fulltabular}{@{}lcccr@{}}
\toprule
 & \multicolumn{3}{c}{Mean ATE (m) $\downarrow$} & \\
\cmidrule(lr){2-4}
$M$ & Bench. & Outside TUM-dyn & TUM-dyn & Writes \\
\midrule
0.2 & 0.0910 & 0.0967 & 0.053 & 2,312 \\
0.25 & 0.0857 & 0.0911 & \underline{0.050} & 1,785 \\
\oursrow 0.3 (ours) & \textbf{0.0805} & \underline{0.0855} & \textbf{0.048} & \oursend 1,421 \\
0.35 & \underline{0.0808} & 0.0856 & \underline{0.050} & 1,166 \\
0.4 & 0.0812 & \textbf{0.0843} & 0.061 & 992 \\
0.5 & 0.0860 & \textbf{0.0843} & 0.098 & 731 \\
\bottomrule
\end{fulltabular}
\end{table}

\begin{table}[t]
\caption{Depth unit of frame 0 scaled by $\rho$ on the benchmark in one state, the unit of a first frame that sees its scene $\rho$ times farther away, fixed and rising with the keyframes.}
\label{tab:unit}
\centering
\footnotesize
\begin{fulltabular}{@{}lcccccr@{}}
\toprule
 & \multicolumn{5}{c}{Mean ATE (m) $\downarrow$} & \\
\cmidrule(lr){2-6}
$\rho$ & TUM-dyn & Bonn & 7-Scenes & NRGBD & Bench. & Writes \\
\midrule
1/4 & 0.058 & 0.086 & 0.105 & 0.145 & 0.0967 & 3,415 \\
1/2 & 0.051 & 0.082 & 0.107 & 0.107 & 0.0890 & 2,100 \\
\oursrow 1 (ours) & \textbf{0.048} & 0.078 & \textbf{0.095} & 0.088 & \underline{0.0804} & \oursend 1,423 \\
2 & 0.055 & 0.082 & \underline{0.096} & \underline{0.084} & 0.0830 & 1,061 \\
4 & 0.050 & \textbf{0.076} & 0.109 & \textbf{0.081} & 0.0830 & 828 \\
\midrule
\multicolumn{7}{@{}l}{\emph{Unit rising with the median depth of the keyframes}} \\
1/4 & \underline{0.049} & 0.082 & \underline{0.096} & 0.090 & 0.0832 & 1,432 \\
1/2 & \textbf{0.048} & 0.080 & 0.097 & 0.089 & 0.0821 & 1,418 \\
1 & 0.050 & \underline{0.077} & \textbf{0.095} & 0.088 & \textbf{0.0801} & 1,374 \\
\bottomrule
\end{fulltabular}
\end{table}

\begin{table}[t]
\caption{Threshold of the drift check on all sets where it ran.}
\label{tab:threshold}
\centering
\footnotesize
\begin{fulltabular}{@{}lcccr@{}}
\toprule
 & \multicolumn{4}{c}{Mean ATE (m) $\downarrow$} \\
\cmidrule(lr){2-5}
 & & \multicolumn{2}{c}{TUM RGB-D} & \\
\cmidrule(lr){3-4}
$\tau$ & Bench. & Dev., 7 & Other, 12 & KITTI, 02--10 \\
\midrule
0.05 & 0.0815 & 0.126 & 0.544 & \textbf{36.9} \\
0.075 & 0.0812 & \underline{0.117} & \underline{0.529} & \underline{39.8} \\
\oursrow 0.1 (ours) & \textbf{0.0804} & \textbf{0.112} & \textbf{0.503} & \oursend 48.0 \\
0.15 & \textbf{0.0804} & 0.210 & 0.543 & 40.9 \\
0.2 & \underline{0.0805} & 0.326 & 0.608 & 42.2 \\
0.3 & \underline{0.0805} & 0.249 & 0.687 & 43.2 \\
0.45 & \underline{0.0805} & 0.268 & 0.722 & 49.6 \\
0.6 & \underline{0.0805} & 0.275 & 0.668 & 42.3 \\
One state & \underline{0.0805} & 0.138 & 0.713 & 136.8 \\
\bottomrule
\end{fulltabular}
\end{table}

The depth unit is the median depth that frame 0 predicts, and \cref{prop:unit} preserves every bound for any fixed unit.
Across the 61 benchmark sequences, the ratio of each frame's median depth to this unit has 5th and 95th percentiles of 0.67 and 1.15 in the median sequence, and 0.11 and 2.47 at the extremes.
\Cref{tab:unit} scales the unit by $\rho$ from $1/4$ to $4$ in both the keyframe test and the drift check, matching a first frame that sees its scene $\rho$ times farther away than at $\rho=1$.
A larger unit measures the scene more coarsely, writes 0.75 and 0.58 times as many frames at $\rho=2$ and $4$, and raises the mean ATE by 3\%.
A smaller unit measures it more finely, writes 1.5 and 2.4 times as many frames at $\rho=1/2$ and $1/4$, and raises the mean ATE by 11\% and 20\%, as \cref{thm:budget} predicts for a larger update budget; the radius sweep of \cref{tab:radius} shows the same response.
A first frame that faces a near wall therefore costs accuracy through the writes it adds, and it leaves memory bounded at every sequence length.
A unit that rises, never falls, to the median of the median depths of frame 0 and of every keyframe written since removes most of this cost.
It never falls below the unit of frame 0, so the count bound at that unit holds (\cref{app:proof-keyframes}), and once the keyframes outnumber the first frame it follows their depth.
On the benchmark it reaches 0.0832 and 0.0821 at $\rho=1/4$ and $1/2$, 3\% and 2\% above the fixed unit at $\rho=1$, with 1,432 and 1,418 writes, and 0.0801 with 1,374 writes at $\rho=1$.
All other results of the paper use the fixed unit of frame 0.

We selected the threshold $\tau=0.1$ among 10 candidates by the lowest mean ATE on seven of the 19 TUM RGB-D sequences.
The candidates are five thresholds from 0.05 to 0.2, each with the plain placement of a submap and with a placement estimated from the anchor and the eight frames after it, each read against both submaps, by chordal-mean rotation, median translation and median depth ratio.
At the five thresholds, the plain placement reaches 0.126, 0.117, 0.112, 0.210 and 0.326, and the bridge reaches 0.121, 0.201, 0.129, 0.115 and 0.116.
\Cref{tab:threshold} reports all thresholds of the plain placement on each set, with three larger ones from a separate run.
The benchmark responds by at most 1.4\%, since the check fires on few of its sequences.
On the 12 other TUM RGB-D sequences, the threshold 0.1 has the lowest error of the eight.
On KITTI, the eight thresholds result in 36.9 to 49.6\,m, the lowest at the smallest threshold, 0.05, and the selected one in 48.0.

\subsection{Closing a submap}
\label{app:closing}

\Cref{tab:closing} lists all ways of closing a state that we ran.
On KITTI sequences 02 to 10, the drift check opens 301 submaps over 17,559 frames, one per 58 frames.
In rooms, a fixed reset comes at a cost.
With \scenebound{} inside, resets at 60 and 30 frames result in 0.1112 and 0.1301 on the benchmark versus 0.0805 for one state, in 0.407 and 0.456 versus 0.138 on the concatenated rooms, and in a point cloud with 2.4 to 2.5 times the accuracy error.
On the revisit sequences, the point clouds and KITTI sequences 04, 07 and 09, \scenebound{} makes zero returns to a stored submap, and its kept and dropped variants give equal errors.
On the eight drives with a return, the dropped variant reaches 59.1\,m compared with 62.1 for the kept one, and 0.494 compared with 0.503 on the 12 TUM RGB-D sequences outside development.
Per drive, returns lower the error on 00, 01 and 03, from 59.4, 105.7 and 21.3\,m with the closed submaps dropped to 36.4, 89.5 and 19.6\,m, and raise it on 02, 05, 06 and 08.
The gain of \scenebound{} over one state on these sets therefore comes from closing submaps, and stored submaps add their bound on the memory.
Evenly spaced closing ends as many submaps as the drift check on each sequence, at frames spaced evenly in time, and releases them as the dropped variant does.
On KITTI 02 to 10 it reaches 32.4 against 41.0\,m for the dropped variant of the drift check and is ahead on seven of the nine drives (\cref{app:driving}).
On the TUM RGB-D sequences it reaches 0.133 against 0.113 on the seven development sequences and 0.596 against 0.494 on the 12 others.
On the benchmark it closes states on two sequences and reaches 0.0798 against 0.0805 for one state, on the concatenated rooms it closes states in two rooms and reaches 0.140 against 0.138, and on the revisit sequences and the point clouds, where the drift check opens no submap, it keeps one state.

\begin{table*}[t]
\caption{Ways of closing a state, kept or dropped, with mean ATE per set and point-cloud accuracy on Neural RGB-D at 500 views\@.}
\label{tab:closing}
\centering
\footnotesize
\begin{fulltabular}{@{}lccccrc@{}}
\toprule
 & \multicolumn{5}{c}{Mean ATE (m) $\downarrow$} & Cloud acc.\ (m) $\downarrow$ \\
\cmidrule(lr){2-6}\cmidrule(lr){7-7}
Closing and keeping a state & Benchmark & Revisits & Concatenated & TUM RGB-D, 12 & KITTI, 02--10 & NRGBD \\
\midrule
One state & 0.0805 & \textbf{0.084} & \underline{0.138} & 0.713 & 136.8 & \textbf{0.0565} \\
Reset every 100 frames, TTT3R~\cite{chen2026ttt3r} gate, dropped & 0.1344 & 0.573 & 0.478 & \textbf{0.449} & 56.0 & 0.1214 \\
Count of writes, kept & 0.0806 & \textbf{0.084} & 0.174 & 0.642 & 53.0 & \textbf{0.0565} \\
\oursrow Drift check, kept (ours) & 0.0804 & \textbf{0.084} & 0.139 & 0.503 & 48.0 & \oursend \textbf{0.0565} \\
Drift check, dropped & 0.0802 & \textbf{0.084} & \textbf{0.137} & \underline{0.494} & 41.0 & \textbf{0.0565} \\
Reset every 60 frames, dropped & 0.1112 & 0.125 & 0.407 & 0.515 & 39.0 & 0.1351 \\
Reset every 30 frames, dropped & 0.1301 & 0.154 & 0.456 & 0.508 & \textbf{30.9} & 0.1397 \\
Radius of 1, kept & \underline{0.0800} & \underline{0.086} & 0.251 & 0.546 & 37.3 & \underline{0.1103} \\
Radius of 1, dropped & 0.0937 & 0.098 & 0.360 & 0.635 & 39.5 & \underline{0.1103} \\
Evenly spaced, count of the drift check, dropped & \textbf{0.0798} & \textbf{0.084} & 0.140 & 0.596 & \underline{32.4} & \textbf{0.0565} \\
\bottomrule
\end{fulltabular}
\end{table*}

We selected three alternatives to the drift check on the two development sets, the seven TUM RGB-D sequences and KITTI sequences 03, 04 and 10.
The selection score, fixed before any candidate ran, is the mean over the two sets of the ATE relative to the drift check.
The first alternative is a submap radius, and the search over six radii from 0.6 to 4 with and without the drift check selected a radius of 1 (score 0.748).
On the other sets, this radius is ahead on KITTI and on the two benchmark datasets with moving people, level on the benchmark mean and the revisit sequences, and behind on the two room datasets of the benchmark, the concatenated rooms, the point clouds and the 12 TUM RGB-D sequences outside development.
The second alternative measures the drift at the camera as the turn of the anchor times one plus the distance of the camera from the anchor in depth units.
It met the adoption criterion at a threshold of 0.05 (score 0.70) and yields 0.0830 on the benchmark, 75.1 versus 63.2\,m on the eight drives outside development and 0.420 versus 0.378 on 16 TUM RGB-D sequences.
The third alternative rolls a failed submap back to its state at the last check it passed.
On the development drives it reaches 17.3 versus 17.4\,m, an improvement of 1\% against the 10\% that the adoption criterion requires, and on the development sequences 0.117 versus 0.112.
Outside development, the rollback reaches 57.8 versus 63.2\,m on the eight drives and 0.448 versus 0.503 on the 12 sequences.
\scenebound{} retains the drift check measured at the anchor, and we made these choices with all of these sets in view.

The drift that a state measures at its anchor tracks the true error of the poses in its submap.
Over the keyframes of all submaps on the 19 TUM RGB-D sequences and 11 drives, the Spearman correlation of the true error relative to the anchor is 0.41 and 0.45 with the drift measured at the anchor, 0.40 and 0.62 with the frames since the anchor and 0.45 and 0.70 with the distance travelled. On KITTI the error grows with the distance from the anchor.

\subsection{Where the error of the submaps lies}
\label{app:graph}

\begin{table}[t]
\caption{Error of \scenebound{} and of one state over whole sequences and inside the pieces that \scenebound{} reads against one submap, each piece aligned to the truth on its own, ATE (m); the lower of the two runs is in bold.}
\label{tab:pieces}
\centering
\footnotesize
\begin{fulltabular}{@{}lrrrrr@{}}
\toprule
 & & \multicolumn{2}{c}{\scenebound{}} & \multicolumn{2}{c}{One state} \\
\cmidrule(lr){3-4}\cmidrule(lr){5-6}
Sequence & Pieces & Whole & Inside & Whole & Inside \\
\midrule
EuRoC MH 01 & 13 & 1.84 & \textbf{0.57} & \textbf{1.41} & 0.73 \\
EuRoC MH 02 & 20 & \textbf{1.91} & \textbf{0.46} & 3.62 & 0.67 \\
EuRoC MH 03 & 5 & 2.77 & \textbf{0.76} & \textbf{2.17} & 1.04 \\
EuRoC MH 04 & 28 & 4.12 & \textbf{0.31} & \textbf{3.54} & 0.45 \\
EuRoC MH 05 & 6 & 4.89 & \textbf{1.11} & \textbf{3.45} & 1.34 \\
EuRoC V1 01 & 14 & 0.84 & \textbf{0.26} & \textbf{0.82} & 0.32 \\
EuRoC V1 02 & 52 & 1.73 & \textbf{0.11} & \textbf{1.29} & 0.16 \\
EuRoC V1 03 & 13 & 1.29 & \textbf{0.28} & \textbf{1.00} & 0.34 \\
EuRoC V2 01 & 13 & 1.44 & \textbf{0.29} & \textbf{0.72} & 0.31 \\
EuRoC V2 02 & 9 & 1.05 & \textbf{0.26} & \textbf{0.75} & 0.34 \\
EuRoC V2 03 & 20 & 1.66 & 0.45 & \textbf{0.82} & \textbf{0.43} \\
EuRoC, mean & & 2.14 & \textbf{0.44} & \textbf{1.78} & 0.56 \\
\midrule
KITTI 00 & 88 & \textbf{36.4} & \textbf{0.8} & 188.0 & 8.1 \\
KITTI 01 & 25 & \textbf{89.5} & \textbf{6.0} & 529.9 & 29.7 \\
KITTI 02 & 92 & \textbf{159.2} & \textbf{2.7} & 275.6 & 13.7 \\
KITTI 03 & 16 & \textbf{19.6} & \textbf{2.2} & 56.9 & 7.8 \\
KITTI 04 & 5 & \textbf{3.5} & \textbf{1.1} & 12.3 & 5.4 \\
KITTI 05 & 44 & \textbf{31.2} & \textbf{0.8} & 115.8 & 8.1 \\
KITTI 06 & 21 & \textbf{79.5} & 25.5 & 136.1 & \textbf{22.6} \\
KITTI 07 & 22 & \textbf{5.6} & \textbf{0.5} & 73.2 & 4.7 \\
KITTI 08 & 85 & \textbf{52.0} & \textbf{3.6} & 232.6 & 10.2 \\
KITTI 09 & 35 & \textbf{52.2} & \textbf{1.1} & 191.2 & 9.8 \\
KITTI 10 & 19 & \textbf{29.2} & \textbf{2.1} & 137.2 & 7.5 \\
KITTI, mean & & \textbf{50.7} & \textbf{4.2} & 177.2 & 11.6 \\
\bottomrule
\end{fulltabular}
\end{table}

\begin{table*}[t]
\caption{A pose graph over the kept submaps, with loops confirmed by reads of the frozen model, on the 11 KITTI drives: ATE (m) and confirmed loops. Drives 01, 03, 04 and 10 revisit nothing.}
\label{tab:graph}
\centering
\footnotesize
\begin{fulltabular}{@{}lrrrrrrrrrrrc@{}}
\toprule
 & 00 & 01 & 02 & 03 & 04 & 05 & 06 & 07 & 08 & 09 & 10 & Mean \\
\midrule
\oursrow \scenebound{} (ours) & \textbf{36.4} & \textbf{89.5} & 159.2 & \textbf{19.6} & \textbf{3.5} & \textbf{31.2} & \textbf{79.5} & \textbf{5.6} & \textbf{52.0} & \textbf{52.2} & \textbf{29.2} & \oursend \textbf{50.7} \\
With the pose graph & 61.1 & \textbf{89.5} & \textbf{154.6} & 55.6 & 101.4 & 51.1 & 96.6 & 40.5 & 121.6 & 71.8 & \textbf{29.2} & 79.4 \\
Confirmed loops & 97 & 0 & 17 & 1 & 2 & 40 & 27 & 1 & 23 & 4 & 0 & 19.3 \\
\bottomrule
\end{fulltabular}
\end{table*}

To locate the error of a run with submaps, we split each sequence at the frames that open or return to a submap, so that each piece is read against a single submap, and align each piece to the ground truth on its own (\cref{tab:pieces}).
On EuRoC, the pieces of \scenebound{} have an error of 0.44\,m, below the 0.56\,m of one state over the same frames; the drift check thus makes each submap more accurate than the state it replaces, and the full flights reach 2.14\,m against 1.78\,m through the relative placement of 5 to 52 submaps.
On KITTI, the pieces of \scenebound{} have an error of 4.2\,m against 50.7\,m for the full drives, and the same frames of one state have 11.6\,m against 177.2\,m, lower than \scenebound{} on drive 06 alone; the remaining error lies in placing the submaps.

\scenebound{} keeps every closed submap with its state, so the submaps can serve as the nodes of a pose graph, as in submap-based SLAM~\cite{bosse2004simultaneous,elvira2019orbslamatlas,maggio2025vggtslam,deng2026vggtlong}.
We tested a pose graph that adds no constant beyond $M$.
Each node is the placement $\Gamma_i$ of submap $i$ in the coordinates of the first submap, and each edge $(a,b,Z)$ asks for $\Gamma_a^{-1}\Gamma_b=Z$.
When submap $b$ opens from submap $a$, the read that places it gives an edge whose $Z$ is the pose of its anchor in submap $a$.
At every keyframe write, the stored keyframe from another submap whose mean encoder token has the highest cosine similarity to the current frame names a candidate, excluding the submap that the current one opened from and those opened from it.
The frame is decoded against the stored state of the candidate, as in a return, and the read passes when its predicted pose lies at distance below $M$ from a keyframe in that submap.
Two consecutive keyframes whose reads pass against the same candidate submap, with relative motions in the two submaps within $M$ of each other, confirm a loop; each read adds an edge $Z=T_aT_b^{-1}$, with $T_a$ and $T_b$ the poses of the frame in the candidate and current submaps.
At the end of the video, the placements minimise the sum of squared edge residuals, with rotation in radians and translation in the depth unit of frame 0 and with the first submap fixed, and every output camera and point moves with the updated placement of its submap.
The graph changes no write, submap or return of the rule.

On the seven KITTI drives with loops, the pose graph raises the error on six and lowers it on drive 02, from 59.5 to 85.3\,m in the mean, and it runs at 15 to 17 frames per second against 22 to 23 (\cref{tab:graph}).
On drives 03 and 04, which revisit nothing, it confirms one and two loops, which must be false, and raises the error from 19.6 and 3.5\,m to 55.6 and 101.4\,m; on eight TUM RGB-D sequences it confirms no loop and leaves the error unchanged.
On drive 07, a single confirmed loop raises the error from 5.6 to 40.5\,m.
Reading a frame against a stored state places the frame near the keyframes of that state, which confirms false loops and yields relative poses too coarse to correct true loops.

\subsection{Cost}
\label{app:cost}

On KITTI 00, where the drift check opens 83 submaps and returns to kept ones four times, the peak GPU memory of \scenebound{} is 6,456\,MB, compared with 4,968\,MB in one state, 4,966\,MB for CUT3R and 5,779\,MB for TTT3R with reset.
A stored submap thus costs about 18\,MB, and \cref{tab:cost} lists the cost on the other drives and on TUM RGB-D.

\begin{table}[t]
\caption{Throughput, peak GPU memory and the most submaps held at once on the 19 TUM RGB-D sequences and KITTI sequences 02--10.}
\label{tab:cost}
\centering
\footnotesize
\begin{fulltabular}{@{}llccr@{}}
\toprule
Rule & Set & FPS $\uparrow$ & GPU (GB) $\downarrow$ & Submaps \\
\midrule
\oursrow \scenebound{} (ours) & TUM & 22.3 & 8.3 & \oursend 8 \\
TTT3R~\cite{chen2026ttt3r} with reset & TUM & 23.6 & 8.3 & 1 \\
\oursrow \scenebound{} (ours) & KITTI & 21.7 & 6.5 & \oursend 88 \\
TTT3R with reset & KITTI & 25.7 & 5.8 & 1 \\
Radius of 1 & KITTI & 23.6 & 7.4 & 237 \\
\bottomrule
\end{fulltabular}
\end{table}

\subsection{The KV cache}
\label{app:kv}

We run the cache rules in the released StreamVGGT checkpoint through the code and evaluator of RegVGGT (commit \texttt{3ec82dbe}, the 518-pixel crop of its pose evaluation, bfloat16) and score each predicted pose as the pose of the camera in the world.
The pose script of RegVGGT scores the inverse, the pose of the world in the camera.
Scored that way, our run at 1\% of keys reproduces the nine values of its Table~1 at 800, 900 and 1,000 frames within 0.5\%.
The rules run the cache unregulated, with all keys of all layers, and decide once per frame, after the camera head, whether the frame stays.
\scenebound{} stores the frame when, with the depth unit of frame 0, its predicted camera is at pose distance 0.3 or more from every kept frame.
Uniform sampling selects frame 0 and the same number of later frames at the spacing of \cref{prop:schedule}, and a fixed interval selects frames a fixed number apart.
A frame that leaves the cache is replaced by an empty tensor, which frees its storage.
The camera head of StreamVGGT keeps its own cache, four refinement steps of four blocks per frame; the fixed intervals, uniform sampling and \scenebound{} of the main tables prune it, so a frame that leaves the aggregator's cache leaves this cache too.
Pruning the camera head lowers the mean ATE of \scenebound{} from 0.0870 to 0.0831 and of a frame every 25, 50, 100 and 200 frames by 2\% to 5\%; with the camera head kept whole, as the published caches run, \scenebound{} is 15\% below RetrieveVGGT and ahead of all seven published caches; the radius sweep keeps the camera head whole.
We retuned nothing for this model.

\begin{table*}[t]
\caption{Signal-free rules in the KV cache of StreamVGGT on the benchmark, beside the scored keys of RegVGGT. Fixed intervals, uniform sampling and \scenebound{} prune the camera head, and the other rules keep it whole.}
\label{tab:cache-detail}
\centering
\footnotesize
\begin{fulltabular}{@{}lccccccr@{}}
\toprule
 & \multicolumn{5}{c}{ATE (m) $\downarrow$} & & \\
\cmidrule(lr){2-6}
Rule & TUM-dyn & Bonn & 7-Scenes & NRGBD & Mean & Rot.\ RPE (\textdegree) $\downarrow$ & GPU (GB) $\downarrow$ \\
\midrule
RegVGGT~\cite{mao2026regvggt}, 1\% of keys & 0.058 & 0.099 & 0.116 & 0.195 & 0.1128 & 0.86 & \textbf{8.3} \\
RegVGGT, 10\% of keys & 0.051 & 0.091 & 0.113 & 0.188 & 0.1067 & 0.73 & 11.9 \\
Random 1\% of keys\textsuperscript{\dag} & 0.051 & 0.110 & 0.119 & 0.208 & 0.1191 & 2.11 & \textbf{8.3} \\
Random 10\% of keys\textsuperscript{\dag} & 0.050 & 0.086 & 0.107 & 0.175 & 0.1006 & 0.69 & 11.9 \\
Uniformly spaced 10\% of keys\textsuperscript{\dag} & 0.051 & 0.090 & 0.109 & 0.173 & 0.1027 & 0.73 & 11.9 \\
Every 10 frames, regulated layers\textsuperscript{\dag} & 0.049 & 0.085 & 0.107 & 0.173 & 0.0995 & 0.70 & 12.0 \\
Every 10 frames\textsuperscript{\dag} & 0.038 & \textbf{0.071} & 0.107 & 0.123 & 0.0850 & \textbf{0.66} & 28.9 \\
Every 25 frames\textsuperscript{\dag} & \underline{0.033} & \textbf{0.071} & 0.102 & \underline{0.121} & \textbf{0.0828} & 0.72 & 16.0 \\
Every 50 frames\textsuperscript{\dag} & \underline{0.033} & 0.079 & \underline{0.101} & 0.124 & 0.0863 & 0.91 & 11.7 \\
Every 100 frames\textsuperscript{\dag} & 0.035 & 0.103 & \textbf{0.100} & 0.150 & 0.1002 & 1.19 & 9.6 \\
Every 200 frames\textsuperscript{\dag} & 0.041 & 0.149 & 0.210 & 0.196 & 0.1596 & 4.08 & \underline{8.5} \\
Uniform sampling\textsuperscript{\dag} & \textbf{0.032} & \underline{0.075} & 0.105 & \textbf{0.115} & 0.0841 & 0.71 & 13.3 \\
\oursrow \scenebound{} (ours) & \underline{0.033} & \textbf{0.071} & 0.104 & 0.122 & \underline{0.0831} & \underline{0.67} & \oursend 13.3 \\
\bottomrule
\end{fulltabular}

\smallskip
\raggedright\scriptsize
$^{\dag}$Constructed baseline.
\end{table*}

The peak GPU memory of a run includes the model and its activations, the cache and the input frames, which the evaluator of RegVGGT stores on the GPU in float32 at 2.35\,MB per frame for 378 by 518 pixels.
We report the peak less the stored input frames in all tables.
A least-squares fit over the runs of \scenebound{}, its uniform sampling and the interval of 50 frames yields about 255\,MB per kept frame, and the full cache fills the 80\,GB GPU at frame 287 on all 61 benchmark sequences.

On TUM-dynamics, Bonn, 7-Scenes and Neural RGB-D, \scenebound{} stores 116, 317, 690 and 310 frames, and the state of CUT3R takes 113, 314, 697 and 299 writes.
Over the 61 sequences, the two counts have a rank correlation of 0.99.
Fixed intervals of 25 and 50 frames fall within 0.009 of \scenebound{} on each dataset (\cref{tab:cache-detail}).
Every published cache is at or above \scenebound{} on each dataset except RetrieveVGGT on Neural RGB-D (0.102 against 0.122), with the largest gaps on Neural RGB-D and 7-Scenes, where GHOST reaches 0.302 and 0.248 compared with 0.122 and 0.104 (\cref{tab:benchmark}).

InfiniteVGGT (commit \texttt{7f9a5a26}) holds at most 1,200,000 tokens over the 24 global layers and shares them among the layers by the diversity of their keys.
OVGGT (commit \texttt{43304e42}) caps the cache at 200,000 tokens and 384 in the camera head, evicts by the size of the residual update of each token and its diversity, retains the first frame and protects up to three anchor frames.
FrameVGGT (commit \texttt{7e76cfbf}) stores two recent frames, 16 middle frames chosen to be complementary in key space and up to 64 anchors, with the memory setting of its released pose evaluation.
Since five benchmark sequences fill the released buffer, we size the token buffer of our runner for the frames that this setting retains.
XStreamVGGT (commit \texttt{3ec1d8d6}) keeps, in every global layer, the first frame, the current frame and the tokens between whose keys best match the queries of the frame, up to 2,048 tokens, and prunes the cache of its camera head to the same length.
RetrieveVGGT (commit \texttt{3b3923cb}) runs its released pose evaluation, in which each frame attends to frame 0 and to the earlier frames that segment sampling retrieves by the similarity of their mean key with its mean query, with its pose-aware memory off as released; it keeps the keys and values of every frame in host memory, 98.7\,MB per frame.
With its pose-aware memory and per-region compression on, at the defaults of its launcher, it reaches 0.1067 on the benchmark (0.053, 0.107, 0.127 and 0.113 per dataset), and \scenebound{} is lower on 49 of the 61 sequences.
This memory files frames in a grid of camera poses and halves the crowded regions every 200 frames, so it thins by the clock: over the nine passes of the revisit sequences it holds 22,710 frames after 6,690 in the first, 3.4 times as many, with a mean peak of 211\,GB of host memory, and it ends at 0.080 against 0.069 for \scenebound{}, which is lower on 13 of the 17 videos (\cref{app:looped}).
We run all four with the constructor of their own pose evaluation on the same checkpoint, crop and evaluator.
OVGGT evaluates the poses of InfiniteVGGT in its Table~S1, and our run agrees with these values within 4\% at all five lengths.
Our run of OVGGT is 20\% to 28\% above the same table, because its authors replaced the anchor code of the repository after the paper version and its loader takes 518 by 392 pixels, compared with 518 by 378 in our pose runs of the cache.
FrameVGGT reports poses on sequences of up to 300 frames.
It processes 0.5 frames per second on the benchmark and 0.24 to 0.42 on the revisit sequences, up to 10.6 hours for one video.

%% file: supp/D_results.tex
\section{Further results}
\label{app:results}

All runs use one NVIDIA H100 with 80\,GB or one H200 and PyTorch 2.8 or 2.9; runs of \scenebound{} on the two GPUs agree to five digits.

\subsection{Each dataset of the benchmark}
\label{app:benchmark}

\Cref{tab:benchmark} lists the mean ATE of every rule on each dataset of the long-sequence benchmark, with the relative pose error (RPE) between consecutive frames and the cost, and adds the constructed baselines and \scenebound{} with the camera head of StreamVGGT kept whole.
Over the 61 sequences, Info3R in full, the Info3R gate and AFG-Pose reach 1.10, 1.12 and 1.18 times the mean ATE of \scenebound{} in the state, with paired resampling intervals of 0.99 to 1.24, 1.04 to 1.21 and 1.08 to 1.30 as in \cref{tab:gates-detail}, and RetrieveVGGT reaches 1.23 times in the cache, 1.07 to 1.39.
On 7-Scenes and Neural RGB-D, AFG-Pose and the Info3R gate are 1\% and 3\% below \scenebound{} in the state through three of the 27 sequences, and in the cache RetrieveVGGT is lower on Neural RGB-D through one kitchen.
The relative pose error of \scenebound{} between consecutive frames is 1.7\,cm, against 1.5\,cm for CUT3R and 1.0\,cm for TTT3R with reset.

Thirteen Bonn sequences change the scene during recording as a person places, removes, moves or carries a box past the camera.
On these sequences, \scenebound{} writes 3 to 24 frames per sequence and keeps views from before the change, reaching 0.085 against 0.094 for the gate of Info3R, 0.106 for AFG-Pose and 0.096 for a full update every 50 frames, with a lower error than the two gates on 11 and 12 of the 13 sequences and than the interval on eight.

\input{tab/benchmark}

\Cref{tab:long-detail} lists the recurrent state on the revisit sequences, the long sequences, the held-out data and the point clouds with the constructed baselines beside the published rules.

\input{tab/long-detail}

\begin{table*}[t]
\caption{Our runs compared with the published values of the same rule.}
\label{tab:protocol}
\centering
\footnotesize
\begin{fulltabular}{@{}llrrl@{}}
\toprule
Rule & Set & Ours & Published & Source \\
\midrule
CUT3R~\cite{wang2025continuous} & TUM-dynamics, 1,000 frames & 0.161 & 0.173 & TTT3R, Table 5 \\
TTT3R~\cite{chen2026ttt3r} & TUM-dynamics, 1,000 frames & 0.115 & 0.106 & TTT3R, Table 5 \\
CUT3R with reset & TUM-dynamics, 1,000 frames & 0.121 & 0.126 & TTT3R, Table 2 \\
TTT3R with reset & TUM-dynamics, 1,000 frames & 0.096 & 0.093 & TTT3R, Table 5 \\
TTSA3R~\cite{zheng2026ttsa3r} & TUM-dynamics, 1,000 frames & 0.093 & 0.092 & TTSA3R, supplement \\
ReCal3R~\cite{li2026recal3r} & TUM-dynamics, 1,000 frames & 0.067 & 0.077 & ReCal3R, Table 2 \\
AFG-Pose~\cite{ren2026rethinking}, our implementation & TUM-dynamics, 1,000 frames & 0.055 & 0.054 & AFG, Table 11 \\
Info3R~\cite{baek2026info3r} in full, our implementation & TUM-dynamics, 1,000 frames & 0.056 & 0.061 & Info3R, Table II \\
CUT3R with reset & KITTI 02--10 & 85.9 & 80.8 & AFG, Table 1 \\
TTT3R with reset & KITTI 02--10 & 56.0 & 59.2 & AFG, Table 1 \\
AFG-Pose with reset, our implementation & KITTI 02--10 & 57.6 & 36.2 & AFG, Table 1 \\
Info3R in full, our implementation & KITTI 02--10 & 62.7 & 23.1 & Info3R, Table I \\
InfiniteVGGT~\cite{yuan2026infinitevggt} & TUM-dynamics, 1,000 frames & 0.067 & 0.066 & OVGGT, Table S1 \\
OVGGT~\cite{lu2026ovggt} & TUM-dynamics, 1,000 frames & 0.071 & 0.058 & OVGGT, Table S1 \\
RegVGGT~\cite{mao2026regvggt} at 1\% of keys, scored as its release scores & TUM-dynamics, 1,000 frames & 0.135 & 0.135 & RegVGGT, Table 1 \\
\bottomrule
\end{fulltabular}
\end{table*}

\subsection{Protocol and published values}
\label{app:protocol}

We run all rules of the recurrent state through one evaluator, the pose, depth and reconstruction scripts of the CUT3R release as packaged with ReCal3R, at 512 pixels and stride 1, with at most 1,000 frames on the long-sequence benchmark and at full length elsewhere.
The checkpoint, the preprocessing, the frames and the alignment of the metric are identical across rows, and inference is deterministic.
On the full-length TUM RGB-D sequences, all RGB frames are fed to the model and the frames with a true pose within 20\,ms are scored.
For KITTI, we use the grey left camera of the odometry benchmark, staged from a public mirror and checked against the official archive.

Rows with official code reproduce the published values within 13\% and OVGGT within 22\%, for the reasons in \cref{app:kv} (\cref{tab:protocol}).
\Cref{tab:published} sets \scenebound{} beside the values that published methods report on TUM-dynamics at 1,000 frames, trained ones included.

\input{tab/published}

FILT3R reports 0.057 on TUM-dynamics at 800 frames, our port obtains 0.074 at 1,000, and MeMix and AFG-Img report no value at 1,000 frames.
Since AFG-Pose and Info3R have no released code, we implement their equations, and our implementations agree with the papers on TUM-dynamics.
On KITTI, our AFG-Pose with the periodic reset of TTT3R is within 20\% of the paper on six of the 11 sequences, and our Info3R with the constants its paper specifies for driving is above the published value on 10 of the 11 sequences, with 72.3 against 30.5\,m in the mean.
The releases of CUT3R and TTT3R with reset, run on the same grey frames, lie within 7\% of the values that AFG reports for them on drives 02 to 10, so the camera changes little for the model, and the gap lies with the two implementations.
MeMix, the two AFG gates and BAT3R scale the token gain of TTT3R by a mask, a frame gain or further factors, and five of the nine gates therefore share that token signal.
RayMap3R resets its state at intervals of 100 frames and smooths the output translation, and the similarity alignment of its resets calls an unimported function, so in the release each segment of 100 frames keeps its own coordinate frame.
Our runs import this function; as released, RayMap3R reaches 0.3797 on the benchmark against 0.1699 with the function imported.
The gate of BAT3R needs an untrained network that the CUT3R checkpoint lacks, and its release therefore draws this network afresh in each process.
Our port fixes draw 0, which yields 0.069 on TUM-dynamics, compared with 0.157 and 0.143 for draws 1 and 2.
With full updates at a fixed interval after an opening window, we reproduce the published TTT3R baseline rows at intervals of 100 and 150 within 15\%; the published row at an interval of 50 is 0.150 and ours is 0.090.

\subsection{KITTI}
\label{app:driving}

\begin{table*}[t]
\caption{KITTI odometry at full length. Reported rows list the published values of each paper, obtained with the colour camera, and are left unmarked.}
\label{tab:kitti}
\centering
\footnotesize
\begin{fulltabular}{@{}lrrrrrrrrrrrrr@{}}
\toprule
 & \multicolumn{11}{c}{ATE (m) per sequence $\downarrow$} & \multicolumn{2}{c}{Mean $\downarrow$} \\
\cmidrule(lr){2-12}\cmidrule(lr){13-14}
Rule & 00 & 01 & 02 & 03 & 04 & 05 & 06 & 07 & 08 & 09 & 10 & All & 02--10 \\
\midrule
CUT3R~\cite{wang2025continuous} & 188.5 & 672.9 & 282.1 & 46.9 & 16.3 & 151.1 & 132.2 & 72.6 & 231.0 & 199.9 & 133.6 & 193.4 & 140.6 \\
TTT3R~\cite{chen2026ttt3r} & 189.4 & 556.0 & 259.7 & 51.3 & 9.2 & 115.8 & 134.9 & 68.9 & 261.2 & 180.6 & 119.5 & 177.0 & 133.5 \\
AFG-Pose\textsuperscript{\ddag}~\cite{ren2026rethinking} & 180.4 & 584.7 & 259.3 & 19.3 & 7.8 & 93.6 & 105.4 & 79.3 & 238.1 & 155.8 & 208.1 & 175.6 & 129.6 \\
Every 50 frames\textsuperscript{\dag} & 170.5 & 578.7 & 245.9 & 82.5 & 8.9 & 152.6 & 99.2 & 86.9 & 233.5 & 193.7 & 81.2 & 175.8 & 131.6 \\
\scenebound{}, one state & 188.0 & 529.9 & 275.6 & 56.9 & 12.3 & 115.8 & 136.1 & 73.2 & 232.6 & 191.2 & 137.2 & 177.2 & 136.8 \\
\midrule
CUT3R with reset & 189.4 & 123.5 & 271.3 & 18.5 & 7.8 & 88.9 & 59.2 & 25.8 & 152.2 & 124.9 & 24.2 & 98.7 & 85.9 \\
TTT3R with reset & 76.6 & 98.4 & 216.5 & 14.0 & 3.6 & 30.7 & 34.7 & 11.0 & 87.5 & 76.6 & 29.4 & 61.7 & 56.0 \\
TTT3R with reset every 50 frames & 70.4 & \textbf{89.4} & 121.8 & 11.0 & \underline{2.3} & 42.6 & 20.7 & 12.0 & 50.1 & 42.9 & 31.4 & 45.0 & 37.2 \\
TTT3R with reset every 30 frames & 79.4 & 107.5 & \textbf{91.5} & 11.4 & 2.9 & 48.9 & 28.2 & 14.3 & 50.3 & \underline{36.3} & 22.6 & 44.9 & 34.1 \\
AFG-Pose\textsuperscript{\ddag} with reset & 54.0 & 90.9 & 265.0 & 13.2 & 3.5 & 36.8 & 35.8 & 20.3 & 60.5 & 54.4 & 28.9 & 60.3 & 57.6 \\
Info3R\textsuperscript{\ddag}~\cite{baek2026info3r}, gate with reset & 69.1 & 124.4 & 173.7 & 16.3 & 3.6 & 51.4 & 45.4 & 18.9 & 78.8 & 65.3 & 32.0 & 61.7 & 53.9 \\
Info3R\textsuperscript{\ddag}, in full, driving constants & 68.0 & 162.6 & 143.1 & 14.6 & 5.3 & 87.9 & 44.6 & 25.0 & 110.0 & 102.9 & 31.5 & 72.3 & 62.7 \\
\midrule
\oursrow \scenebound{} (ours) & \textbf{36.4} & \underline{89.5} & 159.2 & 19.6 & 3.5 & 31.2 & 79.5 & \textbf{5.6} & 52.0 & 52.2 & 29.2 & 50.7 & \oursend 48.0 \\
\scenebound{}, reset every 60 frames & 55.6 & 104.1 & 164.2 & 9.7 & 3.2 & 28.1 & 25.2 & 10.1 & 38.1 & 52.7 & 20.0 & 46.5 & 39.0 \\
\scenebound{}, reset every 30 frames & 91.7 & 107.4 & 130.3 & \underline{7.6} & \underline{2.3} & 26.8 & \textbf{14.3} & 7.7 & \underline{37.3} & \textbf{35.5} & \underline{16.5} & \underline{43.4} & \textbf{30.9} \\
Radius of 1 & 82.5 & 117.5 & 132.4 & \textbf{7.2} & \textbf{1.6} & 49.1 & \underline{14.6} & 16.0 & \textbf{36.2} & 59.7 & 18.7 & 48.7 & 37.3 \\
Drift check, submaps dropped & 59.4 & 105.7 & 141.9 & 21.3 & 3.5 & \underline{23.8} & 45.3 & \textbf{5.6} & 46.7 & 52.2 & 28.3 & 48.5 & 41.0 \\
Evenly spaced, submaps dropped & \underline{44.9} & 94.0 & \underline{118.0} & 12.7 & 3.5 & \textbf{19.9} & 30.7 & \underline{6.8} & 38.1 & 48.3 & \textbf{13.1} & \textbf{39.1} & \underline{32.4} \\
Radius of 1 with a return under drift & 148.3 & 118.1 & 248.5 & \textbf{7.2} & \textbf{1.6} & 29.6 & 17.7 & 16.0 & 42.4 & 58.4 & 16.7 & 64.0 & 48.7 \\
\midrule
Info3R, reported & 30.0 & 97.2 & 86.6 & 9.1 & 6.3 & 23.6 & 11.2 & 7.6 & 30.0 & 22.9 & 10.8 & 30.5 & 23.1 \\
AFG-Pose, reported & 72.2 & 85.5 & 119.0 & 12.1 & 6.5 & 35.2 & 30.1 & 17.4 & 45.9 & 28.6 & 31.1 & 44.0 & 36.2 \\
TTT3R with reset, reported & 118.2 & 103.0 & 242.1 & 14.4 & 4.5 & 31.6 & 37.8 & 12.0 & 81.5 & 79.1 & 29.8 & 68.5 & 59.2 \\
CUT3R with reset, reported & 190.4 & 90.6 & 264.4 & 20.4 & 7.3 & 92.2 & 67.5 & 22.5 & 145.1 & 67.4 & 40.0 & 91.6 & 80.8 \\
LongStream~\cite{cheng2026longstream}, reported, trained & 92.5 & 46.0 & 134.7 & 3.8 & 1.9 & 84.7 & 23.1 & 14.9 & 62.1 & 85.6 & 21.5 & 51.9 & 48.0 \\
\bottomrule
\end{fulltabular}

\smallskip
\raggedright\scriptsize
$^{\dag}$Constructed baseline. $^{\ddag}$Our implementation of a method without released code.
\end{table*}

\Cref{tab:kitti} reports all rules on the 11 sequences.
Systems trained for long sequences, or equipped with a back end and loop closure, report lower errors on the 11 sequences, 25.4\,m for LoGeR~\cite{zhang2026loger}, 19.8 for Horizon-Stream~\cite{cheng2026horizonstream}, 18.3 for ABot-Recon~\cite{han2026revisiting} and 13.1 for AMB3R-SLAM~\cite{wang2026amb3rslam}.
Chunked systems on a frozen VGGT align chunks after the fact and report 26.4\,m for VGGT-Long~\cite{deng2026vggtlong}, 24.1 for LingBot-Map~\cite{chen2026lingbotmap} and 12.8 for CLoSeR~\cite{li2026closer}.

\subsection{TUM RGB-D and concatenated rooms}
\label{app:recordings}

\begin{table*}[t]
\caption{Full-length TUM RGB-D sequences, ATE (m) per sequence; dev marks the seven sequences that selected the threshold.}
\label{tab:recordings}
\centering
\footnotesize
\begin{fulltabular}{@{}lcccccccccc@{}}
\toprule
 & & & & Info3R\textsuperscript{\ddag} & \multicolumn{2}{c}{With reset} & Every & One & & Radius \\
\cmidrule(lr){6-7}
Sequence & CUT3R & TTT3R & AFG-Pose\textsuperscript{\ddag} & in full & CUT3R & TTT3R & 50\textsuperscript{\dag} & state & Ours & of 1 \\
\midrule
fr1 360 (dev) & 0.190 & 0.181 & 0.182 & 0.127 & 0.178 & 0.105 & 0.172 & 0.144 & \underline{0.103} & \textbf{0.065} \\
fr1 desk (dev) & 0.732 & 0.367 & 0.057 & 0.095 & 0.151 & 0.111 & \underline{0.054} & 0.095 & \textbf{0.050} & 0.076 \\
fr1 desk2 (dev) & 0.713 & 0.188 & \underline{0.119} & 0.271 & 0.405 & 0.133 & 0.132 & 0.155 & 0.120 & \textbf{0.101} \\
fr1 room (dev) & 0.995 & 0.623 & 0.461 & \textbf{0.162} & 0.380 & 0.235 & \underline{0.206} & 0.333 & 0.244 & 0.215 \\
fr2 desk (dev) & 1.580 & 1.606 & 0.065 & 0.172 & 0.632 & 0.329 & 0.060 & \underline{0.058} & \textbf{0.056} & 0.117 \\
fr2 xyz (dev) & 0.278 & 0.245 & 0.021 & 0.018 & 0.333 & 0.249 & 0.017 & \textbf{0.010} & \textbf{0.010} & \underline{0.011} \\
fr3 long office (dev) & 1.809 & 1.766 & \underline{0.118} & 0.280 & 0.258 & 0.278 & \textbf{0.103} & 0.168 & 0.202 & 0.173 \\
fr1 floor & 0.726 & 0.569 & 0.456 & 0.235 & 0.342 & 0.267 & 0.214 & \underline{0.106} & 0.109 & \textbf{0.086} \\
fr1 rpy & 0.051 & 0.061 & 0.057 & 0.055 & 0.056 & 0.054 & 0.060 & \underline{0.049} & \underline{0.049} & \textbf{0.041} \\
fr1 xyz & 0.147 & 0.090 & 0.030 & 0.023 & 0.088 & 0.081 & \underline{0.021} & \textbf{0.020} & \textbf{0.020} & \textbf{0.020} \\
fr2 360 hemisphere & 1.113 & 0.963 & 0.656 & 0.433 & 0.696 & \underline{0.419} & 0.748 & 0.852 & 0.825 & \textbf{0.353} \\
fr2 360 kidnap & 1.598 & 1.661 & 1.492 & \underline{1.036} & 1.385 & \textbf{0.500} & 1.144 & 1.617 & 1.610 & 1.316 \\
fr2 large no loop & 1.949 & 0.913 & 0.587 & \textbf{0.179} & 0.770 & 0.949 & 1.752 & 0.710 & \underline{0.438} & 0.685 \\
fr2 large with loop & 1.484 & 1.271 & 1.217 & 1.048 & 1.019 & 0.945 & 1.435 & 1.254 & \textbf{0.615} & \underline{0.863} \\
fr2 pioneer 360 & 1.795 & 0.947 & 0.905 & 0.564 & 0.555 & \textbf{0.360} & \underline{0.458} & 0.831 & 0.462 & 0.882 \\
fr2 pioneer slam & 1.930 & 1.358 & 1.276 & 0.953 & 1.668 & 0.900 & \textbf{0.588} & 1.073 & 1.003 & \underline{0.894} \\
fr2 pioneer slam2 & 1.823 & 1.097 & 1.478 & 0.479 & 0.762 & \underline{0.460} & 1.521 & 1.051 & \textbf{0.420} & 0.817 \\
fr2 pioneer slam3 & 1.979 & 1.825 & 0.975 & \textbf{0.283} & 0.557 & \underline{0.385} & 0.824 & 0.948 & 0.443 & 0.552 \\
fr2 rpy & 0.079 & 0.070 & 0.061 & 0.055 & 0.065 & 0.071 & 0.050 & \underline{0.042} & \underline{0.042} & \textbf{0.041} \\
Mean of the 12 others & 1.223 & 0.902 & 0.766 & \textbf{0.445} & 0.664 & \underline{0.449} & 0.735 & 0.713 & 0.503 & 0.546 \\
Geometric mean of the 12 & 0.742 & 0.564 & 0.442 & \textbf{0.261} & 0.423 & 0.313 & 0.382 & 0.359 & \underline{0.268} & 0.284 \\
Mean of the 19 & 1.104 & 0.832 & 0.538 & \textbf{0.340} & 0.542 & 0.360 & 0.503 & 0.501 & \underline{0.359} & 0.385 \\
\bottomrule
\end{fulltabular}

\smallskip
\raggedright\scriptsize
Ours is \scenebound{}; one state and a radius of 1 are its variants. $^{\dag}$Constructed baseline. $^{\ddag}$Our implementation of a method without released code.
\end{table*}

\Cref{tab:recordings} reports all rules on each of the 19 sequences.
The concatenated rooms are revisits along new paths.
Summed over the six rooms, the first sequences store 107 keyframes, 17.8 per sequence, and the 21 later sequences 401 with submaps, 19.1 per sequence.
Of the keyframes of later sequences, 4.0\% are within $M$ of a keyframe of an earlier sequence in true pose, and the median distance to the nearest earlier keyframe is 0.57 in true and 0.54 in predicted pose.
A later sequence therefore views the room from new poses, and the rule stores those views.

\subsection{Revisit sequences}
\label{app:looped}

\begin{figure*}[t]
\centering
\includegraphics[width=\linewidth]{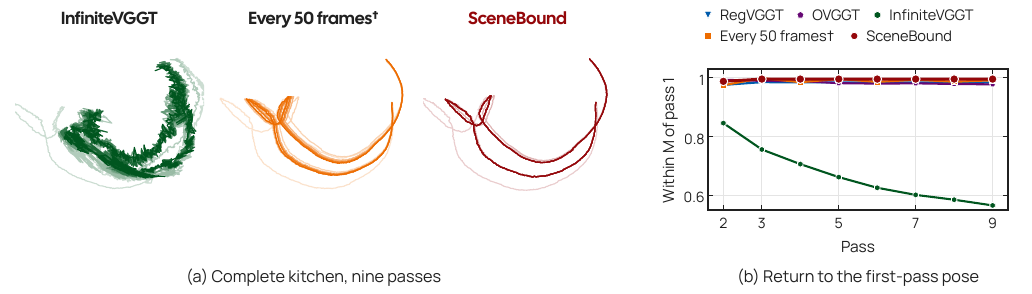}
\caption{Nine passes over the revisit sequences in the KV cache of StreamVGGT. (a) Predicted camera centres on the complete kitchen of Neural RGB-D from above, from the first pass (light) to the ninth (dark). (b) Frames of each later pass within $M$ of the pose predicted for the same image in pass 1, over the 17 sequences.}
\label{fig:revisits}
\end{figure*}

\input{tab/cache}

\begin{table*}[t]
\caption{All revisit sequences in the KV cache, with the frames that \scenebound{} retains after one and nine passes and the ATE (m) of each method; every 50 frames\textsuperscript{\dag} and \scenebound{} prune the camera head.}
\label{tab:looped-cache}
\centering
\footnotesize
\begin{fulltabular}{@{}lcrcccccc@{}}
\toprule
Video & Frames & Kept & RegVGGT, 10\% & InfiniteVGGT & OVGGT & FrameVGGT & Every 50\textsuperscript{\dag} & \scenebound{} \\
\midrule
NRGBD breakfast room & 9,000 & 43, 49 & 0.190 & 0.494 & 0.278 & \textbf{0.042} & \underline{0.050} & 0.215 \\
NRGBD complete kitchen & 9,000 & 49, 62 & 0.415 & 0.717 & \underline{0.201} & 0.219 & 0.206 & \textbf{0.180} \\
NRGBD green room & 9,000 & 25, 25 & 0.120 & 0.578 & 0.210 & 0.060 & \underline{0.052} & \textbf{0.043} \\
NRGBD grey-white room & 9,000 & 25, 29 & 0.460 & 1.033 & 0.494 & \textbf{0.146} & 0.154 & \underline{0.148} \\
NRGBD kitchen & 9,000 & 41, 41 & 0.307 & 0.707 & 0.322 & 0.170 & \underline{0.142} & \textbf{0.133} \\
NRGBD morning apartment & 8,280 & 28, 28 & 0.055 & 0.322 & 0.147 & 0.077 & \underline{0.045} & \textbf{0.036} \\
NRGBD staircase & 9,000 & 42, 42 & 0.150 & 0.327 & 0.147 & \underline{0.074} & 0.083 & \textbf{0.062} \\
NRGBD thin geometry & 3,555 & 18, 19 & 0.097 & 0.096 & 0.093 & \textbf{0.059} & 0.075 & \underline{0.070} \\
NRGBD whiteroom & 9,000 & 39, 39 & 0.127 & 0.452 & 0.205 & 0.085 & \underline{0.072} & \textbf{0.065} \\
TUM-dyn sitting halfsphere & 9,000 & 28, 28 & 0.097 & 0.181 & 0.104 & 0.085 & \underline{0.038} & \textbf{0.037} \\
TUM-dyn sitting rpy & 7,353 & 20, 20 & 0.033 & 0.058 & 0.044 & \underline{0.027} & \textbf{0.021} & \textbf{0.021} \\
TUM-dyn sitting static & 6,327 & 1, 1 & \underline{0.015} & 0.038 & 0.031 & \textbf{0.012} & 0.018 & \textbf{0.012} \\
TUM-dyn sitting xyz & 9,000 & 6, 6 & 0.038 & 0.092 & 0.073 & 0.031 & \underline{0.025} & \textbf{0.022} \\
TUM-dyn walking halfsphere & 9,000 & 28, 28 & 0.099 & 0.274 & 0.208 & 0.097 & \underline{0.060} & \textbf{0.043} \\
TUM-dyn walking rpy & 8,154 & 24, 24 & 0.060 & 0.147 & 0.140 & 0.049 & \underline{0.044} & \textbf{0.039} \\
TUM-dyn walking static & 6,660 & 1, 1 & 0.019 & 0.025 & 0.022 & \underline{0.016} & 0.018 & \textbf{0.015} \\
TUM-dyn walking xyz & 7,713 & 8, 8 & 0.066 & 0.152 & 0.102 & 0.058 & \underline{0.033} & \textbf{0.026} \\
Mean & & & 0.138 & 0.335 & 0.166 & 0.077 & \textbf{0.067} & \underline{0.069} \\
\bottomrule
\end{fulltabular}

\smallskip
\raggedright\scriptsize
$^{\dag}$Constructed baseline.
\end{table*}

On the 17 revisit sequences whose first pass is a benchmark sequence, one state writes 412 views in the first pass and 430 in all nine, and on 12 of them the eight further passes add no write.
For these sequences, uniform sampling of the same writes yields 0.088 and a constant gain of the same budget 0.082, compared with 0.078 for the rule, and on all 24 the rule reaches 0.084 against 0.124 for the best gate, and most of the lead of the rule over the gates is therefore a lead of its budget, which the baselines receive as an input and the rule finds online.
In the cache, \scenebound{} retains 426 frames in the first passes and 450 over all nine, no growth on 13 of the 17 sequences and at most 27\% (\cref{tab:cache,tab:looped-cache}).
Spreading the same 450 frames uniformly over the nine passes requires the final count and length in advance, and it holds 0.112 after the first pass against 0.080 for \scenebound{}, and 0.076 over all nine against 0.069, above \scenebound{} on 13 of the 17 sequences at the same memory and equal on the two static scenes where both keep the first frame alone.
Played once, the same sequences give 0.076 for uniform sampling of \scenebound{}'s count, so the count sets the error of a single pass, and placing keyframes by place lowers it by 10\% once the camera returns.
Each later pass reads the stored views because it receives their poses again (\cref{fig:revisits}): the predicted pose of a frame in passes 2 to 9 lies at a median distance of 0.019 from that of the same image in pass 1 under \scenebound{}, flat from the second pass, against 0.047 under RegVGGT at 10\% of keys, 0.078 under OVGGT and 0.210 under InfiniteVGGT, whose share within $M$ falls from 85\% in pass 2 to 57\% in pass 9 as it drops the views of the first passes.

\subsection{Held-out sets}
\label{app:heldout}

EuRoC~\cite{burri2016euroc} comprises the 11 sequences MH~01 to V2~03 of the left grey camera at 752 by 480 pixels and 20\,Hz, ETH3D SLAM~\cite{schops2019bad} the 61 training sequences with public ground truth, and the TUM set all TUM RGB-D sequences with public ground truth in the categories Structure vs.~Texture, Dynamic Objects and 3D Object Reconstruction that no earlier result of this work used, excluding the calibration sequences.
We froze the inputs of each set before any method ran, and each method ran once with the settings of \cref{sec:setup}.
\scenebound{} achieves a lower error than TTT3R with reset on six of the 11 EuRoC sequences, 40 of the 61 ETH3D sequences and 18 of the 19 TUM sequences, and a lower error than CUT3R and TTT3R on nine and six of 11, 50 and 50 of 61 and all 19.
Its submaps lower the error of one state on ETH3D and the TUM set and raise it on EuRoC (\cref{tab:long-detail}).

\begin{table*}[t]
\caption{ZipMap and STream3R on the 61 benchmark sequences, ATE (m) per dataset, RPE and cost: the frames written into the fast weights of ZipMap, summed over the sequences, and the peak GPU memory (GB) of STream3R.}
\label{tab:models-detail}
\centering
\footnotesize
\begin{fulltabular}{@{}llcccccccr@{}}
\toprule
 & & \multicolumn{5}{c}{ATE (m) $\downarrow$} & \multicolumn{2}{c}{RPE $\downarrow$} & \\
\cmidrule(lr){3-7}\cmidrule(lr){8-9}
Model & Rule & TUM-dyn & Bonn & 7-Scenes & NRGBD & Mean & Trans.\ (m) & Rot.\ (\textdegree) & Cost $\downarrow$ \\
\midrule
ZipMap~\cite{jin2026zipmap} & Every frame, as released & 0.205 & 0.355 & 0.675 & 1.439 & 0.5897 & 0.0514 & 7.294 & 51,256 \\
 & Every 5 frames\textsuperscript{\dag} & 0.130 & 0.200 & 0.367 & 0.880 & 0.3406 & 0.0448 & 2.815 & 10,262 \\
 & Every 10 frames\textsuperscript{\dag} & 0.062 & 0.112 & 0.161 & 0.502 & 0.1774 & 0.0391 & 1.396 & 5,137 \\
 & Every 25 frames\textsuperscript{\dag} & \underline{0.041} & \underline{0.080} & \underline{0.092} & 0.119 & 0.0841 & \textbf{0.0291} & \underline{0.854} & 2,065 \\
 & Every 50 frames\textsuperscript{\dag} & \textbf{0.040} & 0.086 & \textbf{0.087} & \underline{0.103} & \textbf{0.0827} & 0.0306 & \textbf{0.843} & 1,039 \\
 & Every 100 frames\textsuperscript{\dag} & 0.043 & 0.117 & 0.100 & \textbf{0.101} & 0.0998 & 0.0344 & 1.408 & 526 \\
 & Every 200 frames\textsuperscript{\dag} & 0.044 & 0.155 & 0.178 & 0.156 & 0.1473 & 0.0509 & 3.258 & 269 \\
 & Uniform sampling\textsuperscript{\dag} & \underline{0.041} & \textbf{0.079} & 0.095 & 0.111 & \underline{0.0837} & \underline{0.0293} & 0.858 & 1,437 \\
 & \cellcolor{ours}\scenebound{} (ours) & \cellcolor{ours}0.045 & \cellcolor{ours}0.081 & \cellcolor{ours}0.102 & \cellcolor{ours}0.127 & \cellcolor{ours}0.0890 & \cellcolor{ours}0.0295 & \cellcolor{ours}\textbf{0.843} & \cellcolor{ours}1,437 \\
\midrule
STream3R~\cite{lan2026stream3r} & Every 10 frames\textsuperscript{\dag} & 0.037 & 0.088 & 0.095 & 0.067 & 0.0800 & \textbf{0.0203} & \textbf{0.778} & 20.7 \\
 & Every 25 frames\textsuperscript{\dag} & 0.034 & \underline{0.086} & \underline{0.092} & 0.056 & \underline{0.0763} & 0.0221 & 0.981 & 12.7 \\
 & Every 50 frames\textsuperscript{\dag} & \textbf{0.032} & 0.091 & \underline{0.092} & \textbf{0.052} & 0.0776 & 0.0248 & 1.051 & 10.1 \\
 & Every 100 frames\textsuperscript{\dag} & \underline{0.033} & 0.117 & 0.104 & \textbf{0.052} & 0.0925 & 0.0274 & 1.506 & 8.8 \\
 & Every 200 frames\textsuperscript{\dag} & 0.036 & 0.167 & 0.205 & 0.069 & 0.1462 & 0.0507 & 3.648 & 8.1 \\
 & Uniform sampling\textsuperscript{\dag} & 0.034 & 0.089 & \textbf{0.090} & \underline{0.055} & 0.0769 & 0.0238 & 0.917 & 11.1 \\
 & \cellcolor{ours}\scenebound{} (ours) & \cellcolor{ours}\underline{0.033} & \cellcolor{ours}\textbf{0.081} & \cellcolor{ours}0.094 & \cellcolor{ours}0.058 & \cellcolor{ours}\textbf{0.0751} & \cellcolor{ours}\underline{0.0211} & \cellcolor{ours}\underline{0.793} & \cellcolor{ours}11.1 \\
\bottomrule
\end{fulltabular}

\smallskip
\raggedright\scriptsize
$^{\dag}$Constructed baseline.
\end{table*}

\subsection{ZipMap and STream3R}
\label{app:models}

ZipMap~\cite{jin2026zipmap} runs its streaming checkpoint (\texttt{checkpoint\_online.pt} of coast01/ZipMap at \texttt{71e449db}, source \texttt{e0f1f400}) with its online window of one frame, Muon-normalised updates of the fast weights and the camera of its MLP head, in the RegVGGT evaluator with the release's image loader.
We read one frame per call with the fast weights of every global block carried; with every frame kept this reproduces the release's online pass to $8\cdot10^{-7}$ in a behavioural check with the released model code.
A frame that \scenebound{} or a fixed interval skips returns the fast weights to their values before it, and under such a rule each frame's pose equals that of an online pass over the kept frames before it and the frame itself.
To measure the drift of the fast weights, frame 0 is read again after 100 to 500 frames with a forward pass through every block that applies the fast weights and leaves them unchanged.
STream3R~\cite{lan2026stream3r} runs its causal streaming session (\texttt{7bdca8cf}, weights \texttt{0af188b4}); a frame that leaves the cache leaves the keys and values of every global attention layer and of the camera head.
Its whole cache fills the 80\,GB GPU at frame 473, and the six sequences shorter than 473 frames complete.
Over the nine passes of the 17 revisit sequences, \scenebound{} keeps 404 frames after the first passes and 408 after all nine in ZipMap and 410 and 411 in STream3R, against 311 and 2,766 for every 50 frames in both.
\Cref{tab:models-detail} lists every rule per dataset.

\subsection{Frame rate}
\label{app:frame-rate}

\begin{table}[t]
\caption{The benchmark at four strides: frames written, summed over the 61 sequences, and mean ATE (m). The fixed time interval writes every 50, 25, 12 and 6 input frames.}
\label{tab:frame-rate}
\centering
\footnotesize
\begin{fulltabular}{@{}lcrrrcccc@{}}
\toprule
 & \multicolumn{4}{c}{Frames written} & \multicolumn{4}{c}{Mean ATE (m) $\downarrow$} \\
\cmidrule(lr){2-5}\cmidrule(lr){6-9}
Stride & 1 & 2 & 4 & 8 & 1 & 2 & 4 & 8 \\
\midrule
\multicolumn{9}{@{}l}{\emph{Recurrent state of CUT3R}} \\
\oursrow \scenebound{} (ours) & 1,423 & 1,379 & 1,325 & 1,220 & \textbf{0.080} & \textbf{0.081} & \textbf{0.080} & \oursend \underline{0.082} \\
Every 50 frames\textsuperscript{\dag} & 1,039 & 526 & 269 & 160 & \underline{0.084} & 0.096 & 0.147 & 0.187 \\
Fixed time interval\textsuperscript{\dag} & 1,039 & 1,039 & 1,089 & 1,089 & \underline{0.084} & \underline{0.084} & \underline{0.081} & \textbf{0.081} \\
AFG-Pose\textsuperscript{\ddag}~\cite{ren2026rethinking} & 1,390 & 935 & 743 & 694 & 0.095 & 0.099 & 0.103 & 0.107 \\
Info3R\textsuperscript{\ddag}, gate & 2,901 & 2,717 & 2,312 & 1,756 & 0.090 & 0.089 & 0.087 & 0.085 \\
\midrule
\multicolumn{9}{@{}l}{\emph{KV cache of StreamVGGT}} \\
\oursrow \scenebound{} (ours) & 1,433 & 1,395 & 1,322 & 1,216 & \textbf{0.083} & \textbf{0.082} & \textbf{0.083} & \oursend \textbf{0.082} \\
Every 50 frames\textsuperscript{\dag} & 1,039 & 526 & 269 & 160 & \underline{0.086} & 0.101 & 0.159 & 0.182 \\
Fixed time interval\textsuperscript{\dag} & 1,039 & 1,039 & 1,089 & 1,089 & \underline{0.086} & \underline{0.086} & \underline{0.086} & \underline{0.086} \\
\midrule
\multicolumn{9}{@{}l}{\emph{KV cache of STream3R}} \\
\oursrow \scenebound{} (ours) & 1,435 & 1,392 & 1,329 & 1,235 & \textbf{0.075} & \textbf{0.076} & \textbf{0.079} & \oursend \textbf{0.078} \\
Every 50 frames\textsuperscript{\dag} & 1,039 & 526 & 269 & 160 & \underline{0.078} & \underline{0.092} & \underline{0.146} & \underline{0.180} \\
\midrule
\multicolumn{9}{@{}l}{\emph{Fast weights of ZipMap}} \\
\oursrow \scenebound{} (ours) & 1,437 & 1,394 & 1,326 & 1,211 & \underline{0.089} & \textbf{0.086} & \textbf{0.086} & \oursend \textbf{0.084} \\
Every 50 frames\textsuperscript{\dag} & 1,039 & 526 & 269 & 160 & \textbf{0.083} & \underline{0.100} & \underline{0.146} & \underline{0.215} \\
\bottomrule
\end{fulltabular}

\smallskip
\raggedright\scriptsize
$^{\dag}$Constructed baseline. $^{\ddag}$Our implementation of a method without released code.
\end{table}

We feed the first 1,000 frames of each benchmark sequence to all four models at every first, second, fourth and eighth frame, so that the camera visits the same poses in 1,000, 500, 250 and 125 frames, and every constant stays as it is.
\Cref{tab:frame-rate} lists the frames written and the mean ATE of each rule, and \cref{fig:budget}f plots the error relative to every frame.
At every frame, the runs give the values of \cref{tab:benchmark} for \scenebound{}, AFG-Pose and the gate of Info3R.
In all four models, \scenebound{} writes 14\% to 16\% fewer frames at every eighth frame, since a coarser sampling of the same path offers the keyframe test fewer poses, and its error changes by $-6\%$ to $+3\%$.
Per dataset, the cache of StreamVGGT holds \scenebound{} within 0.003 of its error at every frame on each of the four at every stride, and in the state every rule rises on TUM-dynamics, whose scenes hold walking people, at every eighth frame, \scenebound{} from 0.048 to 0.073 and AFG-Pose from 0.055 to 0.088, and \scenebound{} stays within 0.004 on the other three datasets.
A fixed interval of 50 frames writes in proportion to the frames fed, and at every eighth frame it writes about three frames per sequence and has 2.1 to 2.6 times its error at every frame in the four models.
The gates read the change from the previous frame, which grows with the step: AFG-Pose writes half its budget at every eighth frame and its error rises by 12\%, and the gate of Info3R writes 39\% less and its error falls by 6\%, toward the minimum of the curve in \cref{fig:budget}b.
A fixed time interval writes every 25, 12 and 6 input frames at strides 2, 4 and 8, every 48 to 50 frames of video, and holds its count at 1,039 to 1,089.
In the state it reaches 0.0836, 0.0813 and 0.0811 against 0.0812, 0.0800 and 0.0821 for \scenebound{}, and in the cache 0.0865, 0.0859 and 0.0855 against 0.0823, 0.0827 and 0.0819, keeping 1,039, 1,089 and 1,089 frames against 1,395, 1,322 and 1,216.
Its point clouds at every eighth frame reach 0.027 and 0.079\,m in the state and 0.044 and 0.073\,m in the cache on 7-Scenes and Neural RGB-D, within 6\% of \scenebound{}.
A fixed time interval thus follows the frame rate as \scenebound{} does; it reads the frame rate of the video, and on the revisit sequences it writes on every pass (\cref{tab:cache}).

\subsection{Replica and video depth}
\label{app:replica}

\begin{table*}[t]
\caption{Replica, eight scenes of 2,000 frames held out until all constants were fixed, ATE (m) per scene, RPE and budget.}
\label{tab:replica}
\centering
\footnotesize
\begin{fulltabular}{@{}lcccccccccccr@{}}
\toprule
 & \multicolumn{9}{c}{ATE (m) $\downarrow$} & \multicolumn{2}{c}{RPE $\downarrow$} & \\
\cmidrule(lr){2-10}\cmidrule(lr){11-12}
Rule & room0 & room1 & room2 & office0 & office1 & office2 & office3 & office4 & Mean & Trans.\ (m) & Rot.\ (\textdegree) & Budget $\downarrow$ \\
\midrule
CUT3R~\cite{wang2025continuous} & 1.023 & 0.926 & 0.905 & 0.819 & 0.423 & 1.189 & 1.387 & 1.074 & 0.9682 & 0.0162 & 2.464 & 15,992 \\
TTT3R~\cite{chen2026ttt3r} & 1.069 & 0.622 & 0.721 & 0.651 & 0.332 & 0.930 & 1.077 & 0.825 & 0.7784 & 0.0157 & 1.569 & 4,967 \\
MeMix~\cite{dong2026memix} & 1.071 & 0.538 & 0.712 & 0.661 & 0.354 & 1.020 & 0.631 & 0.902 & 0.7362 & 0.0168 & 1.701 & 4,534 \\
TTSA3R~\cite{zheng2026ttsa3r} & 0.580 & 0.525 & 0.617 & 0.529 & 0.311 & 0.651 & 0.620 & 0.578 & 0.5514 & 0.0154 & 1.111 & 3,076 \\
FILT3R~\cite{jin2026filt3r} & 0.542 & 0.474 & 0.856 & 0.481 & 0.325 & 0.573 & 0.502 & 0.442 & 0.5244 & 0.0166 & 1.995 & 2,509 \\
ReCal3R~\cite{li2026recal3r} & 0.528 & 0.444 & 0.517 & 0.792 & 0.307 & 0.557 & 0.362 & 0.402 & 0.4885 & 0.0140 & 0.997 & 1,805 \\
AFG-Img\textsuperscript{\ddag}~\cite{ren2026rethinking} & 0.353 & 0.409 & 0.373 & 0.851 & 0.294 & 0.468 & 0.311 & 0.451 & 0.4387 & 0.0145 & 0.783 & 1,399 \\
AFG-Pose\textsuperscript{\ddag}~\cite{ren2026rethinking} & 0.168 & 0.261 & 0.160 & 0.246 & 0.260 & 0.200 & 0.103 & 0.251 & 0.2062 & 0.0142 & 0.371 & 393 \\
Info3R\textsuperscript{\ddag}~\cite{baek2026info3r}, gate & \underline{0.132} & 0.245 & \underline{0.113} & 0.182 & 0.212 & \textbf{0.102} & \underline{0.073} & \textbf{0.160} & \underline{0.1524} & 0.0157 & 1.076 & \textbf{283} \\
Info3R\textsuperscript{\ddag}, in full & \underline{0.132} & 0.277 & \underline{0.113} & 0.182 & 0.212 & \textbf{0.102} & \underline{0.073} & \textbf{0.160} & 0.1564 & 0.0158 & 1.078 & \textbf{283} \\
CUT3R with reset & 0.705 & 0.488 & 0.845 & 0.564 & 0.399 & 1.000 & 0.611 & 1.098 & 0.7138 & 0.0149 & 0.821 & 15,992 \\
TTT3R with reset & 0.346 & 0.344 & 0.420 & 0.382 & 0.363 & 0.380 & 0.274 & 0.319 & 0.3534 & 0.0130 & 0.464 & 5,781 \\
Every 50 frames\textsuperscript{\dag} & \textbf{0.130} & \underline{0.196} & \textbf{0.111} & \textbf{0.151} & 0.204 & \underline{0.123} & \textbf{0.060} & \underline{0.163} & \textbf{0.1422} & 0.0152 & 0.513 & \underline{320} \\
\scenebound{}, one state & 0.159 & 0.275 & 0.164 & \underline{0.160} & \underline{0.195} & 0.241 & 0.088 & 0.188 & 0.1837 & 0.0149 & 0.462 & 604 \\
\oursrow \scenebound{} (ours) & 0.161 & 0.246 & 0.188 & 0.162 & \textbf{0.176} & 0.241 & 0.088 & 0.188 & 0.1813 & 0.0172 & 0.677 & \oursend 612 \\
\scenebound{}, reset every 60 frames & 0.177 & \textbf{0.162} & 0.256 & 0.256 & 0.311 & 0.208 & 0.135 & 0.192 & 0.2120 & \underline{0.0090} & \textbf{0.243} & 751 \\
\scenebound{}, reset every 30 frames & 0.369 & 0.338 & 0.298 & 0.489 & 0.342 & 0.435 & 0.265 & 0.377 & 0.3641 & \textbf{0.0088} & \underline{0.280} & 866 \\
\bottomrule
\end{fulltabular}

\smallskip
\raggedright\scriptsize
$^{\dag}$Constructed baseline. $^{\ddag}$Our implementation of a method without released code.
\end{table*}

On Replica, a full update every 50 frames achieves the lowest mean, 0.142, followed by the gate of Info3R, 0.152, and \scenebound{} obtains 0.181 (\cref{tab:replica}).
Its error is lower than that of CUT3R, six gates and both reset baselines on all eight scenes, than that of AFG-Pose on six, than that of the gate of Info3R on two and than that of the fixed interval on one.
\scenebound{} opens nine submaps beyond the first on the eight scenes and is 1.3\% below one state.

Video depth over the first 500 frames depends little on the write rule, with an absolute relative error after scale alignment of 0.112 to 0.127 for the seven gates on TUM-dynamics and 0.056 to 0.061 on Bonn, compared with 0.116 and 0.057 for \scenebound{}.

\subsection{Point clouds}
\label{app:points}

\begin{table*}[t]
\caption{Point clouds fused from the first 1,000 frames of the benchmark sequences of 7-Scenes and Neural RGB-D and from every eighth of these frames: accuracy and completion errors (m) and normal consistency.}
\label{tab:recon}
\centering
\footnotesize
\begin{fulltabular}{@{}lcccccc@{}}
\toprule
 & \multicolumn{3}{c}{7-Scenes, 18} & \multicolumn{3}{c}{Neural RGB-D, 9} \\
\cmidrule(lr){2-4}\cmidrule(lr){5-7}
Rule & Acc.\ $\downarrow$ & Comp.\ $\downarrow$ & NC $\uparrow$ & Acc.\ $\downarrow$ & Comp.\ $\downarrow$ & NC $\uparrow$ \\
\midrule
\multicolumn{7}{@{}l}{\emph{Recurrent state of CUT3R, 1,000 frames}} \\
CUT3R~\cite{wang2025continuous} & 0.2569 & 0.1088 & 0.536 & 0.3965 & 0.2103 & 0.549 \\
TTT3R~\cite{chen2026ttt3r} & 0.1439 & 0.0545 & 0.563 & 0.2713 & 0.1143 & 0.578 \\
TTT3R with reset & 0.0851 & 0.0359 & 0.609 & 0.1385 & 0.0502 & 0.645 \\
AFG-Pose\textsuperscript{\ddag}~\cite{ren2026rethinking} & 0.0289 & 0.0228 & 0.619 & 0.0847 & \underline{0.0296} & 0.689 \\
Info3R\textsuperscript{\ddag}~\cite{baek2026info3r}, gate & 0.0255 & \underline{0.0218} & 0.626 & 0.0852 & 0.0344 & 0.671 \\
Info3R\textsuperscript{\ddag}, in full & 0.0255 & \underline{0.0218} & 0.626 & \underline{0.0836} & 0.0341 & 0.677 \\
Every 50 frames\textsuperscript{\dag} & \textbf{0.0238} & \textbf{0.0211} & \textbf{0.630} & \textbf{0.0769} & \textbf{0.0284} & \textbf{0.701} \\
\scenebound{}, one state & \underline{0.0254} & 0.0226 & \underline{0.629} & 0.0863 & 0.0338 & \underline{0.693} \\
\oursrow \scenebound{} (ours) & 0.0258 & 0.0226 & \underline{0.629} & 0.0863 & 0.0338 & \oursend \underline{0.693} \\
\midrule
\multicolumn{7}{@{}l}{\emph{KV cache of StreamVGGT, 1,000 frames}} \\
RegVGGT~\cite{mao2026regvggt}, 10\% of keys & 0.0457 & \underline{0.0243} & 0.600 & 0.1100 & 0.0427 & 0.675 \\
InfiniteVGGT~\cite{yuan2026infinitevggt} & 0.0769 & 0.0413 & 0.610 & 0.0914 & 0.0484 & 0.694 \\
OVGGT~\cite{lu2026ovggt} & 0.0541 & 0.0259 & 0.605 & \underline{0.0568} & \textbf{0.0309} & 0.697 \\
RetrieveVGGT~\cite{zou2026attention} & 0.0470 & \textbf{0.0227} & 0.616 & \textbf{0.0542} & \underline{0.0339} & 0.722 \\
Every 50 frames\textsuperscript{\dag} & \underline{0.0433} & 0.0244 & 0.643 & 0.0735 & 0.0423 & \textbf{0.745} \\
Uniform sampling\textsuperscript{\dag} & \textbf{0.0428} & 0.0250 & \textbf{0.649} & 0.0691 & 0.0412 & 0.729 \\
\oursrow \scenebound{} (ours) & 0.0445 & 0.0253 & \underline{0.648} & 0.0721 & 0.0417 & \oursend \underline{0.730} \\
\midrule
\multicolumn{7}{@{}l}{\emph{Recurrent state of CUT3R, every eighth frame}} \\
CUT3R~\cite{wang2025continuous} & 0.0564 & 0.0328 & 0.626 & 0.1375 & 0.0649 & 0.657 \\
TTT3R~\cite{chen2026ttt3r} & 0.0317 & 0.0258 & 0.637 & 0.0884 & 0.0350 & 0.702 \\
TTT3R with reset & 0.0436 & 0.0283 & 0.638 & 0.0928 & 0.0354 & \underline{0.705} \\
AFG-Pose\textsuperscript{\ddag}~\cite{ren2026rethinking} & 0.0308 & 0.0238 & 0.639 & 0.0868 & \underline{0.0333} & 0.704 \\
Info3R\textsuperscript{\ddag}~\cite{baek2026info3r}, gate & \underline{0.0269} & \textbf{0.0232} & \underline{0.645} & 0.0854 & 0.0356 & 0.703 \\
Info3R\textsuperscript{\ddag}, in full & \underline{0.0269} & \textbf{0.0232} & \underline{0.645} & \underline{0.0840} & 0.0353 & 0.704 \\
Every 50 frames\textsuperscript{\dag} & 0.1063 & 0.0357 & 0.625 & 0.0974 & 0.0351 & 0.695 \\
\scenebound{}, one state & \textbf{0.0251} & \underline{0.0235} & \textbf{0.649} & \textbf{0.0816} & \textbf{0.0315} & \textbf{0.713} \\
\oursrow \scenebound{} (ours) & \textbf{0.0251} & \underline{0.0235} & \textbf{0.649} & \textbf{0.0816} & \textbf{0.0315} & \oursend \textbf{0.713} \\
\midrule
\multicolumn{7}{@{}l}{\emph{KV cache of StreamVGGT, every eighth frame}} \\
RegVGGT~\cite{mao2026regvggt}, 10\% of keys & \underline{0.0434} & \textbf{0.0253} & 0.635 & 0.0855 & 0.0435 & 0.698 \\
RetrieveVGGT~\cite{zou2026attention} & 0.0444 & \underline{0.0255} & 0.648 & 0.0840 & 0.0492 & \underline{0.720} \\
Every 50 frames\textsuperscript{\dag} & 0.1163 & 0.0386 & 0.626 & 0.0817 & \underline{0.0428} & 0.705 \\
Uniform sampling\textsuperscript{\dag} & \textbf{0.0432} & 0.0269 & \textbf{0.662} & \textbf{0.0703} & \textbf{0.0427} & \textbf{0.741} \\
\oursrow \scenebound{} (ours) & 0.0452 & 0.0268 & \underline{0.661} & \underline{0.0733} & 0.0431 & \oursend \textbf{0.741} \\
\bottomrule
\end{fulltabular}

\smallskip
\raggedright\scriptsize
$^{\dag}$Constructed baseline. $^{\ddag}$Our implementation of a method without released code.
\end{table*}

The reconstruction evaluator of the CUT3R release, as ReCal3R packages it, aligns the fused points of all views to the depth of the dataset by scale, median depth and ICP and scores the central 224 by 224 pixels of each view, at 512 by 384 pixels in the state.
It encodes all views in one batch, which fills an 80\,GB GPU at 1,000 views, so we encode them 32 at a time, which leaves every token unchanged, and keep the maps and views of all frames on the host until they are scored.
At 1,000 frames, \scenebound{} reaches an accuracy error of 0.026 and 0.086\,m on 7-Scenes and Neural RGB-D, where CUT3R has 10 and 4.6 times this error and TTT3R 5.6 and 3.1 times (\cref{tab:recon}).
A full update every 50 frames, which writes 340 and 167 frames on the two datasets against 696 and 297 for \scenebound{}, achieves the lowest errors, 0.024 and 0.077\,m; the gates of Info3R lie between it and \scenebound{} on both datasets and that of AFG-Pose on Neural RGB-D, and on 7-Scenes AFG-Pose lies 12\% above \scenebound{}.
In the KV cache of StreamVGGT, the views of the same evaluator at 518 by 392 pixels give each rule the world points of its point head.
On 7-Scenes, the rules that keep whole frames give the most accurate clouds and the highest normal consistency, 0.043 to 0.045 for \scenebound{}, every 50 frames and uniform sampling at the count of \scenebound{}, against 0.046 to 0.077 for RegVGGT, RetrieveVGGT, OVGGT and InfiniteVGGT.
On Neural RGB-D, RetrieveVGGT and OVGGT, which read tokens of every earlier frame, reach 0.054 and 0.057 against 0.072 for \scenebound{}, and RegVGGT and InfiniteVGGT 0.110 and 0.091.
In the cache, the points of a frame thus depend on the tokens it reads more than on the trajectory, which \scenebound{} sets with 2.8\% of the frames (\cref{tab:cache}).
From 500 to 1,000 views, the accuracy error of \scenebound{} in the state grows by 8\% and 53\% and that of TTT3R doubles; at 200 and 500 views, \scenebound{} is within 3\% of a full update every 50 frames, CUT3R has 2.5 to 7.4 times its error, and on 7-Scenes at 200 views it achieves 0.0198, where the cache models Point3R, InfiniteVGGT and OVGGT report 0.041, 0.046 and 0.024.
At every eighth of the first 1,000 frames, the input of the frame-rate test of \cref{sec:models}, \scenebound{} in the state achieves the lowest accuracy error of the nine rules on both datasets, 0.025 and 0.082\,m, 3\% and 5\% below its error at 1,000 frames, with 33 and 29 keyframes per sequence.
A full update every 50 frames, the lowest at 1,000 frames, writes once every 400 frames of the video and reaches 0.106 and 0.097\,m as its pose error grows (\cref{fig:budget}f).
In the cache, \scenebound{} keeps 32 and 30 frames per sequence and reaches 0.045 and 0.073\,m, within 5\% of uniform sampling at its count, which needs the count in advance; every 50 frames reaches 2.6 and 1.1 times \scenebound{}, and RetrieveVGGT, at 0.054 on Neural RGB-D with 1,000 frames, reaches 0.084 there.

\subsection{The assumptions beside their measurements}
\label{app:assumptions}

\begin{figure*}[t]
\centering
\includegraphics[width=\linewidth]{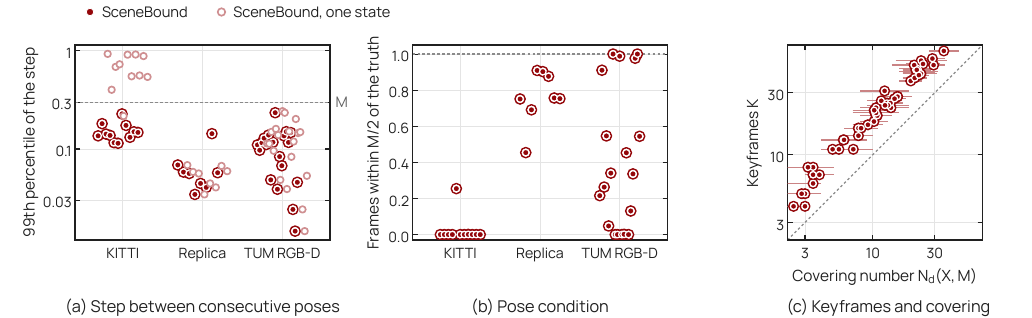}
\caption{Measured assumptions of the theorems. (a) The 99th percentile of the step between consecutive poses per sequence. (b) Share of frames within half a radius of the truth. (c) Keyframes versus the covering number.}
\label{fig:assumptions}
\end{figure*}

We measure each assumption of the theorems on the stored runs (\cref{fig:assumptions}).
\begin{itemize}
  \item A read-out that drifts in a fixed direction (\cref{thm:budget,prop:drift}). The shift of frame 0 grows with exponent 1.2 in the budget and stays within a factor of 2.4 per unit budget across eight rules (\cref{app:increments}).
  \item Consecutive poses less than $\delta<M$ apart (\cref{thm:submaps}). On each stream, 99.3\% to 100\% of the steps fall below $M$, and one step exceeds $M$ on 13 of the 38 streams.
  \item Covering numbers of the predicted poses (\cref{thm:keyframes}). In the state, $1{,}421\le N_d(X,M/2)\le2{,}231$, and in the cache, $571\le N_d(X,M)\le965$ compared with 1,433 keyframes, a ratio of 1.49 to 2.51.
  \item A place receives the same predicted pose on a return (\cref{thm:keyframes}). In the cache, 99.5\% of later-pass frames are within $M$ of their first-pass pose (\cref{fig:revisits}).
  \item Predicted poses within $\varepsilon<M/2$ of the truth (\cref{prop:true-poses}). This holds for 72\% of Replica frames, for nine in ten frames on four of the 19 TUM RGB-D sequences and on no KITTI drive.
  \item The amount stays fixed as the video grows. The revisit sequences take 678 writes in one pass and 698 in nine.
\end{itemize}
Of the 49,823 frames that \scenebound{} skips on the benchmark, 15.4\% are nearest to a keyframe other than the latest one, 33\% on TUM-dynamics and 2.6\% on Neural RGB-D.
The benchmark thus tests exploration, and the revisit sequences test the returns.
On the revisit sequences, the covering number of the predicted poses at half the radius grows by 23\% from one pass to nine, and the number of writes by 3\%.
In one state, submaps lower the median distance between predicted and true pose on 13 of the 19 TUM RGB-D sequences and on all 11 KITTI drives, where the median over the drives falls from 9.6 to 1.4.

%% file: tab/benchmark.tex
\begin{table*}[t]
\caption{Camera pose on the long-sequence benchmark of 61 sequences, with the cost of each rule: the update budget in the state and the GPU memory, less the stored input frames, in the cache.}
\label{tab:benchmark}
\centering
\footnotesize
\begin{fulltabular}{@{}lcccccccr@{\hspace{3.5pt}}}
\toprule
 & \multicolumn{5}{c}{ATE (m) $\downarrow$} & \multicolumn{2}{c}{RPE $\downarrow$} & \\
\cmidrule(lr){2-6}\cmidrule(lr){7-8}
Method & TUM-dyn & Bonn & 7-Scenes & NRGBD & Mean & Trans.\ (m) & Rot.\ (\textdegree) & Cost $\downarrow$ \\
\midrule
\multicolumn{8}{@{}l}{\emph{Recurrent state of CUT3R}} & \emph{budget} \\
CUT3R~\cite{wang2025continuous} & 0.161 & 0.322 & 0.655 & 1.156 & 0.5224 & 0.0154 & 1.231 & 51,195 \\
MeMix~\cite{dong2026memix} & 0.102 & 0.264 & 0.393 & 0.604 & 0.3312 & 0.0175 & 0.801 & 14,432 \\
TTT3R~\cite{chen2026ttt3r} & 0.115 & 0.269 & 0.410 & 0.457 & 0.3182 & 0.0179 & 0.897 & 15,716 \\
TTSA3R~\cite{zheng2026ttsa3r} & 0.093 & 0.220 & 0.315 & 0.289 & 0.2414 & 0.0181 & 0.677 & 11,322 \\
RayMap3R~\cite{wang2026raymap3r} & 0.131 & 0.127 & 0.206 & 0.257 & 0.1699 & \textbf{0.0100} & 0.673 & 9,566 \\
BAT3R~\cite{li2026bat3r} & 0.069 & 0.155 & 0.170 & 0.262 & 0.1642 & 0.0183 & 0.759 & 6,920 \\
FILT3R~\cite{jin2026filt3r} & 0.074 & 0.143 & 0.190 & 0.245 & 0.1629 & 0.0167 & 0.630 & 7,790 \\
AFG-Img\textsuperscript{\ddag}~\cite{ren2026rethinking} & 0.083 & 0.165 & 0.163 & 0.162 & 0.1534 & 0.0170 & 0.639 & 6,798 \\
ReCal3R~\cite{li2026recal3r} & 0.067 & 0.110 & 0.146 & 0.185 & 0.1258 & 0.0159 & \underline{0.625} & 5,794 \\
AFG-Pose\textsuperscript{\ddag}~\cite{ren2026rethinking} & \underline{0.055} & 0.107 & \underline{0.094} & 0.098 & 0.0950 & 0.0175 & 0.641 & \underline{1,390} \\
Info3R\textsuperscript{\ddag}~\cite{baek2026info3r}, gate & 0.060 & 0.094 & 0.101 & \underline{0.085} & 0.0901 & 0.0163 & 0.635 & 2,901 \\
Info3R\textsuperscript{\ddag}, in full & 0.056 & \underline{0.091} & 0.095 & 0.099 & 0.0888 & 0.0152 & 0.642 & 2,127 \\
TTT3R with reset & 0.096 & 0.115 & 0.163 & 0.167 & 0.1344 & \underline{0.0102} & \textbf{0.548} & 17,825 \\
Every 50 frames\textsuperscript{\dag} & 0.083 & \textbf{0.078} & \textbf{0.092} & \textbf{0.083} & \underline{0.0839} & 0.0199 & 1.066 & \textbf{1,039} \\
\oursrow \scenebound{} (ours) & \textbf{0.048} & \textbf{0.078} & 0.095 & 0.088 & \textbf{0.0804} & 0.0173 & 0.697 & \oursend[3.5pt] 1,423 \\
\midrule
\multicolumn{8}{@{}l}{\emph{KV cache of StreamVGGT}} & \emph{GB} \\
RegVGGT~\cite{mao2026regvggt}, 1\% of keys & 0.058 & 0.099 & 0.116 & 0.195 & 0.1128 & 0.0263 & 0.863 & \textbf{8.3} \\
RegVGGT, 10\% of keys & 0.051 & 0.091 & 0.113 & 0.188 & 0.1067 & 0.0222 & 0.730 & 11.9 \\
InfiniteVGGT~\cite{yuan2026infinitevggt} & 0.067 & 0.129 & 0.161 & 0.178 & 0.1374 & 0.0287 & 0.870 & 16.1 \\
OVGGT~\cite{lu2026ovggt} & 0.071 & 0.192 & 0.167 & 0.203 & 0.1702 & 0.0442 & 1.385 & \underline{8.8} \\
GHOST~\cite{chen2026ghost} & 0.072 & 0.173 & 0.248 & 0.302 & 0.2009 & 0.0382 & 1.793 & 16.5 \\
FrameVGGT~\cite{xu2026framevggt} & 0.047 & 0.139 & 0.153 & 0.133 & 0.1304 & 0.0323 & 1.726 & 23.0 \\
XStreamVGGT~\cite{su2026xstreamvggt} & 0.065 & 0.224 & 0.287 & 0.316 & 0.2354 & 0.0531 & 3.990 & 10.0 \\
RetrieveVGGT~\cite{zou2026attention} & 0.050 & 0.101 & 0.126 & \textbf{0.102} & 0.1019 & 0.0290 & 0.947 & 11.3\rlap{$^{\S}$} \\
Every 50 frames\textsuperscript{\dag} & \underline{0.033} & 0.079 & \textbf{0.101} & 0.124 & 0.0863 & 0.0227 & 0.915 & 11.7 \\
Uniform sampling\textsuperscript{\dag} & \textbf{0.032} & 0.075 & 0.105 & \underline{0.115} & \underline{0.0841} & 0.0193 & 0.711 & 13.3 \\
\scenebound{}, camera head kept whole & 0.051 & \underline{0.073} & 0.105 & 0.123 & 0.0870 & \underline{0.0190} & \underline{0.667} & 13.6 \\
\oursrow \scenebound{} (ours) & \underline{0.033} & \textbf{0.071} & \underline{0.104} & 0.122 & \textbf{0.0831} & \textbf{0.0189} & \textbf{0.666} & \oursend[3.5pt] 13.3 \\
\bottomrule
\end{fulltabular}

\smallskip
\raggedright\scriptsize
$^{\dag}$Constructed baseline; in the cache, a frame it drops leaves the camera head, as for \scenebound{} unless that head is kept whole. $^{\ddag}$Our implementation of a method without released code. $^{\S}$Plus 98.7\,MB of host memory per frame.
\end{table*}

%% file: tab/long-detail.tex
\begin{table*}[t]
\caption{Mean ATE (m) in the recurrent state on the revisit sequences over nine passes, with the update budget, on long sequences and on held-out data, with the number of sequences, and accuracy error (m) of point clouds fused from the first 1,000 frames of benchmark sequences.}
\label{tab:long-detail}
\centering
\footnotesize
\begin{fulltabular}{@{}lcrccrcccccc@{}}
\toprule
 & \multicolumn{2}{c}{Revisits} & \multicolumn{3}{c}{Long sequences, ATE $\downarrow$} & \multicolumn{4}{c}{Held out, ATE $\downarrow$} & \multicolumn{2}{c}{Point cloud, Acc.\ $\downarrow$} \\
\cmidrule(lr){2-3}\cmidrule(lr){4-6}\cmidrule(lr){7-10}\cmidrule(lr){11-12}
Method & ATE $\downarrow$ & Budget $\downarrow$ & Concat., 6 & TUM, 12 & KITTI, 11 & Replica, 8 & EuRoC, 11 & ETH3D, 61 & TUM, 19 & 7-Scenes, 18 & NRGBD, 9 \\
\midrule
CUT3R~\cite{wang2025continuous} & 0.815 & 196,518 & 0.736 & 1.223 & 193.4 & 0.968 & 3.027 & 0.948 & 0.952 & 0.257 & 0.397 \\
AFG-Pose\textsuperscript{\ddag}~\cite{ren2026rethinking} & 0.134 & 5,304 & 0.142 & 0.766 & 175.6 & 0.206 & \textbf{1.245} & 0.628 & 0.183 & 0.029 & 0.085 \\
Info3R\textsuperscript{\ddag}~\cite{baek2026info3r}, gate & \underline{0.124} & 7,642 & 0.228 & 0.772 & 176.7 & \underline{0.152} & 1.743 & 0.604 & 0.182 & 0.026 & 0.085 \\
Info3R\textsuperscript{\ddag}, in full & 0.252 & 7,618 & 0.145 & \textbf{0.445} & 72.3 & 0.156 & 2.053 & 0.535 & 0.195 & 0.026 & \underline{0.084} \\
TTT3R~\cite{chen2026ttt3r} with reset & 0.573 & 68,399 & 0.478 & \underline{0.449} & \underline{61.7} & 0.353 & 2.261 & \underline{0.525} & 0.230 & 0.085 & 0.139 \\
Every 50 frames\textsuperscript{\dag} & 0.130 & \underline{3,936} & 0.142 & 0.735 & 175.8 & \textbf{0.142} & \underline{1.301} & 0.672 & 0.178 & \textbf{0.024} & \textbf{0.077} \\
\scenebound{}, one state & \textbf{0.084} & \textbf{698} & \textbf{0.138} & 0.713 & 177.2 & 0.184 & 1.780 & 0.583 & \underline{0.169} & \underline{0.025} & 0.086 \\
\oursrow \scenebound{} (ours) & \textbf{0.084} & \textbf{698} & \underline{0.139} & 0.503 & \textbf{50.7} & 0.181 & 2.140 & \textbf{0.492} & \textbf{0.127} & 0.026 & \oursend 0.086 \\
\bottomrule
\end{fulltabular}

\smallskip
\raggedright\scriptsize
$^{\dag}$Constructed baseline. $^{\ddag}$Our implementation of a method without released code.
\end{table*}

%% file: tab/published.tex
\begin{table}[t]
\caption{TUM-dynamics at 1,000 frames, published values beside \scenebound{}, ATE (m).}
\label{tab:published}
\centering
\footnotesize
\begin{fulltabular}{@{}lclc@{}}
\toprule
Training-free & ATE $\downarrow$ & Trained / ours & ATE $\downarrow$ \\
\midrule
TTT3R~\cite{chen2026ttt3r} with reset & 0.093 & Mem3R + TTSA3R~\cite{liu2026mem3r} & 0.064 \\
ReCal3R~\cite{li2026recal3r} & 0.077 & PAS3R~\cite{xu2026pas3r} & 0.052 \\
Info3R~\cite{baek2026info3r} & 0.061 & LoGeR~\cite{zhang2026loger}, hybrid & 0.050 \\
OVGGT~\cite{lu2026ovggt}, cache & 0.058 & \cellcolor{ours}\scenebound{}, state & \cellcolor{ours}\underline{0.048} \\
AFG-Pose~\cite{ren2026rethinking} & 0.054 & \cellcolor{ours}\scenebound{}, cache & \cellcolor{ours}\textbf{0.033} \\
\bottomrule
\end{fulltabular}

\smallskip
\raggedright\scriptsize
The radius of \scenebound{} was chosen on these eight sequences, and radii from 0.25 to 0.35 give 0.048 to 0.050 in the state.
\end{table}

%% file: tab/cache.tex
\begin{table}[t]
\caption{The KV cache of StreamVGGT on the 17 revisit sequences after one pass and after all nine, with the setting that bounds each cache.}
\label{tab:cache}
\centering
\footnotesize
\begin{fulltabular}{@{}llccr@{\hspace{3.5pt}}}
\toprule
 & & \multicolumn{2}{c}{ATE (m) $\downarrow$} & GPU (GB) $\downarrow$ \\
\cmidrule(lr){3-4}\cmidrule(l){5-5}
Method & Setting & 1 pass & 9 passes & 9 passes \\
\midrule
RegVGGT~\cite{mao2026regvggt} & 1\% of keys & 0.131 & 0.129 & 15.2 \\
RegVGGT & 10\% of keys & 0.124 & 0.138 & 52.6 \\
InfiniteVGGT~\cite{yuan2026infinitevggt} & 1.2M tokens & 0.125 & 0.335 & 18.1 \\
OVGGT~\cite{lu2026ovggt} & 0.2M tokens & 0.141 & 0.166 & \textbf{8.9} \\
GHOST~\cite{chen2026ghost} & 1.2M tokens & 0.194 & 0.768 & 18.6 \\
FrameVGGT~\cite{xu2026framevggt} & $\le$82 frames & 0.093 & 0.077 & 33.1 \\
RetrieveVGGT~\cite{zou2026attention} & pose-aware & 0.085 & 0.080 & 15.6\rlap{$^{\S}$} \\
Every 50 frames\textsuperscript{\dag} & 2.0\% of frames & 0.081 & \textbf{0.067} & 49.6 \\
Uniform sampling\textsuperscript{\dag} & 2.8\% of frames & \textbf{0.076} & 0.076 & \underline{14.7} \\
\oursrow \scenebound{} (ours) & 2.8\% of frames & \underline{0.080} & \underline{0.069} & \oursend[3.5pt] 14.8 \\
\bottomrule
\end{fulltabular}

\smallskip
\raggedright\scriptsize
$^{\dag}$Constructed baseline. $^{\S}$Plus 211\,GB of host memory.
\end{table}

%% file: supp/E_limitations.tex
\section{Limitations}
\label{app:limitations}

Our analysis covers the recurrent state of two released checkpoints of CUT3R, the fast weights of ZipMap and the KV caches of StreamVGGT and STream3R.
STAC, RetrieveVGGT and OVGGT report that scored retention beats random retention; of the scored caches, we compare only RegVGGT with a signal-free rule at equal budget.

The keyframe test and the drift check act on predicted poses, so their guarantees carry pose accuracy as a condition.
A frame whose predicted pose jumps beyond $M$ from its predecessor is written as a new view, and on one of the 17 revisit sequences in the cache such writes raise the error of all later passes (\cref{app:tests}).
Every later pass reads these misplaced views, as errors in the experience that a language-model agent distils into skills persist into its later tasks~\cite{yu2026rsebench}.
A first frame that sees the scene closer than the frames after it sets a finer depth unit for the whole video and adds writes (\cref{tab:unit}).

A return finds a stored submap by predicted pose up to an accumulated drift of one keyframe radius, and place recognition on the stored keyframes extends returns beyond this radius.
The error of \scenebound{} on long drives lies in placing its submaps (\cref{tab:pieces}); a pose graph whose loops the frozen model confirms by its own reads raises this error (\cref{tab:graph}), so global placement needs a back end with independent place recognition and geometric verification.
On KITTI, resets at fixed intervals with \scenebound{} inside are ahead of \scenebound{}, and Info3R, with constants set for driving, and AFG-Pose, both without released code, report lower errors.
On EuRoC, submaps raise the error compared with one state, and in the second CUT3R checkpoint writes at new views are on par with uniform sampling.

In the KV cache on Neural RGB-D, RetrieveVGGT and OVGGT, which read tokens of every earlier frame, fuse more accurate clouds than \scenebound{} (\cref{app:points}).
\scenebound{} keeps one state of about 18\,MB per submap, so its memory grows with the visited region; a closed submap is read only on a return and can wait in host memory, where each return adds a host-to-device transfer, a cost that lies on the critical path of decoding when mixture-of-experts language models offload their experts~\cite{cao2026specfit}.

Moving people enter memory as the rest of the scene does, and a stored view stays in memory after the scene it shows has changed; longer-term changes call for a test that retires stored views, as the persistent memory of a language-model agent resolves a new fact that contradicts a stored one~\cite{wang2026toki}.
Executing a program renders a video together with the ground-truth answer and time interval of every question about it~\cite{zhu2026tempozero}, so a program that changes a scene at known times could supply the ground truth for such a test.